\documentclass{article} 
\usepackage{iclr2027_conference,times}

\usepackage{amsmath,amsfonts,bm}

\def\eqref#1{equation~\ref{#1}}

\def\1{\bm{1}}

\DeclareMathAlphabet{\mathsfit}{\encodingdefault}{\sfdefault}{m}{sl}
\SetMathAlphabet{\mathsfit}{bold}{\encodingdefault}{\sfdefault}{bx}{n}

\usepackage[hidelinks]{hyperref}
\usepackage{url}
\usepackage{booktabs}
\usepackage{xltabular}
\usepackage{multirow}
\usepackage{amssymb}
\usepackage{xcolor}
\usepackage{colortbl}
\usepackage{graphicx}
\usepackage{float}
\usepackage{tikz}
\usetikzlibrary{fadings}

\definecolor{titleblue}{RGB}{76,133,178}
\definecolor{titlecream}{RGB}{232,216,181}
\newcommand{\gradienttitle}{%
  {\bfseries
  \textcolor{titleblue}{R}%
  \textcolor{titleblue!80!titlecream}{a}%
  \textcolor{titleblue!60!titlecream}{w}%
  \textcolor{titleblue!40!titlecream}{V}%
  \textcolor{titleblue!20!titlecream}{L}%
  \textcolor{titlecream}{A}}%
}

\title{\texorpdfstring{\gradienttitle}{RawVLA}: Embodied Neural Image Signal Processor For Robotic Manipulation}

\author{
\normalfont\bfseries
Shuhong Liu\textsuperscript{1,2},
Heng Zhou\textsuperscript{2,\textdaggerdbl},
Lingfeng Qian\textsuperscript{2},
Yuhao Fang\textsuperscript{2},
Xianbao Hou\textsuperscript{2},
\\
\bfseries
Qianyu Zhou\textsuperscript{1},
Lin Gu\textsuperscript{4},
Wei Sui\textsuperscript{2},
Jianfei Yang\textsuperscript{3},
Ziteng Cui\textsuperscript{1,5,\textdagger}
\\
\small\normalfont
\textsuperscript{1}UTokyo
\quad
\textsuperscript{2}D-Robotics
\quad
\textsuperscript{3}NTU
\quad
\textsuperscript{4}TohokuU
\quad
\textsuperscript{5}HKUSTGZ
\\
\textsuperscript{\textdaggerdbl}Project lead
\quad
\textsuperscript{\textdagger}Corresponding author
}

\newcommand{\methodname}{RawVLA}
\newcommand{\benchmarkname}{RawVLA-Bench}
\definecolor{changegreen}{RGB}{42,112,72}
\definecolor{changered}{RGB}{158,58,58}
\definecolor{improvementgreen}{RGB}{83,126,98}
\definecolor{fst}{RGB}{165,199,226}
\definecolor{sed}{RGB}{216,235,244}
\definecolor{thd}{RGB}{253,244,225}
\definecolor{codeblue}{RGB}{76,133,178}
\definecolor{linkcream}{RGB}{176,145,94}
\hypersetup{
  hypertexnames=false,
  colorlinks=true,
  linkcolor=linkcream,
  citecolor=codeblue,
  urlcolor=codeblue
}
\newcommand{\positivechange}[1]{\rlap{\hspace{0.20ex}\scalebox{0.80}{\textcolor{changegreen}{\ensuremath{+#1}}}}}
\newcommand{\negativechange}[1]{\rlap{\hspace{0.20ex}\scalebox{0.80}{\textcolor{changered}{\ensuremath{-#1}}}}}
\newsavebox{\rankresultbox}
\newcommand{\rankshade}[2]{%
  \begingroup
  \sbox{\rankresultbox}{#2}%
  \leavevmode
  \rlap{\hspace{-1.2pt}\raisebox{0pt}[0pt][0pt]{%
    \setlength{\fboxsep}{1.2pt}%
    \colorbox{#1}{\phantom{\usebox{\rankresultbox}}}}}%
  \usebox{\rankresultbox}%
  \endgroup
}
\newcommand{\firstbest}[1]{\rankshade{fst}{#1}}
\newcommand{\secondbest}[1]{\rankshade{sed}{#1}}
\newcommand{\thirdbest}[1]{\rankshade{thd}{#1}}

\iclrfinalcopy 
\begin{document}

\maketitle

\vspace{-8pt}
\begin{figure}[H]
\raggedright
\includegraphics[width=\textwidth]{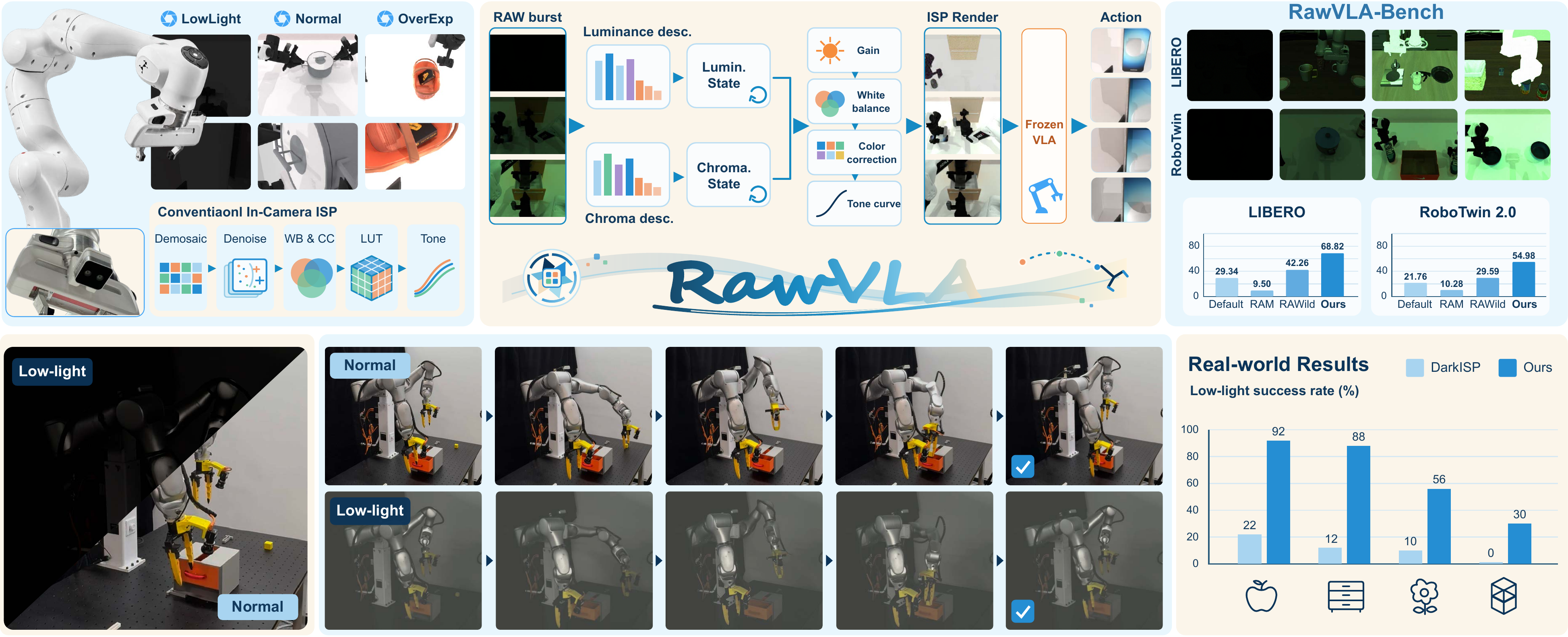}
\vspace{-4pt}
\caption{We propose \methodname{}, an embodied and adaptive neural ISP module for VLA models, and
\benchmarkname{}, a RAW-domain manipulation benchmark based on LIBERO and
RoboTwin~2.0.}
\label{fig:teaser}
\end{figure}

\begin{abstract}
Vision-language-action (VLA) models typically operate on RGB images produced by a
fixed camera image signal processor (ISP), leaving the imaging pipeline outside the learning
and evaluation loop. We systematically examine the consequences of this
overlooked design choice across five fundamental ISP dimensions: gain, sensor noise, chromatic response, tonal response, and bit depth. Our analysis reveals
that RAW-to-RGB processing materially shapes both action prediction and
manipulation success, with different ISP dimensions exerting substantially
different effects. Guided by these findings, we introduce \methodname, a
streaming neural ISP that adaptively renders RAW observations for frozen VLA
policies while concentrating its capacity on the imaging factors relevant
to embodied behavior. We further present \benchmarkname, a RAW-domain
manipulation benchmark to expose image processing as an explicit evaluation variable across clean and adverse
acquisition conditions. Experiments on \benchmarkname{} show that \methodname{}
preserves performance under standard conditions while substantially improving
robustness under degraded imaging, establishing adaptive RAW processing as an
effective interface between physical cameras and embodied policies. Code available at
\href{https://shuhongll.github.io/rawvla}{\textcolor{linkcream}{\nolinkurl{https://shuhongll.github.io/rawvla}}}.
\end{abstract}

\section{Introduction}

Developing agents that can perceive, reason, and act in the physical world is
a central goal of embodied AI and physical intelligence
\citep{brohan2023rt2,black2024pi0}. Vision-language-action
(VLA) models offer a promising path toward this goal by unifying visual
observations, language instructions, and robot actions
\citep{brohan2023rt2,kim2024openvla,black2024pi0,physicalintelligence2025pi05}.
Recent systems further scale continuous action generation, cross-embodiment
learning, and embodied reasoning through diffusion transformers and generalist
robot foundation models
\citep{hou2024ditpolicy,bjorck2025gr00tn1,geminirobotics2025}. Leveraging
vision-language pretraining and robot demonstrations, they can acquire
generalizable manipulation capabilities across tasks, scenes, and robot
embodiments
\citep{octo2024,kim2024openvla,li2024robovlms,physicalintelligence2025pi05}.
However, existing VLA
research typically begins with camera-rendered RGB images
\citep{brohan2023rt2,kim2024openvla} and abstracts the visual pathway as
\emph{RGB-to-action}. This abstraction overlooks the physical imaging pipeline
that precedes model inference \citep{brooks2019unprocessing,wu2019visionisp}.
Light from a scene is captured as RAW sensor measurements and then transformed
into RGB images by an image signal processor (ISP)
\citep{brooks2019unprocessing,yu2021reconfigisp}. Consequently, systems
implicitly treat the ISP as a fixed and task-neutral preprocessing component
\citep{diamond2021dirtypixels,huang2025drraw}.

This assumption can constrain downstream robot control. ISPs adjust exposure, white
balance and color, tone, denoising, and quantization
\citep{brooks2019unprocessing}, potentially discarding signals needed for
manipulation \citep{diamond2021dirtypixels,huang2025drraw}. Embodied Image
Compression \citep{li2025embodiedcompression} and SPARC \citep{kim2026sparc}
further show that visual processing should be evaluated by closed-loop task
performance rather than perceptual fidelity. Consequently, an unchanged scene
can induce different VLA behaviors across ISP configurations
\citep{wang2024adaptiveisp,huang2025drraw}. This issue is especially important
under low light, high dynamic range, backlighting, and unusual illumination
\citep{guo2025darkisp,fei2025liberoplus}, where clipping, noise, or aggressive
tone compression can obscure task-relevant visual details
\citep{chen2018sid,watanabe2026flare}.

RAW observations offer a natural opportunity to revisit this overlooked part
of the VLA system \citep{cui2024rawadapter,huang2025drraw}.
Compared with a single RGB image produced by a fixed ISP, RAW measurements
preserve sensor information that can be weakened or irreversibly lost during
RGB rendering \citep{chen2018sid,huang2025drraw}. They
also retain the flexibility to select or learn a RAW-to-RGB transformation
according to the scene, task, and downstream model
\citep{yu2021reconfigisp,wang2024adaptiveisp}.
These properties suggest that the imaging pipeline should be treated as an
explicit and optimizable component of a VLA system rather than as hidden
camera firmware \citep{diamond2021dirtypixels,huang2025drraw}.
Nevertheless, existing task-oriented ISP studies focus mainly on static vision
tasks \citep{cui2024rawadapter,liu2026rawild}, leaving open
how strongly ISP choices affect VLA behavior and whether task-adaptive
RAW-to-RGB processing can improve manipulation performance.

In this work, we conduct a systematic analysis of VLA sensitivity to five
fundamental ISP dimensions: exposure, sensor noise, chromatic response, tonal response, and bit depth \citep{brooks2019unprocessing,yu2021reconfigisp}. By varying one dimension at a time while holding the underlying
scene, task, and robot state fixed, we isolate the effect of RAW-to-RGB
processing on downstream action prediction and task success. Our results show
that ISP processing is far from task-neutral and that different ISP dimensions
affect VLA models to substantially different degrees. These observations both
motivate RAW-domain evaluation and provide direct guidance for designing an
adaptive ISP.

We propose \textbf{\methodname}, a lightweight neural ISP that adaptively
renders RAW inputs for downstream VLA models, as illustrated in
Figure~\ref{fig:teaser}. Unlike a conventional
frozen ISP, \methodname{} adapts behavior-relevant imaging factors without
modifying the VLA backbone. To evaluate RAW-domain processing, we introduce
\textbf{\benchmarkname}, a benchmark that exposes
the RAW-to-RGB pipeline as an explicit degree of freedom. \benchmarkname{}
spans clean and challenging acquisition conditions for evaluating robustness
and RAW-domain processing headroom, complementing RGB
robustness benchmarks
\citep{fei2025liberoplus,chen2025robotwin,zhang2026wamrobustness}.

We summarize our contributions below.
\begin{itemize}
    \item We systematically analyze how the imaging pipeline affects VLA
    models through gain, sensor noise, chromatic response, tonal response, and
    bit depth.
    Our analysis demonstrates that ISP choices materially influence VLA action
    prediction.
    \item We construct \textbf{\benchmarkname}, a RAW-domain benchmark that makes
    image processing an explicit evaluation variable and supports controlled
    study of VLA robustness under clean and challenging imaging conditions.
    \item We propose \textbf{\methodname}, a real-time neural ISP for VLA
    models. Extensive experiments on \benchmarkname{} and real-world tasks demonstrate state-of-the-art
    performance.
\end{itemize}

\section{Analyzing VLA Sensitivity to ISP}
\label{sec:isp_sensitivity}

To identify the imaging factors relevant to embodied actions, we conduct a controlled ISP sensitivity study on LIBERO and RoboTwin~2.0. We synthesize pseudo-RAW inputs from renderer buffers via parametric unprocessing \citep{brooks2019unprocessing}. Following standard camera-formation and ISP decompositions \citep{yu2021reconfigisp}, we examine five principal
factors governing signal level, acquisition noise, color, tone, and numerical precision. Holding the underlying pseudo-RAW observation $R$ fixed, we isolate
their effects through the factorized RAW-to-RGB mapping below, with detailed formulations provided in Appendices~\ref{app:unprocessing} and~\ref{app:perturbation_settings}:
\begin{equation}
Y_{e,\eta,\zeta,c_{\mathrm{tone}},n}
=\mathcal{Q}_{n}\!\left(
\mathcal{T}_{c_{\mathrm{tone}}}\!\left(
\mathcal{C}_{\zeta}\!\left(
\mathcal{N}_{\eta}\!\left(
\mathcal{E}_{e}(R)
\right)\right)\right)\right).
\label{eq:isp_analysis_overview}
\end{equation}

\begin{figure}[!t]
\centering
\includegraphics[width=\textwidth]{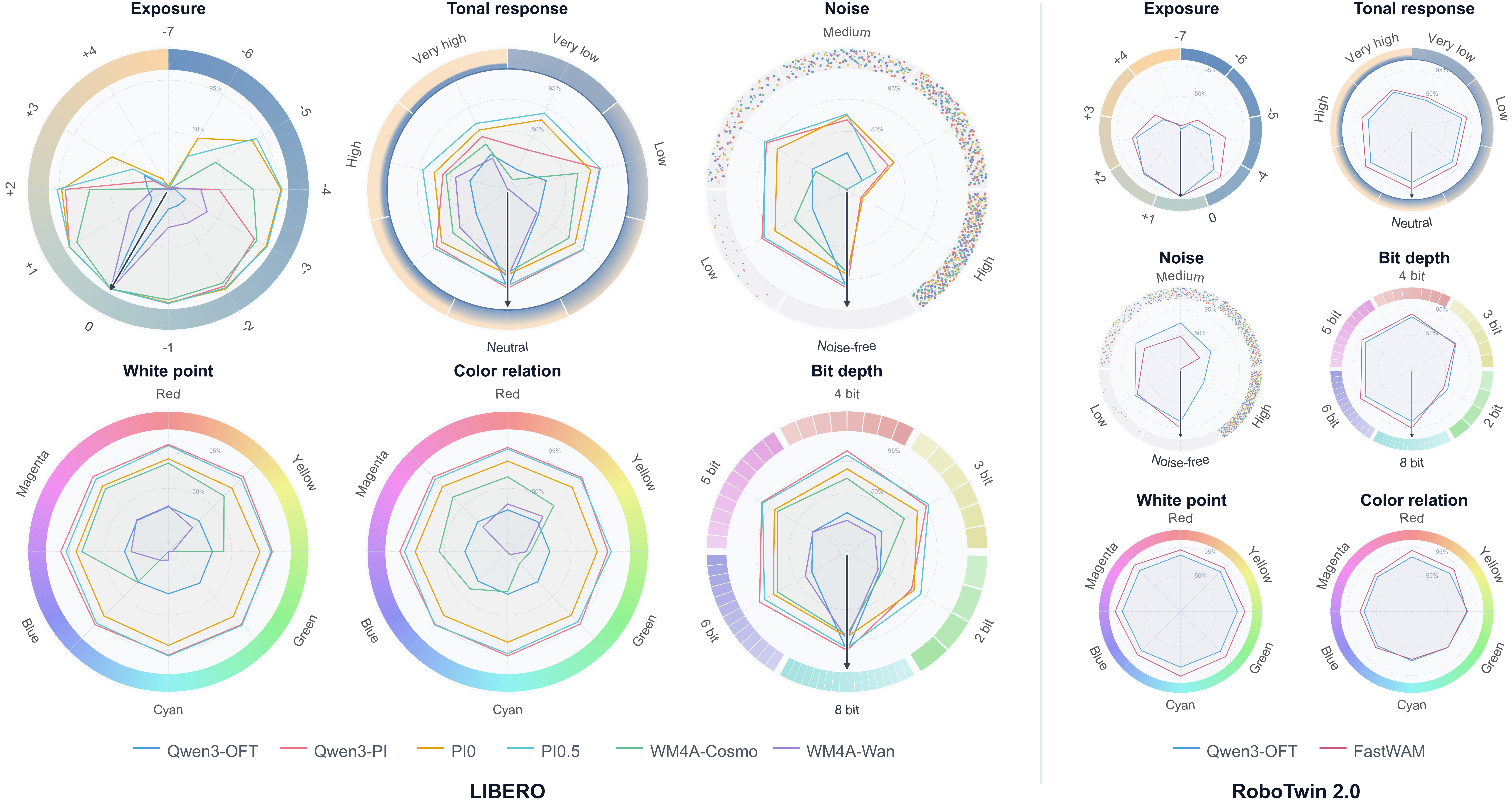}
\vspace{-5pt}
\caption{ISP sensitivity across simulation benchmarks. The left panel shows LIBERO results for fine-tuned VLA and WAM policies provided by StarVLA \citep{starvla2026}. The right panel shows Qwen3-OFT and FastWAM fine-tuned on RoboTwin~2.0 Easy and Hard (Hard adds randomized clutter, lighting, textures,
and tabletop heights), which moderately improves robustness.}
\label{fig:isp_analysis}
\end{figure}

\paragraph{Exposure.}
$\mathcal{E}_e$ scales the RAW signal to control shadow visibility and highlight saturation. As shown in Figure~\ref{fig:isp_analysis}, exposure is the most influential axis. VLA models maintain high task performance only within a limited range around the default and collapse under severe underexposure or overexposure. World-action model (WAM) approaches have the narrowest stable range. Both model families on RoboTwin~2.0 remain vulnerable at extreme exposure settings despite fine-tuning with visual domain randomization.

\paragraph{Sensor Noise.}
$\mathcal{N}_\eta$ isolates sensor-noise sensitivity by varying noise independently of its physical coupling with illumination. VLA models tolerate mild noise after brightness restoration but degrade rapidly from moderate noise onward. WAMs degrade earlier and more sharply than VLA PI-family policies. On RoboTwin~2.0, Qwen3-OFT retains partial task success under high noise whereas FastWAM approaches complete failure.

\paragraph{Chromatic Response.}
$\mathcal{C}_\zeta$ changes chromatic rendering through absolute white-point shifts and relative hue and saturation transformations while preserving the underlying scene. VLA PI-family models benefit from visual augmentation during training and remain comparatively robust to these changes whereas WAMs exhibit stronger directional preferences. On LIBERO, the WAM agents reveal a pronounced color bias, preferring warmer renderings to cooler ones. After fine-tuning on the visually randomized RoboTwin~2.0 (Hard), Qwen3-OFT and FastWAM still exhibit white-point-dependent bias, despite remaining stable across relative-color transformations.

\paragraph{Tonal Response.}
$\mathcal{T}_{c_{\mathrm{tone}}}$ controls the distribution of luminance contrast from shadows through mid-tones to highlights. VLA policies retain high success near the default response but degrade abruptly when the response becomes very flat or highly contrasted. WAMs are consistently more sensitive across the tonal range. Both model families on RoboTwin~2.0 respond more smoothly near the default setting but still lose substantial task performance at both extremes.

\paragraph{Bit Depth.}
$\mathcal{Q}_{n}$ quantizes the rendered output to $n$ bits for digital image encoding. Performance remains largely stable above two bits and drops only under the most aggressive quantization. VLA models remain stable down to three bits and show a clear decline at two bits whereas WAMs begin to weaken earlier. Overall, bit depth affects both model families less than other ISP dimensions.

\begin{figure}[!tp]
    \centering
    \includegraphics[width=\textwidth]{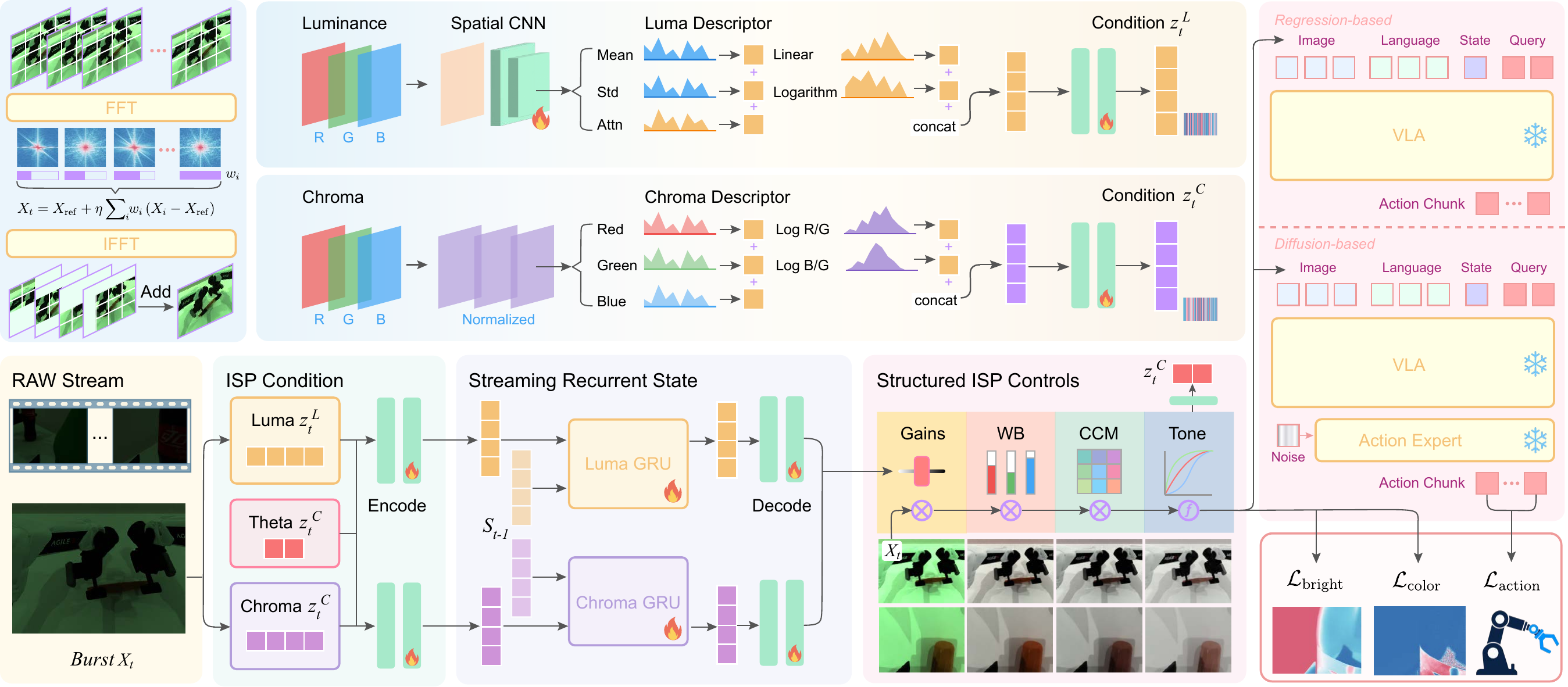}
    \vspace{-5pt}
    \caption{Overview of \methodname. A causal linear RAW burst is summarized by
    factorized luminance and chroma descriptors, fused with previous ISP
    parameters through decoupled recurrent states, and decoded into structured
    exposure, white-balance, color-correction, and monotonic tone
    controls. The rendered image is passed to a frozen VLA policy for action
    prediction.}
    \label{fig:rawvla_pipeline}
\end{figure}

\section{\methodname}
\label{sec:rawvla}

To account for the observed variation in sensitivity, we design \methodname{},
a causal, structured ISP that selectively adapts behavior-relevant photometric
controls while keeping the VLA policy frozen.
We next describe its
streaming formulation (Section~\ref{sec:rawvla_formulation}), recurrent
luminance--chroma conditioning (Section~\ref{sec:rawvla_conditioning}),
structured decoding (Section~\ref{sec:rawvla_isp}), and action-driven
optimization (Section~\ref{sec:rawvla_training}).

\subsection{Causal Streaming Formulation}
\label{sec:rawvla_formulation}

\methodname{} is a causal, task-driven ISP that transforms demosaiced linear RAW
observations for a downstream VLA policy. It obtains the denoised RAW frame
$X_t$ from consecutive RAW frames through burst denoising while recurrently adapting factorized
luminance and chroma controls. To preserve temporal context and avoid
frame-to-frame rendering discontinuities, \methodname{} formulates the streaming update
as
\begin{equation}
    (Y_t,S_t,\Theta_t)
    =
    \operatorname{\methodname}
    (X_t,S_{t-1},\Theta_{t-1}),
\end{equation}
where $Y_t$ is the rendered observation passed to the VLA policy,
$S_t$ provides latent temporal context for illumination and color adaptation,
and $\Theta_t$ is the explicit ISP parameter set controlling the current
RAW-to-RGB processing.

\subsection{Recurrent Photometric Conditioning}
\label{sec:rawvla_conditioning}

A central difficulty in learning an adaptive ISP is parameter identifiability,
as global brightness can otherwise be explained by exposure, white balance,
color correction, or channel-dependent tone mapping
\citep{brooks2019unprocessing,yu2021reconfigisp}. \methodname{} resolves this ambiguity
by encoding orthogonal achromatic and chromatic conditions into decoupled
recurrent states.

\paragraph{Luminance State.}
We compute an equal-channel luminance map $L_t$\footnote{We define
$L_t=\frac{1}{3}\sum_{k\in\{\mathrm R,\mathrm G,\mathrm B\}}X_t^k$ to capture achromatic intensity structure.}, from which we extract a compact spatial
luminance feature $g_t$. We complement $g_t$ with an absolute luminance
descriptor $h_t^L$ that summarizes intensity distributions, quantiles, and
dark-clipped pixel ratios detailed in
Appendix~\ref{app:photometric_conditioning}. The logarithmic representation emphasizes
low-intensity variation, whereas the linear histogram and clipping statistics
capture over-exposure. The luminance
recurrent state is then updated with a GRU \citep{cho2014gru} as
\begin{equation}
    z_t^L=\phi_L(g_t,h_t^L,\Theta_{t-1}),
    \qquad
    S_t^L=\operatorname{GRU}_L(z_t^L,S_{t-1}^L).
\end{equation}
Here, $\phi_L$ combines the current spatial feature $g_t$, the luminance
descriptor $h_t^L$, and the previous ISP parameters $\Theta_{t-1}$. The
recurrent state $S_t^L$ provides temporal context for predicting the current
exposure and tone parameters.

\paragraph{Chroma State.}
To prevent global brightness from leaking into color estimation, the chroma
pathway uses only scale-invariant channel statistics. We compute chromaticities
$\boldsymbol{\chi}_t$ and log-chrominance ratios ${\mathbf q}_t$ directly from the
RAW channels,
\begin{equation}
    [\boldsymbol{\chi}_t]^k
    =
    \frac{X_t^k}
    {\sum_{j\in\{\mathrm R,\mathrm G,\mathrm B\}}X_t^j+\epsilon},
    \quad k\in\{\mathrm R,\mathrm G,\mathrm B\},
    \quad
    {\mathbf q}_t
    =
    \left(
        \log\frac{X_t^{\mathrm R}+\epsilon}{X_t^{\mathrm G}+\epsilon},
        \log\frac{X_t^{\mathrm B}+\epsilon}{X_t^{\mathrm G}+\epsilon}
    \right).
\end{equation}
We form the chroma descriptor $h_t^C$ by concatenating histograms together with
first- and second-order statistics, as detailed in
Appendix~\ref{app:chroma_descriptor}.
We fuse $h_t^C$ with the previous ISP parameters $\Theta_{t-1}$ through
$\phi_C$ and update the chroma recurrent state as
\begin{equation}
    z_t^C=\phi_C(h_t^C,\Theta_{t-1}),
    \qquad
    S_t^C=\operatorname{GRU}_C(z_t^C,S_{t-1}^C).
\end{equation}
The pathway receives neither absolute luminance statistics nor RGB spatial
features. Together, the luminance and chroma states form the full recurrent
state $S_t=[S_t^L,S_t^C]$.

\subsection{Structured Neural ISP Decoding}
\label{sec:rawvla_isp}

From the decoupled recurrent state $[S_t^L,S_t^C]$, \methodname{} predicts the
structured ISP parameter set
$\Theta_t=\{e_t,w_t,A_t,\ell_t\}$, where $e_t$ specifies exposure in EV, $w_t$ encodes white balance,
$A_t$ denotes the $3\times3$ color-correction matrix, and
$\ell_t\in\mathbb{R}^{7}$ parameterizes the tone curve. \methodname{} first applies
fixed six-frame burst denoising to linear RAW inputs, producing $X_t$ as shown
in Figure~\ref{fig:rawvla_pipeline} (top left).

\paragraph{Exposure and Color Correction.}
We parameterize global exposure in EV space as
$e_t=E_{\max}\tanh(\hat e_t)$, with gain $2^{e_t}$, where $\hat e_t$ is the predicted exposure logit, following standard RAW processing
conventions \citep{brooks2019unprocessing}. White balance uses zero-sum channel
offsets and the corresponding diagonal gain matrix,
\begin{equation}
    w_t=(w_t^{\mathrm R},w_t^{\mathrm G},w_t^{\mathrm B})
    \quad\text{with}~
    \sum w_t^k=0,
    \qquad\text{and}\qquad
    W_t=\operatorname{diag}(2^{w_t^{\mathrm R}},2^{w_t^{\mathrm G}},2^{w_t^{\mathrm B}}),
\end{equation}
which restricts white balance to relative channel gains while $e_t$ controls
global intensity. The two-coordinate parameterization is detailed in
Appendix~\ref{app:isp_parameterization}. The color-correction matrix uses an
identity-centered bounded residual, and the linear ISP output is processed as
\begin{equation}
    A_t=\mathbb{I}_3+\Delta A_t,
    \qquad Z_t=2^{e_t}A_tW_tX_t.
\end{equation}
$\Delta A_t$ is the predicted residual; white balance and color correction
precede exposure adjustment.

\paragraph{Monotonic Achromatic Tone-mapping.}
\methodname{} predicts a single tone curve shared across RGB channels, preventing the
luminance pathway from introducing color casts. We use an eighth-degree
Bernstein polynomial \citep{farouki2012bernstein}. We set $p_0=0$ and construct
the remaining control points directly from the predicted logits as
$p_k=\sum_{j=1}^{k}[\operatorname{softmax}([\ell_t,0])]_j$, yielding the curve
\begin{equation}
    T_t(x)
    =
    \sum_{k=0}^{8}
    p_k \binom{8}{k}
    x^k(1-x)^{8-k}
\end{equation}
maps a normalized channel intensity $x\in[0,1]$. It is monotonic and endpoint
preserving, with zero logits corresponding exactly to the identity mapping.
The final VLA-facing image is
\begin{equation}
    Y_t
    =
    \operatorname{clip}
    \left(
        T_t\!\left(
            \operatorname{clip}(Z_t,0,1)
        \right),
        0,1
    \right).
\end{equation}

\subsection{Action-Driven Optimization}
\label{sec:rawvla_training}

\methodname{} is optimized through the native action objective of the
downstream VLA policy. This objective retains the backbone's original training
formulation across direct action regression \citep{kim2025oft} and
diffusion-based generation \citep{chi2023diffusionpolicy} and is denoted
generically as
\begin{equation}
    \mathcal{L}_{\mathrm{action}}
    =
    \mathcal{L}_{\mathrm{VLA}}(Y_t,\mathbf{a}_t),
\end{equation}
where $\mathbf{a}_t$ is the ground-truth action signal. The VLA parameters remain fixed, while gradients through $Y_t$ update
\methodname{}. The learned ISP is therefore trained with an action-driven objective
without requiring pixel-wise supervision.

Although the action objective effectively aligns ISP rendering with downstream
policy behavior, extreme illumination can hinder \methodname{} from learning
aggressive exposure adjustments and reliable chromatic correction early in
training. We therefore introduce two pathway-specific regularizers
\begin{equation}
    \mathcal{L}_{\mathrm{bright}}=\frac{1}{B}\sum_{b=1}^{B}\left|\mu(Y_t^{(b)})-0.48\right|,
    \qquad
    \mathcal{L}_{\mathrm{chroma}}=\frac{1}{B}\sum_{b=1}^{B}\operatorname{dist}_{\mathrm{SL1}}\!\left(\mathbf{d}(Y_t^{(b)}),\Omega_C\right).
\end{equation}
In this expression, $\mu(\cdot)$ is the mean image intensity,
$\mathbf{d}(\cdot)$ denotes the output log-chroma descriptor, $B$ is the batch
size, and $\Omega_C$ is the log-chroma range estimated from RGB images
randomly sampled in the simulator. The smooth distance term
$\operatorname{dist}_{\mathrm{SL1}}(\cdot)$ penalizes violations outside
$\Omega_C$.
The complete objective therefore becomes
\begin{equation}
    \mathcal{L}
    =
    \lambda_a \mathcal{L}_{\mathrm{action}}
    +
    \lambda_b \mathcal{L}_{\mathrm{bright}}
    +
    \lambda_{\mathrm{chroma}} \mathcal{L}_{\mathrm{chroma}},
\end{equation}
where $\lambda_a$ is model-specific due to diverse action losses, while
$\lambda_b=0.001$ and
$\lambda_{\mathrm{chroma}}=0.001$. The complete training settings are given in
Appendix~\ref{app:training_details}.


\begin{figure}[!tp]
\centering
\includegraphics[width=\textwidth]{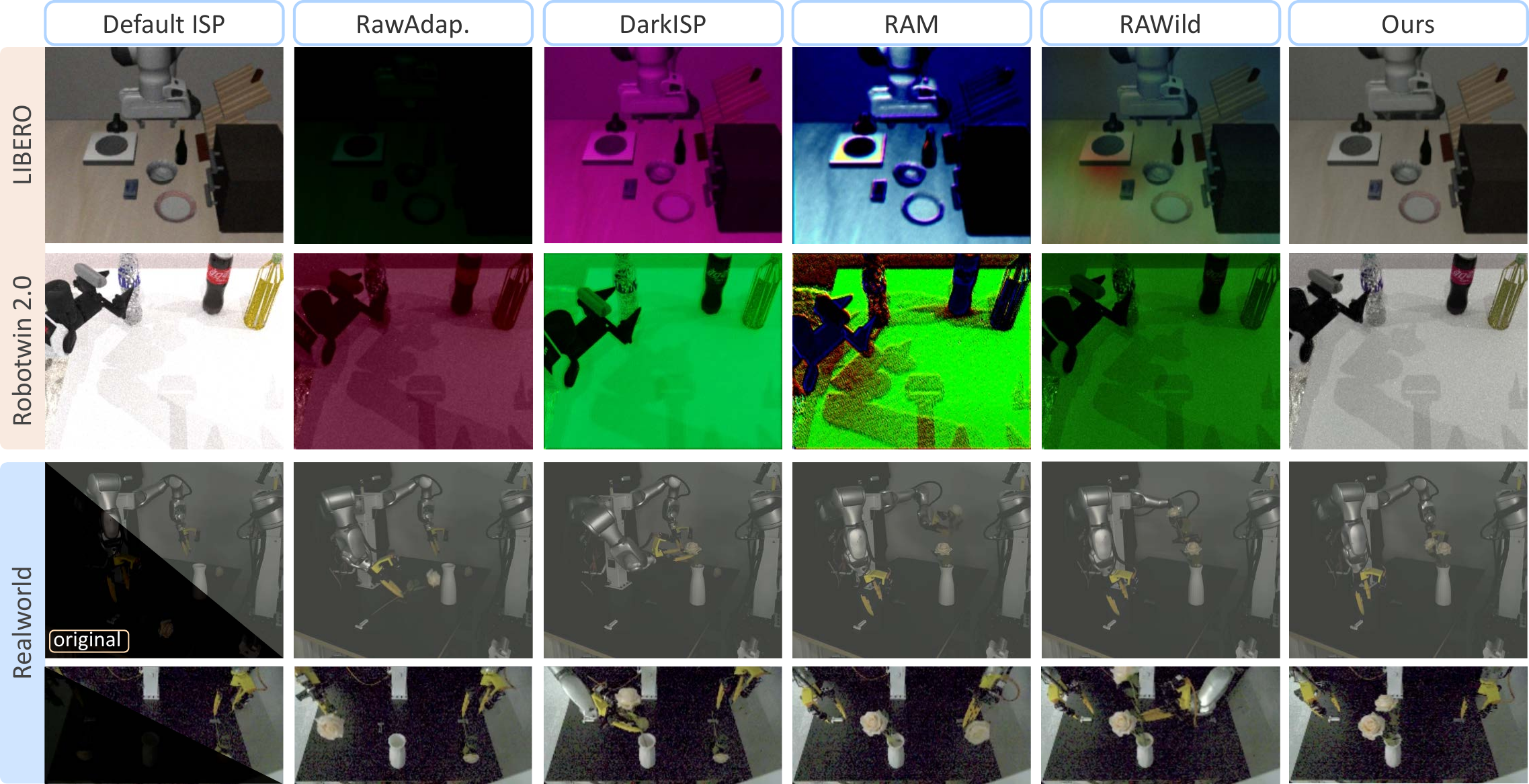}
\vspace{-5pt}
\caption{Qualitative results in simulation and the real world. The first and
second rows compare agent-view observations rendered by the Default ISP and
\methodname{} on LIBERO and RoboTwin~2.0, respectively. The final row shows a
real-world task completed under low-light conditions.}
\label{fig:experiments}
\end{figure}

\section{\benchmarkname}

\benchmarkname{} exposes image acquisition as an explicit evaluation variable
by changing simulator illumination before RAW formation. We sample five
illumination regimes spanning \emph{ExtremeLow}, \emph{Low}, \emph{Normal},
\emph{Over}, and \emph{ExtremeOver}, with environment-specific ranges calibrated
for comparable visual difficulty. To synthesize expert trajectories in the RAW
domain while preserving successful expert actions and environment states, we
replay demonstrations in both benchmarks. For LIBERO, we replay 2,000 official
demonstrations from its four suites and retain 1,771 successful trajectories,
pairing normal- and target-light observations under identical states and
actions. Because MuJoCo exposes only tone-mapped RGB8, we apply fixed
unprocessing \citep{brooks2019unprocessing} to produce three-channel
pseudo-RAW10 for the $256\times256$ agent and wrist views. For RoboTwin~2.0, we
replay 650 demonstrations from 13 dual-arm tasks and obtain 632 successful
trajectory pairs, synchronizing clean and target-light environments at every
frame. We form RAW observations directly from SAPIEN pre-tonemap HDR buffers
with heteroscedastic sensor noise, using head and dual-wrist views at
$240\times320$. Detailed construction and evaluation settings are provided in
Appendix~\ref{app:rawvla_bench}.

\begin{table}[t]
\centering
\definecolor{splitextremelow}{RGB}{112,126,168}
\definecolor{splitlow}{RGB}{145,168,196}
\definecolor{splitnormal}{RGB}{177,207,190}
\definecolor{splitover}{RGB}{244,194,151}
\definecolor{splitextremeover}{RGB}{250,218,126}
\definecolor{splitoverall}{RGB}{184,188,192}
\newcommand{\splitcrescent}[1]{\tikz[baseline=-0.55ex,scale=0.15]{\fill[#1] (0,0) circle (1);\fill[white] (0.43,0.20) circle (0.88);}}
\newcommand{\splithalfmoon}[1]{\tikz[baseline=-0.55ex,scale=0.15]{\begin{scope}\clip (0,0) circle (1);\fill[#1!25!white] (-1,-1) rectangle (1,1);\fill[#1] (-1,-1) rectangle (0,1);\end{scope}\draw[#1!85!black,line width=0.55pt] (0,0) circle (1);}}
\newcommand{\splitsoftday}[1]{\tikz[baseline=-0.55ex,scale=0.15]{\foreach \a in {0,90,180,270}{\draw[#1!85!black,line width=0.65pt,line cap=round] (\a:1.05)--(\a:1.28);}\fill[#1!55!white] (0,0) circle (0.78);\draw[#1!85!black,line width=0.55pt] (0,0) circle (0.78);}}
\newcommand{\splitsun}[1]{\tikz[baseline=-0.55ex,scale=0.15]{\foreach \a in {0,45,...,315}{\draw[#1!90!black,line width=0.65pt,line cap=round] (\a:0.96)--(\a:1.32);}\fill[#1] (0,0) circle (0.67);}}
\newcommand{\splitbrightsun}[1]{\tikz[baseline=-0.55ex,scale=0.15]{\foreach \a in {0,30,...,330}{\draw[#1!92!black,line width=0.70pt,line cap=round] (\a:0.98)--(\a:1.38);}\fill[#1] (0,0) circle (0.72);}}
\newsavebox{\splittextbox}
\newsavebox{\spliticonbox}
\newcommand{\splitlabel}[2]{%
  \begingroup
  \sbox{\splittextbox}{#1}%
  \sbox{\spliticonbox}{#2}%
  \dimen0=\ht\splittextbox
  \advance\dimen0 by -\dp\splittextbox
  \advance\dimen0 by -\ht\spliticonbox
  \advance\dimen0 by \dp\spliticonbox
  \divide\dimen0 by 2
  \makebox[5.3em][l]{%
    \usebox{\splittextbox}\hfill
    \makebox[1.2em][c]{\raisebox{\dimen0}{\usebox{\spliticonbox}}}%
  }%
  \endgroup
}
\caption{Suite-resolved success rates (\%) of ISP modules on LIBERO. SOG averages Spatial, Object, and Goal; Long denotes LIBERO-10; Avg averages all four suites.}
\label{tab:libero_illumination_split}
\footnotesize
\setlength{\tabcolsep}{0.4pt}
\renewcommand{\arraystretch}{0.95}
\resizebox{\textwidth}{!}{%
\begin{tabular}{@{}lcl*{18}{>{\centering\arraybackslash}p{2.65em}}@{}}
\toprule
\multirow{2}{*}{Illumin.} & & \multirow{2}{*}{Backbone}
& \multicolumn{3}{c}{Default}
& \multicolumn{3}{c}{DarkISP}
& \multicolumn{3}{c}{RAM}
& \multicolumn{3}{c}{RAWAdapter}
& \multicolumn{3}{c}{RAWild}
& \multicolumn{3}{c}{\textbf{Ours}} \\
\cmidrule(lr){4-6}\cmidrule(lr){7-9}\cmidrule(lr){10-12}\cmidrule(lr){13-15}\cmidrule(lr){16-18}\cmidrule(lr){19-21}
& & & SOG & Long & Avg
& SOG & Long & Avg
& SOG & Long & Avg
& SOG & Long & Avg
& SOG & Long & Avg
& SOG & Long & Avg \\
\midrule
\multirow{3}{*}{\splitlabel{\shortstack[l]{Extreme\\Lowlight}}{\splitcrescent{splitextremelow}}}
& \multirow{3}{*}{\textcolor{splitextremelow}{\rule{2.6pt}{2.84em}}} & PI-0
& \secondbest{7.87} & \secondbest{3.20} & \secondbest{6.70} & 1.93 & 0.00 & 1.45 & 1.80 & 0.00 & 1.35 & 2.80 & 0.00 & 2.10 & \thirdbest{2.87} & 0.00 & \thirdbest{2.15} & \firstbest{46.00} & \firstbest{24.80} & \firstbest{40.70} \\
& & PI-0.5
& 2.60 & 6.60 & 3.60 & \secondbest{5.80} & \secondbest{9.00} & \secondbest{6.60} & 1.73 & 0.00 & 1.30 & 0.07 & 0.00 & 0.05 & \thirdbest{3.93} & \thirdbest{8.00} & \thirdbest{4.95} & \firstbest{41.63} & \firstbest{30.00} & \firstbest{38.72} \\
& & Qwen3-OFT
& 0.00 & 0.00 & 0.00 & \secondbest{4.13} & \thirdbest{2.40} & \secondbest{3.70} & 0.00 & 0.00 & 0.00 & 0.00 & 0.00 & 0.00 & \thirdbest{0.87} & \secondbest{2.60} & \thirdbest{1.30} & \firstbest{15.20} & \firstbest{10.60} & \firstbest{14.05} \\
\midrule
\multirow{3}{*}{\splitlabel{Lowlight}{\splithalfmoon{splitlow}}}
& \multirow{3}{*}{\textcolor{splitlow}{\rule{2.6pt}{2.84em}}} & PI-0
& \secondbest{68.67} & \secondbest{44.60} & \secondbest{62.65} & 2.67 & 0.00 & 2.00 & 1.80 & 0.00 & 1.35 & 2.60 & 0.00 & 1.95 & \thirdbest{6.67} & \thirdbest{0.40} & \thirdbest{5.10} & \firstbest{90.47} & \firstbest{66.00} & \firstbest{84.35} \\
& & PI-0.5
& \thirdbest{62.47} & \thirdbest{50.80} & \thirdbest{59.55} & 53.60 & 43.00 & 50.95 & 1.13 & 0.00 & 0.85 & 0.20 & 0.00 & 0.15 & \secondbest{72.93} & \secondbest{60.40} & \secondbest{69.80} & \firstbest{91.67} & \firstbest{81.80} & \firstbest{89.20} \\
& & Qwen3-OFT
& 36.40 & 19.00 & 32.05 & \thirdbest{47.73} & \thirdbest{29.40} & \thirdbest{43.15} & 0.67 & 0.00 & 0.50 & 0.00 & 0.00 & 0.00 & \secondbest{74.73} & \secondbest{45.20} & \secondbest{67.35} & \firstbest{87.13} & \firstbest{64.00} & \firstbest{81.35} \\
\midrule
\multirow{3}{*}{\splitlabel{Normal}{\splitsoftday{splitnormal}}}
& \multirow{3}{*}{\textcolor{splitnormal}{\rule{2.6pt}{2.84em}}} & PI-0
& \firstbest{96.60} & \firstbest{79.60} & \firstbest{92.35} & 2.53 & 0.00 & 1.90 & 2.33 & 0.20 & 1.80 & 2.80 & 0.00 & 2.10 & \thirdbest{43.13} & \thirdbest{11.60} & \thirdbest{35.25} & \secondbest{95.00} & \secondbest{76.20} & \secondbest{90.30} \\
& & PI-0.5
& \firstbest{98.40} & \firstbest{91.80} & \firstbest{96.75} & \thirdbest{97.67} & \thirdbest{88.60} & 95.40 & 81.87 & 29.80 & 68.85 & 86.07 & 38.20 & 74.10 & \secondbest{98.20} & \secondbest{89.20} & \thirdbest{95.95} & \firstbest{98.40} & \secondbest{89.20} & \secondbest{96.10} \\
& & Qwen3-OFT
& \thirdbest{98.60} & \firstbest{92.80} & \firstbest{97.15} & \secondbest{98.80} & 91.00 & 96.85 & 69.00 & 23.00 & 57.50 & 97.13 & 84.80 & 94.05 & \firstbest{98.87} & \thirdbest{91.40} & \secondbest{97.00} & 98.53 & \secondbest{92.20} & \thirdbest{96.95} \\
\midrule
\multirow{3}{*}{\splitlabel{OverExp}{\splitsun{splitover}}}
& \multirow{3}{*}{\textcolor{splitover}{\rule{2.6pt}{2.84em}}} & PI-0
& \secondbest{66.00} & \secondbest{35.40} & \secondbest{58.35} & 1.60 & 0.00 & 1.20 & 2.53 & 0.00 & 1.90 & 2.93 & 0.00 & 2.20 & \thirdbest{37.13} & \thirdbest{6.20} & \thirdbest{29.40} & \firstbest{81.93} & \firstbest{55.40} & \firstbest{75.30} \\
& & PI-0.5
& \thirdbest{85.20} & \thirdbest{48.20} & \thirdbest{75.95} & 57.80 & 39.60 & 53.25 & 1.67 & 0.20 & 1.30 & 75.73 & 39.00 & 66.55 & \secondbest{86.67} & \firstbest{64.40} & \secondbest{81.10} & \firstbest{89.00} & \secondbest{59.00} & \firstbest{81.50} \\
& & Qwen3-OFT
& 19.27 & \thirdbest{13.60} & 17.85 & \thirdbest{29.80} & 12.80 & \thirdbest{25.55} & 4.20 & 0.20 & 3.20 & 14.80 & 4.40 & 12.20 & \secondbest{38.87} & \secondbest{17.80} & \secondbest{33.60} & \firstbest{91.67} & \firstbest{55.00} & \firstbest{82.50} \\
\midrule
\multirow{3}{*}{\splitlabel{\shortstack[l]{Extreme\\OverExp}}{\splitbrightsun{splitextremeover}}}
& \multirow{3}{*}{\textcolor{splitextremeover}{\rule{2.6pt}{2.84em}}} & PI-0
& \thirdbest{20.67} & \secondbest{19.80} & \thirdbest{20.45} & 1.60 & 0.00 & 1.20 & 2.00 & 0.00 & 1.50 & 2.73 & 0.00 & 2.05 & \secondbest{32.87} & \thirdbest{6.20} & \secondbest{26.20} & \firstbest{68.27} & \firstbest{28.20} & \firstbest{58.25} \\
& & PI-0.5
& 20.93 & 24.20 & 21.75 & 52.20 & 19.40 & 44.00 & 1.40 & 0.00 & 1.05 & \thirdbest{58.47} & \thirdbest{24.80} & \thirdbest{50.05} & \firstbest{84.93} & \firstbest{45.20} & \firstbest{75.00} & \secondbest{78.40} & \secondbest{31.40} & \secondbest{66.65} \\
& & Qwen3-OFT
& 0.00 & 0.00 & 0.00 & \secondbest{17.13} & 0.00 & \secondbest{12.85} & 0.00 & 0.00 & 0.00 & 2.33 & 0.00 & 1.75 & \thirdbest{13.07} & 0.00 & \thirdbest{9.80} & \firstbest{47.80} & \firstbest{2.00} & \firstbest{36.35} \\
\midrule
\multirow{4}{*}[-0.45ex]{Overall}
& \multirow{4}{*}{\raisebox{-0.5\height}[0pt][0pt]{\textcolor{splitoverall}{\rule{2.6pt}{4.65em}}}} & PI-0
& \secondbest{51.96} & \secondbest{36.52} & \secondbest{48.10} & 2.07 & 0.00 & 1.55 & 2.09 & 0.04 & 1.58 & 2.77 & 0.00 & 2.08 & \thirdbest{24.53} & \thirdbest{4.88} & \thirdbest{19.62} & \firstbest{76.33} & \firstbest{50.12} & \firstbest{69.78} \\
& & PI-0.5
& \thirdbest{53.92} & \thirdbest{44.32} & \thirdbest{51.52} & 53.41 & 39.92 & 50.04 & 17.56 & 6.00 & 14.67 & 44.11 & 20.40 & 38.18 & \secondbest{69.33} & \secondbest{53.44} & \secondbest{65.36} & \firstbest{79.82} & \firstbest{58.28} & \firstbest{74.44} \\
& & Qwen3-OFT
& 30.85 & 25.08 & 29.41 & \thirdbest{39.52} & \thirdbest{27.12} & \thirdbest{36.42} & 14.77 & 4.64 & 12.24 & 22.85 & 17.84 & 21.60 & \secondbest{45.28} & \secondbest{31.40} & \secondbest{41.81} & \firstbest{68.07} & \firstbest{44.76} & \firstbest{62.24} \\
\cmidrule(lr){3-21}
& & Avg
& \thirdbest{45.58} & \secondbest{35.31} & \secondbest{43.01} & 31.67 & 22.35 & 29.34 & 11.47 & 3.56 & 9.50 & 23.24 & 12.75 & 20.62 & \secondbest{46.38} & \thirdbest{29.91} & \thirdbest{42.26} & \firstbest{74.74} & \firstbest{51.05} & \firstbest{68.82} \\
\bottomrule
\end{tabular}%
}
\end{table}

\section{Experiments}

\paragraph{Implementation Details.}
We train ISP modules per frozen policy for 2,000 updates on a single NVIDIA A100 80GB GPU. To alleviate temporal inconsistency in ISP baselines designed for single images, we jointly optimize them over 8-frame trajectory windows and average their corresponding losses. \methodname{} instead unrolls $N_{\mathrm{rec}}=6$ recurrent endpoints with stride 8 for Qwen3-series and 5 for PI0 and PI0.5, and applies full BPTT. More details are provided in Appendix~\ref{app:training_details}.

\paragraph{Baseline Methods.}
We compare \methodname{} against a fixed Default ISP and four
learning-based neural ISPs. The Default ISP applies the paired deterministic
RAW-to-RGB reprocessing described in Appendix~\ref{app:unprocessing}, without
task adaptation. The learned baselines include DarkISP
\citep{guo2025darkisp}, RAM \citep{gamrian2025ram}, RAW-Adapter
\citep{cui2024rawadapter}, and RAWild \citep{liu2026rawild}. We use task
success rate as the evaluation metric and conduct 50 rollouts per task with
distinct random seeds and initial positions. Best results are shaded as
\firstbest{first}, \secondbest{second}, and \thirdbest{third}, respectively.

\paragraph{VLA Backbones.}
On the LIBERO split of \benchmarkname{}, the main ISP comparison uses
PI-0~\citep{black2024pi0}, PI-0.5~\citep{physicalintelligence2025pi05}, and
Qwen3-OFT across the four LIBERO suites and five illumination regimes;
Qwen3-PI is additionally included in the ISP perturbation analysis.
On RoboTwin~2.0, we use PI-0 and PI-0.5 under the same five illumination
regimes. The Qwen3-series VLA models are task-finetuned with StarVLA
\citep{starvla2026}.


\paragraph{Evaluation on LIBERO.}

Across the suite-resolved results in
Table~\ref{tab:libero_illumination_split}, \methodname{} achieves the highest overall
performance for every backbone, raising the cross-backbone average from
43.01\% for the strongest baseline to 68.82\%. This improvement is driven
primarily by adverse illumination, with particularly large gains for Qwen3-OFT
under overexposure, while the strong performance under normal lighting is
largely preserved. Moreover, the gains hold for both the Spatial, Object, and
Goal suites and the long-horizon LIBERO-10 suite, demonstrating that \methodname{}
benefits different task types and horizons rather than a particular benchmark
subset. The first row of Figure~\ref{fig:experiments} qualitatively compares
the agent-view outputs of the Default ISP and \methodname{} on LIBERO.

\paragraph{Evaluation on RoboTwin~2.0.}

\begin{table}[!tp]
\centering
\definecolor{rtillumextremelow}{RGB}{112,126,168}
\definecolor{rtillumlow}{RGB}{145,168,196}
\definecolor{rtillumnormal}{RGB}{177,207,190}
\definecolor{rtillumover}{RGB}{244,194,151}
\definecolor{rtillumextremeover}{RGB}{250,218,126}
\definecolor{rtillumoverall}{RGB}{184,188,192}
\newcommand{\rtminiicon}[2]{%
  \tikz[baseline=-0.55ex,scale=0.10]{%
    \ifcase#1\relax
      \fill[#2] (0,0) circle (1);\fill[white] (0.43,0.20) circle (0.88);%
    \or
      \begin{scope}\clip (0,0) circle (1);\fill[#2!25!white] (-1,-1) rectangle (1,1);\fill[#2] (-1,-1) rectangle (0,1);\end{scope}\draw[#2!85!black,line width=0.55pt] (0,0) circle (1);%
    \or
      \foreach \a in {0,90,180,270}{\draw[#2!85!black,line width=0.65pt,line cap=round] (\a:1.05)--(\a:1.28);}\fill[#2!55!white] (0,0) circle (0.78);\draw[#2!85!black,line width=0.55pt] (0,0) circle (0.78);%
    \or
      \foreach \a in {0,45,...,315}{\draw[#2!90!black,line width=0.65pt,line cap=round] (\a:0.96)--(\a:1.32);}\fill[#2] (0,0) circle (0.67);%
    \or
      \foreach \a in {0,30,...,330}{\draw[#2!92!black,line width=0.70pt,line cap=round] (\a:0.98)--(\a:1.38);}\fill[#2] (0,0) circle (0.72);%
    \fi
  }%
}
\newsavebox{\rtminitextbox}
\newsavebox{\rtminiiconbox}
\newcommand{\rtminilabel}[2]{%
  \begingroup
  \sbox{\rtminitextbox}{#1}%
  \sbox{\rtminiiconbox}{#2}%
  \dimen0=\ht\rtminitextbox
  \advance\dimen0 by -\dp\rtminitextbox
  \advance\dimen0 by -\ht\rtminiiconbox
  \advance\dimen0 by \dp\rtminiiconbox
  \divide\dimen0 by 2
  \makebox[4.9em][l]{%
    \usebox{\rtminitextbox}\hfill
    \makebox[1.0em][c]{\raisebox{\dimen0}{\usebox{\rtminiiconbox}}}%
  }%
  \endgroup
}

\begin{minipage}[!tp]{0.555\textwidth}
\centering
\caption{Success rates (\%) on RoboTwin~2.0 across five illumination levels. \textit{Default} denotes the default SAPIEN tone-mapped RGB baseline.}
\label{tab:robotwin_illumination_baselines}
\scriptsize
\setlength{\tabcolsep}{0.5pt}
\setlength{\heavyrulewidth}{0.6pt}
\renewcommand{\arraystretch}{1.05}
\begin{tabularx}{\linewidth}{@{}l@{\hspace{1pt}}c@{\hspace{2pt}}l*{6}{>{\centering\arraybackslash}X}@{}}
\toprule
Illumin. & & Backbone & Default & DarkISP & RAM & R.Adp. & RAWild & \textbf{Ours} \\
\midrule
\multirow{2}{*}{\rtminilabel{\shortstack[l]{Extreme\\Lowlight}}{\rtminiicon{0}{rtillumextremelow}}}
& \multirow{2}{*}{\textcolor{rtillumextremelow}{\rule{2.2pt}{2.05em}}} & PI-0 & \secondbest{23.62} & 4.72 & \thirdbest{12.60} & 1.57 & 8.66 & \firstbest{32.28} \\
& & PI-0.5 & \secondbest{17.32} & 1.57 & 3.15 & 0.00 & \thirdbest{5.51} & \firstbest{25.98} \\
\midrule
\multirow{2}{*}{\rtminilabel{Lowlight}{\rtminiicon{1}{rtillumlow}}}
& \multirow{2}{*}{\textcolor{rtillumlow}{\rule{2.2pt}{2.05em}}} & PI-0 & \secondbest{40.94} & 0.79 & \thirdbest{10.24} & 3.15 & 6.30 & \firstbest{59.06} \\
& & PI-0.5 & \secondbest{51.18} & 0.00 & 2.36 & 0.00 & \thirdbest{4.72} & \firstbest{63.78} \\
\midrule
\multirow{2}{*}{\rtminilabel{Normal}{\rtminiicon{2}{rtillumnormal}}}
& \multirow{2}{*}{\textcolor{rtillumnormal}{\rule{2.2pt}{2.05em}}} & PI-0 & \thirdbest{55.91} & 34.65 & 17.32 & \secondbest{56.69} & 28.35 & \firstbest{57.48} \\
& & PI-0.5 & \firstbest{70.87} & 35.43 & 20.47 & \thirdbest{41.73} & 40.94 & \secondbest{67.72} \\
\midrule
\multirow{2}{*}{\rtminilabel{OverExp}{\rtminiicon{3}{rtillumover}}}
& \multirow{2}{*}{\textcolor{rtillumover}{\rule{2.2pt}{2.05em}}} & PI-0 & 48.41 & 34.92 & 19.05 & \thirdbest{51.59} & \secondbest{55.56} & \firstbest{58.73} \\
& & PI-0.5 & \secondbest{56.35} & 26.98 & 4.76 & 38.89 & \thirdbest{45.24} & \firstbest{65.08} \\
\midrule
\multirow{2}{*}{\rtminilabel{\shortstack[l]{Extreme\\OverExp}}{\rtminiicon{4}{rtillumextremeover}}}
& \multirow{2}{*}{\textcolor{rtillumextremeover}{\rule{2.2pt}{2.05em}}} & PI-0 & 36.00 & \thirdbest{36.80} & 11.20 & 15.20 & \secondbest{47.20} & \firstbest{56.80} \\
& & PI-0.5 & 38.40 & \thirdbest{42.40} & 1.60 & 20.00 & \secondbest{54.40} & \firstbest{63.20} \\
\midrule
\multirow{3}{*}{Overall}
& \multirow{3}{*}{\raisebox{-0.5\height}[0pt][0pt]{\textcolor{rtillumoverall}{\rule{2.2pt}{4.05em}}}} & PI-0 & \secondbest{40.98} & 22.31 & 14.08 & 25.63 & \thirdbest{29.11} & \firstbest{52.85} \\
& & PI-0.5 & \secondbest{46.84} & 21.20 & 6.49 & 20.09 & \thirdbest{30.06} & \firstbest{57.12} \\
\cmidrule(lr){3-9}
& & Avg & \secondbest{43.91} & 21.76 & 10.28 & 22.86 & \thirdbest{29.59} & \firstbest{54.98} \\
\bottomrule
\end{tabularx}
\end{minipage}
\hfill
\begin{minipage}[!tp]{0.415\textwidth}
\centering
\caption{Real-world success rates (\%) on four tasks using PI-0.5. \textit{Default} denotes the constant default ISP baseline.}
\label{tab:realworld_illumination}
\scriptsize
\setlength{\tabcolsep}{1.0pt}
\setlength{\heavyrulewidth}{0.6pt}
\renewcommand{\arraystretch}{1.05}
\begin{tabularx}{\linewidth}{@{}>{\raggedright\arraybackslash}p{3.7em}>{\raggedright\arraybackslash}p{5.7em}*{3}{>{\centering\arraybackslash}X}@{}}
\toprule
Illumin. & Task & Default & DarkISP & \textbf{Ours} \\
\midrule
\multirow{5}{*}{Normal}
& \textit{pick \& place}  & \firstbest{96.00} & \secondbest{66.00} & \firstbest{96.00} \\
& \textit{drug drawer}  & \secondbest{92.00} & \thirdbest{52.00} & \firstbest{94.00} \\
& \textit{pick flower}  & \firstbest{72.00} & \thirdbest{44.00} & \secondbest{66.00} \\
& \textit{stack \& cover}  & \firstbest{46.00} & \thirdbest{38.00} & \secondbest{44.00} \\
& Avg & \firstbest{76.50} & \thirdbest{50.00} & \secondbest{75.00} \\
\midrule
\multirow{5}{*}{Lowlight}
& \textit{pick \& place}  & \thirdbest{0.00} & \secondbest{22.00} & \firstbest{92.00} \\
& \textit{drug drawer}  & \thirdbest{0.00} & \secondbest{12.00} & \firstbest{88.00} \\
& \textit{pick flower}  & \thirdbest{0.00} & \secondbest{10.00} & \firstbest{56.00} \\
& \textit{stack \& cover}  & \secondbest{0.00} & \secondbest{0.00} & \firstbest{30.00} \\
& Avg & \thirdbest{0.00} & \secondbest{11.00} & \firstbest{66.50} \\
\bottomrule
\end{tabularx}
\vspace{0em}

\caption{Inference frame-per-second.}
\label{tab:realworld_fps}
\setlength{\tabcolsep}{1.0pt}
\renewcommand{\arraystretch}{1.05}
\begin{tabularx}{\linewidth}{@{}*{5}{>{\centering\arraybackslash}X}@{}}
\toprule
DarkISP & RAM & R.Adp. & RAWild & \textbf{Ours} \\
\midrule
176 & \firstbest{972} & \secondbest{414} & \thirdbest{320} & 168 \\
\bottomrule
\end{tabularx}
\end{minipage}
\end{table}

As shown in Table~\ref{tab:robotwin_illumination_baselines}, \methodname{} achieves a
cross-backbone average of 54.98\% on RoboTwin~2.0, improving the success rate
by 85.81\% relative to the strongest neural ISP baseline. \methodname{} also ranks
first in every adverse illumination regime for both policies. Its largest
gains occur under low light, while the improvements under overexposure confirm
its effectiveness for both signal-starved and saturated observations.
The second row of Figure~\ref{fig:experiments} compares the corresponding
agent-view outputs of the Default ISP and \methodname{} on RoboTwin~2.0.

\paragraph{Evaluation on Real-World Dual-Arm.}

We evaluate \methodname{} on four real-world dual-arm tasks under normal and low-light conditions with a fine-tuned PI-0.5 policy backbone. As shown in Table~\ref{tab:realworld_illumination}, \methodname{} achieves 75.00\% average success under normal illumination, comparable to the Default ISP at 76.50\% and above DarkISP at 50.00\%. Under low light, \methodname{} maintains 66.50\% success, while the Default ISP and DarkISP reach only 0.00\% and 11.00\%, respectively. As shown in Table~\ref{tab:realworld_fps}, all evaluated ISP modules support real-time operation and run faster than the cameras' native frame rate, with \methodname{} reaching 168 FPS. The final row of Figure~\ref{fig:experiments} shows a representative real-world task completion sequence. Further implementation details and task execution visualizations are provided in Appendix~\ref{app:real_world_experiments}.

\section{Ablation}

\definecolor{ablationgroupgray}{RGB}{240,240,240}

\begin{table}[!tp]
\centering
\caption{Ablations of \methodname{} on LIBERO with Qwen3-OFT and RoboTwin~2.0 with PI-0.5. The two blocks isolate photometric descriptors and recurrent module structures.}
\label{tab:ablation_core_main_full}
\scriptsize
\setlength{\tabcolsep}{2.1pt}
\renewcommand{\arraystretch}{1.05}
\resizebox{\textwidth}{!}{%
\begin{tabular}{@{}l*{6}{c}@{\hspace{7pt}}*{6}{c}@{}}
\toprule
& \multicolumn{6}{c}{LIBERO / Qwen3-OFT}
& \multicolumn{6}{c}{RoboTwin~2.0 / PI-0.5} \\
\cmidrule(lr){2-7}\cmidrule(lr){8-13}
Variant & Ext.Low & Low & Normal & Over & Ext.Over & Avg.
& Ext.Low & Low & Normal & Over & Ext.Over & Avg. \\
\midrule
\rowcolor{ablationgroupgray}
\multicolumn{13}{@{}l}{\textit{Photometric descriptors}} \\
w/o spatial feature $g_t$        & 0.00 & 20.35 & \secondbest{97.10} & 51.15 & 13.15 & 36.35 & 14.96 & 12.60 & \thirdbest{62.99} & 60.32 & 59.20 & 41.93 \\
w/o luminance descriptor $h_t^L$ & \secondbest{27.75} & \firstbest{95.05} & 96.80 & 50.80 & 25.95 & 59.27 & 21.26 & 47.24 & 50.39 & 56.35 & 52.80 & 45.57 \\
w/o chroma descriptor $h_t^C$    & 0.00 & 0.00 & 96.65 & 48.75 & 16.60 & 32.40 & \thirdbest{23.62} & \thirdbest{57.48} & 48.03 & 60.32 & 53.60 & 48.58 \\
\midrule
\rowcolor{ablationgroupgray}
\multicolumn{13}{@{}l}{\textit{Recurrent estimator}} \\
w/o hidden state $S_{t-1}^{L/C}$ & 0.00 & 0.35 & \firstbest{97.20} & \thirdbest{51.65} & 22.85 & 34.41 & 22.05 & \thirdbest{57.48} & 46.46 & 57.94 & 53.60 & 47.47 \\
w/o $\Theta_{t-1}$ conditioning  & 0.00 & 0.00 & 96.65 & 49.95 & 25.25 & 34.37 & \secondbest{24.41} & 49.61 & 51.18 & 57.94 & 55.20 & 47.63 \\
Single shared $\mathrm{GRU}_{L,C}$ & \firstbest{28.10} & 91.20 & 96.75 & \secondbest{55.55} & \secondbest{35.30} & \secondbest{61.38} & 22.83 & 55.91 & 40.94 & \thirdbest{63.49} & 56.80 & 47.94 \\
Lumin. only ($S_t=S_t^L$)        & \firstbest{28.10} & \secondbest{94.60} & 96.35 & 51.40 & \thirdbest{29.75} & \thirdbest{60.04} & 21.26 & \secondbest{61.42} & \secondbest{66.14} & \firstbest{67.46} & \thirdbest{62.40} & \secondbest{55.70} \\
Chrom. only ($S_t=S_t^C$)        & \firstbest{28.10} & \thirdbest{93.45} & 96.90 & 50.35 & 25.15 & 58.79 & 22.83 & 51.97 & 58.27 & 61.11 & \firstbest{64.00} & \thirdbest{51.58} \\
\midrule
Full \methodname{}                      & \thirdbest{14.05} & 81.35 & \thirdbest{96.95} & \firstbest{82.50} & \firstbest{36.35} & \firstbest{62.24} & \firstbest{25.98} & \firstbest{63.78} & \firstbest{67.72} & \secondbest{65.08} & \secondbest{63.20} & \firstbest{57.12} \\
\bottomrule
\end{tabular}%
}
\end{table}

Table~\ref{tab:ablation_core_main_full} summarizes controlled ablations on two representative VLA models, Qwen3-OFT with a regression head on LIBERO and PI-0.5 with a diffusion head on RoboTwin~2.0.

\paragraph{Photometric Descriptors.}
The complete descriptor set achieves average success rates of 62.24\% on
LIBERO and 57.12\% on RoboTwin~2.0. Removing the spatial feature $g_t$ reduces these averages to 36.35\% and 41.93\%, respectively, confirming the importance of spatial luminance structure. The global luminance descriptor $h_t^L$ is especially important on RoboTwin~2.0, whereas removing the chroma descriptor $h_t^C$ causes the largest drop on LIBERO. These complementary trends support separate descriptors for intensity distribution, spatial structure, and color.

\paragraph{Recurrent Estimator.}
Removing either the hidden state or previous-parameter conditioning sharply degrades LIBERO performance and consistently lowers success on RoboTwin~2.0. A shared GRU also underperforms the decoupled design, particularly on RoboTwin~2.0, while either the luminance-only or chroma-only state remains inferior to their combination. Together, these results show that temporal context, parameter feedback, and factorized luminance--chroma estimation each contribute to robust adaptation. Additional ablations of the training losses, individual ISP operators, and burst-denoising settings are provided in Appendix~\ref{app:additional_ablations}.

\section{Related Works}


\paragraph{VLA \& WAM Visual Robustness.}
Generalist policies span token-based and continuous-action VLAs
\citep{kim2024openvla,black2024pi0,physicalintelligence2025pi05} and
video-predictive WAMs \citep{kim2025cosmospolicy,ye2026dreamzero}, increasingly targeting cross-embodiment control and physical reasoning
\citep{bjorck2025gr00tn1,geminirobotics2025}. Yet both remain vulnerable to visual changes. Robustness benchmarks use perturbations and scalable simulation to cover appearance, lighting, viewpoint, scene composition, and physical shifts \citep{pumacay2024colosseum,li2025simpler,wang2025vlatest,wang2024ladev,chen2025robotwin,zhang2026wamrobustness,fei2025liberoplus,morgan2026colosseumv2,chen2026radar}. Existing methods edit task-irrelevant regions, enforce action consistency, restore observations, or learn invariance to visual and illumination changes \citep{hancock2024byovla,guo2025multirobustvla,zhang2025robustvla,orjuela2026crt,xie2026strongvla,luo2026rovla,watanabe2026flare}. Additional robustness can come from event, thermal, or tactile sensing, at the cost of extra sensors and cross-modal data \citep{zhai2026evla,liu2026eventvla,yu2026safenightvla,huang2025tactilevla}. We instead operate on the existing camera signal and study robustness at the RAW and ISP processing levels.

\paragraph{Task-Oriented and Neural ISP.}
Conventional ISPs target human visual quality and may not preserve signals most useful to downstream perception. Prior work therefore co-optimizes RAW processing and recognition, builds compact machine-oriented pipelines, and searches task-specific ISP structures
\citep{diamond2021dirtypixels,wu2019visionisp,yu2021reconfigisp,shi2022refactoring}. Later methods introduce content-adaptive operators and connect learnable ISP stages to downstream models for sensor- or task-conditioned processing \citep{sun2024rlseqisp,wang2024adaptiveisp,gamrian2025ram,won2026posisp,morawski2022genisp,cui2024rawadapter,huang2025drraw}. Complementary work improves adverse-condition and sensor-general RAW perception through synthesis, augmentation, enhancement, adaptation, and efficient models \citep{punnappurath2022daytonight,yoshimura2023rawgment,guo2025darkisp,hashmi2025torchadapt,chen2026taisp,liu2026rawild,li2026uniisp}.
Our work extends this direction to sequential VLA behavior and manipulation success.

\section{Conclusion}

We introduced \methodname, a lightweight neural ISP for frozen VLA policies, and \benchmarkname, a RAW-domain manipulation benchmark. Across simulation and real-world tasks, \methodname{} maintains nominal-condition performance while substantially improving robustness to adverse illumination, demonstrating that adaptive camera processing supports robust robot control.

\appendix
\clearpage
\section{RGB Unprocessing and Default ISP}
\label{app:unprocessing}

This section defines the RAW representation used throughout the paper and specifies the paired RGB-to-RAW unprocessing and default RAW-to-RGB ISP.

\paragraph{RAW Input Representation.}
We distinguish the sensor measurement from the representation processed by
\methodname{}. A physical camera first records a single-channel, mosaiced Bayer
RAW measurement. We subtract the sensor black level, normalize by the usable
white level, and demosaic the Bayer array into a three-channel linear signal.
Throughout the paper, we refer to this black-level-corrected, demosaiced signal
as \emph{linear RAW}. It is the direct input representation of \methodname{}.
The simulated observations are three-channel \emph{pseudo-RAW} signals created
directly in the same linear RAW representation and therefore do not contain a
Bayer mosaic.

We construct the simulated pseudo-RAW observations using a fixed unprocessing pipeline adapted from \citet{brooks2019unprocessing}. Let
$Y^{\mathrm{render}}\in[0,1]^{3\times H\times W}$ denote the display-referred RGB
observation resized to the VLA input resolution. All photometric operations are
applied channel-wise. The default smoothstep tone curve is
\begin{equation}
    T_0(z)=3z^2-2z^3,
\end{equation}
whose inverse $T_0^{-1}$ is uniquely defined on $[0,1]$, and the
standard sRGB transfer function
\begin{equation}
    \mathcal{G}(z)=
    \begin{cases}
        12.92z, & z\leq 0.0031308,\\
        1.055z^{1/2.4}-0.055, & z>0.0031308.
    \end{cases}
\end{equation}
Its inverse is
\begin{equation}
    \mathcal{G}^{-1}(z)=
    \begin{cases}
        z/12.92, & z\leq 0.04045,\\
        \left((z+0.055)/1.055\right)^{2.4}, & z>0.04045.
    \end{cases}
\end{equation}

We first invert the default tone curve and sRGB transfer function, then remove
the fixed channel-gain vector $\mathbf m_{\mathrm{rgb}}=(1.8,1.0,1.7)$. The unprocessing operator
$\mathcal{U}$ produces the simulated camera RAW observation
\begin{equation}
    R
    =\mathcal{U}(Y^{\mathrm{render}})
    =\operatorname{clip}\!\left(
        \operatorname{diag}(\mathbf m_{\mathrm{rgb}})^{-1}
        \mathcal{G}^{-1}\!\left(T_0^{-1}(Y^{\mathrm{render}})\right),0,1\right).
\end{equation}
The resulting $R$ retains the precision available from the corresponding
renderer. The complete unprocessing sequence applies inverse tone mapping,
inverse sRGB transfer, inverse channel gains, and clipping. Controlled
bit-depth changes use $\mathcal{Q}_n$ from
Equation~\ref{eq:bit_depth_perturbation}, with their environment-specific
placement given in Appendix~\ref{app:perturbation_settings}.

Let $\mathcal{P}(\cdot,\xi)$ denote the parameterized ISP and $\xi_0$ its
default parameter setting. The paired default operator
$\mathcal{P}_0(\cdot)\equiv\mathcal{P}(\cdot,\xi_0)$ reverses the fixed
photometric transformations
\begin{equation}
    Y_0
    =\mathcal{P}_0(R)
    =T_0\!\left(
        \mathcal{G}\!\left(
        \operatorname{clip}(\operatorname{diag}(\mathbf m_{\mathrm{rgb}})R,0,1)
        \right)\right)
    =Y^{\mathrm{render}}.
\end{equation}
This produces a three-channel pseudo-RAW signal in the linear RAW
representation rather than a camera-specific Bayer measurement. The baseline unprocessing does not simulate
mosaicing, sensor noise, or a camera-specific color-correction matrix. The
noise axis in Appendix~\ref{app:perturbation_settings} introduces its sensor
model separately.

\section{\benchmarkname{} Construction}
\label{app:rawvla_bench}

\benchmarkname{} changes illumination inside each simulator before RAW
formation rather than multiplying the final RGB image. It contains paired
training caches and frozen evaluation manifests for LIBERO~\citep{liu2023libero}
and RoboTwin~2.0~\citep{chen2025robotwin}.
Each training trajectory stores one target-light RAW sequence together with a
normal-light RGB reference under the same actions and physical states. To avoid
increasing the training cache fivefold, each successful trajectory is assigned
one illumination regime. During evaluation, the same task initialization is
instead expanded across all five regimes for paired comparison. In total, the
training cache contains 2,403 paired trajectories and 408,075 frames, while the
evaluation manifests specify 13,250 rollouts.
Section~\ref{app:bench_libero} describes the LIBERO~\citep{liu2023libero}
construction, and Section~\ref{app:bench_robotwin} details the
RoboTwin~2.0~\citep{chen2025robotwin} construction.

\begin{figure}[t]
\centering
\includegraphics[width=\textwidth]{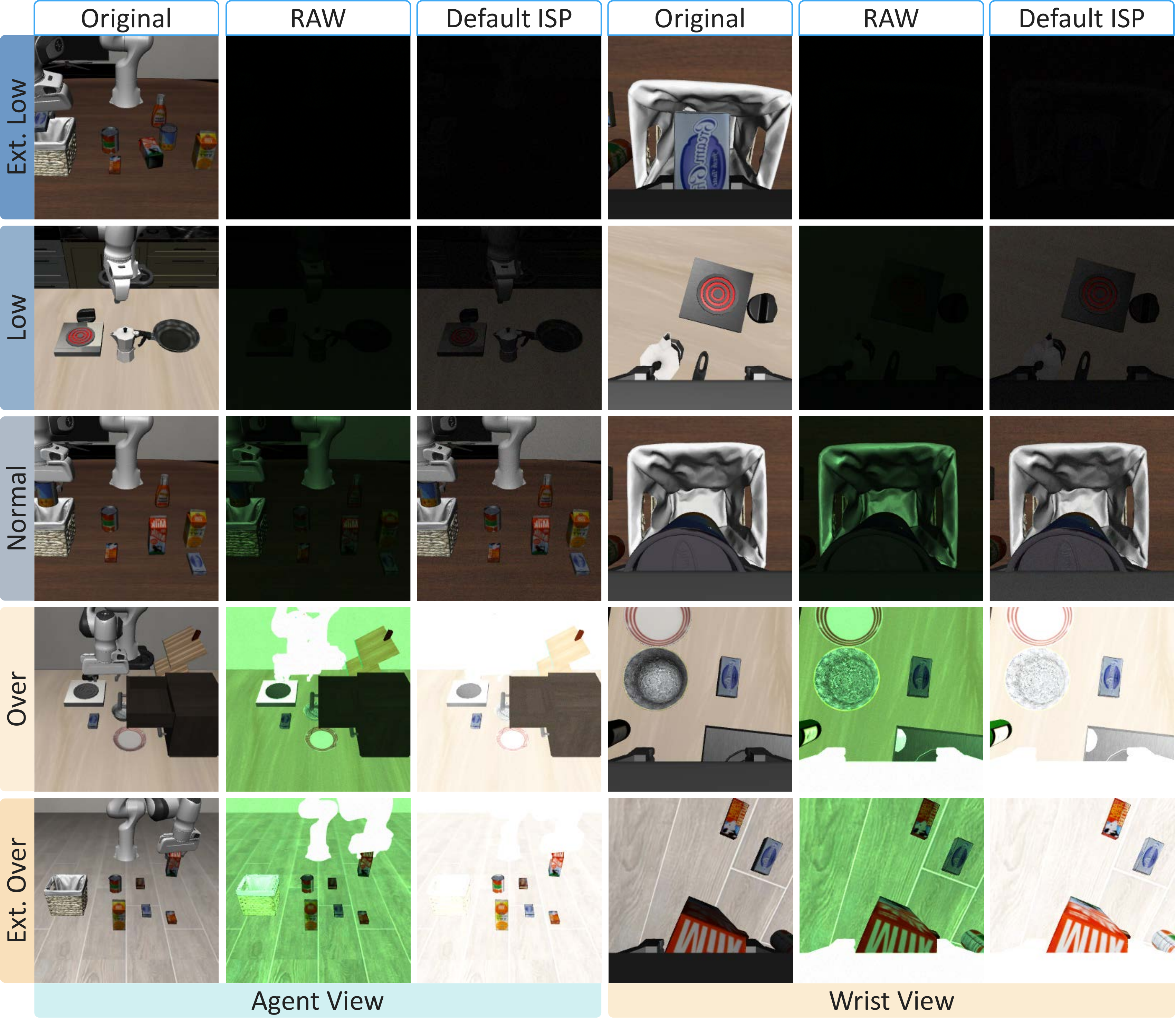}
\caption{\benchmarkname{} observations on LIBERO~\citep{liu2023libero} across five ambient-light
intensity levels ranging from \textit{extreme low light} and \textit{low light}
to \textit{normal}, \textit{overexposure}, and \textit{extreme overexposure}.
Each level shows the original RGB reference, synthesized RAW
observation, and its default-ISP rendering. Agent and wrist views use the same
unprocessing procedure.}
\label{fig:libero_unprocessing}
\end{figure}

\subsection{LIBERO Construction}
\label{app:bench_libero}

We use the four standard suites LIBERO-Spatial, LIBERO-Object, LIBERO-Goal,
and LIBERO-10 from LIBERO~\citep{liu2023libero}, comprising 40 tasks. From at most 50 official demonstrations per
task, we obtain 2,000 candidate trajectories. Each demonstration is recreated
using its recorded MuJoCo XML, initial simulator state, and seven-dimensional
action sequence. We replay the actions in a clean environment, remove no-op
actions, and retain a trajectory only when the task-specific success predicate
is reached. The same state is copied before every rendering step to a second
environment whose illumination is modified, ensuring that the RGB reference
and RAW observation differ only in the image-formation branch. This procedure
retains 1,771 successful trajectories and 261,131 frames. These include 454 from
LIBERO-Spatial, 462 from LIBERO-Object, 456 from LIBERO-Goal, and 399 from
LIBERO-10~\citep{liu2023libero}. Both the agent and wrist cameras render at $256\times256$.

\begin{figure}[t]
\centering
\includegraphics[width=\textwidth]{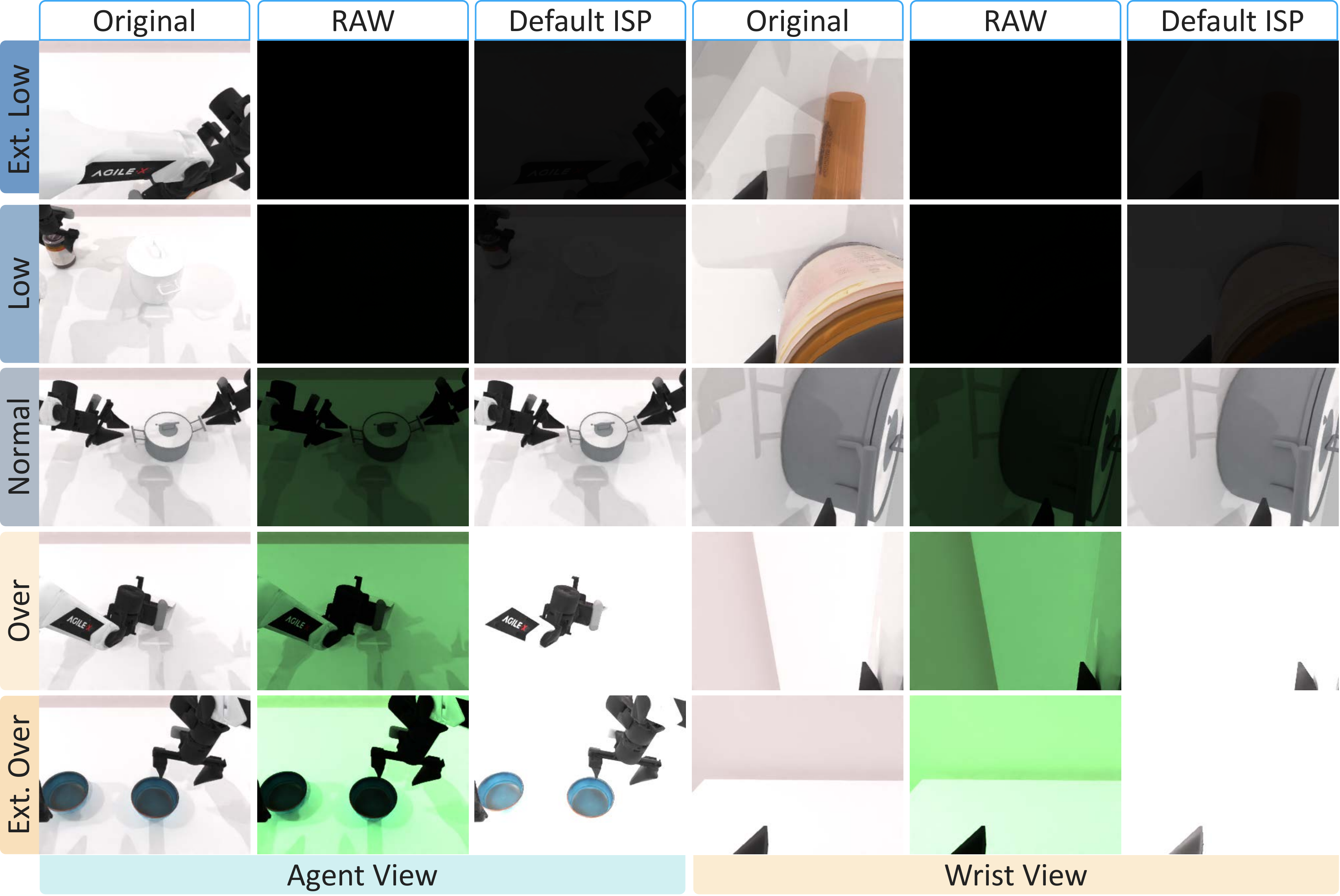}
\caption{\benchmarkname{} observations on RoboTwin~2.0~\citep{chen2025robotwin} across five ambient-light
intensity levels ranging from \textit{extreme low light} and \textit{low light}
to \textit{normal}, \textit{overexposure}, and \textit{extreme overexposure}.
Each level shows the original RGB reference, synthesized RAW
observation, and its default-ISP rendering. Agent and wrist views use the same
unprocessing procedure.}
\label{fig:robotwin_unprocessing}
\end{figure}

\begin{figure}[t]
\centering
\includegraphics[width=\textwidth]{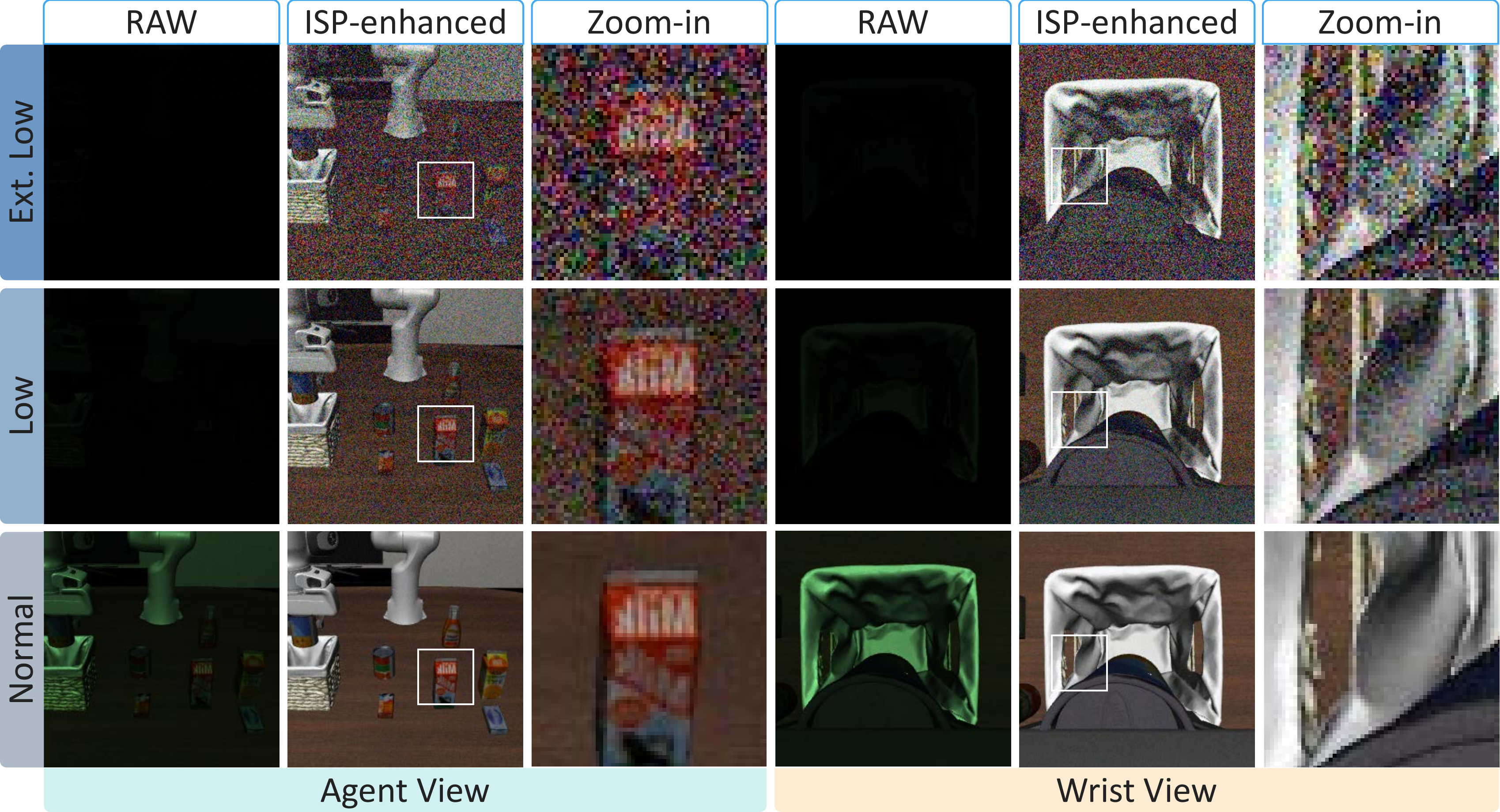}
\caption{Synthetic sensor noise on LIBERO~\citep{liu2023libero}. The figure shows noisy RAW
observations and their corresponding RGB renderings after ISP gain
amplification and subsequent processing across progressively lower illumination
levels. Noise becomes increasingly pronounced as the illumination decreases.}
\label{fig:libero_noise_synthesis}
\end{figure}

MuJoCo provides only tone-mapped RGB8 for these environments. We therefore
construct a three-channel pseudo-RAW signal by inverting the smoothstep tone
curve and sRGB transfer function, removing fixed RGB gains, clipping, and
quantizing to 10 bits. Overexposed regimes additionally apply controlled
full-well saturation before clipping. We then inject heteroscedastic Gaussian
noise with variance $4\times10^{-4}x+10^{-5}$ using deterministic seeds that
vary across trajectories, frames, and views. Figure~\ref{fig:libero_noise_synthesis}
visualizes how the synthesized RAW noise becomes increasingly visible after ISP
gain amplification as illumination decreases. The five regimes and EV intervals
are ExtremeLow $[-5.5,-4]$, Low $[-4,-2.5]$, Normal $[-0.5,0.5]$, Over
$[1.5,3]$, and ExtremeOver $[6,8]$. Over and ExtremeOver also activate
additional table-directed illumination and saturation pressure. These
pseudo-RAW observations model the processing degrees of freedom available to
the learned ISP, but are not Bayer measurements and do not contain additional
renderer highlight information beyond the source RGB8.

The frozen LIBERO~\citep{liu2023libero} evaluation manifest contains 50 initial states for each of
the 40 tasks. Every initial state is evaluated in all five illumination
regimes, producing
$4\times10\times50\times5=10{,}000$ rollouts, or 2,000 rollouts per regime.
Entries sharing an initial state use the same task, language instruction, and
camera configuration, while only illumination and sensor-noise seeds differ.

\subsection{RoboTwin~2.0 Construction}
\label{app:bench_robotwin}

We use 13 tasks from RoboTwin~2.0~\citep{chen2025robotwin} with the ALOHA-AgileX embodiment and the
clean task configuration. The 650 candidates comprise 50 recorded episodes per
task. We reconstruct each scene from its dataset seed and replay the stored
left- and right-arm joint paths through the original controller and physics
stack. An episode is retained only when both trajectory planning and the
task-specific success check succeed. Of 650 candidates, 633 pass this clean
replay audit. One trajectory then fails during paired regeneration because its
saved right-arm path is exhausted, leaving 632 complete pairs and 146,944
frames.

To prevent contact or numerical differences from causing paired trajectories
to diverge, only the clean environment executes actions. Before every camera
capture, actor poses and velocities, articulation root states, and joint
positions and velocities are copied to a target-light environment, which is
used only for rendering. Each frame includes the head, left-wrist, and
right-wrist views at $240\times320$. Unlike LIBERO~\citep{liu2023libero}, SAPIEN exposes a
pre-tonemap HDR buffer. We normalize this linear radiance by a fixed white
level, remove fixed RGB gains, and add heteroscedastic noise with variance
$2.5\times10^{-5}x+3.90625\times10^{-8}$. The resulting three-channel pseudo-RAW is
stored as \textit{float32} without 10-bit quantization.

RoboTwin~2.0~\citep{chen2025robotwin} uses environment-only illumination scaling, applied from the
captured baseline to ambient, directional, and point lights. Its five EV
intervals are ExtremeLow $[-9,-8.5]$, Low $[-7,-6]$, Normal $[-0.5,0.5]$,
Over $[1,1.5]$, and ExtremeOver $[2,2.5]$. The environment-specific ranges are
calibrated separately because MuJoCo and SAPIEN differ in indirect lighting,
materials, HDR rendering, and tone mapping. The frozen evaluation manifest
contains 50 dataset seeds for each task and expands each seed across all five
regimes, yielding $13\times50\times5=3{,}250$ rollouts, or 650 per regime.
Figures~\ref{fig:libero_unprocessing} and~\ref{fig:robotwin_unprocessing}
show the environment-specific pipelines.

\section{\methodname{} Architecture}
\label{app:photometric_conditioning}

This section expands the components of \methodname{}. We describe burst
denoising in Section~\ref{app:burst_denoising}, the luminance and chroma
descriptors in Sections~\ref{app:luminance_descriptor}
and~\ref{app:chroma_descriptor}, and the structured ISP parameterization in
Section~\ref{app:isp_parameterization}.

\subsection{Implementation Details}
\label{app:rawvla_implementation}

Our implementation receives a six-frame causal linear-RAW burst and contains
886,966 trainable parameters. The FFT burst merge has no trainable parameters. The spatial
luminance encoder consists of six $3\times3$ convolutional blocks with channel
transitions $3\!\to\!32$, $32\!\to\!32$, $32\!\to\!64$,
$64\!\to\!64$, $64\!\to\!128$, and $128\!\to\!128$. Their strides are
$1,1,2,1,2,1$, respectively, and each block uses single-group normalization
and a SiLU activation. The equal-channel luminance input is replicated across
three channels before entering this encoder. Spatial mean, standard deviation,
and learned attention-weighted mean pooling produce a 384-dimensional feature,
which is encoded by a $384\!\to\!256\!\to\!128$ MLP. The attention score is
predicted by a $1\times1$ convolution from 128 channels to one channel.

The 135-dimensional luminance descriptor is encoded by a
$135\!\to\!128\!\to\!64$ MLP. We set the chroma histogram count to
$B_C=64$, giving a $5B_C+10=330$ dimensional chroma descriptor that is encoded
by a $330\!\to\!128\!\to\!64$ MLP. The two log-chrominance ratios are clipped
to $[-4,4]$ before histogramming. Previous luminance and chroma operating
points contain nine and eight values and are encoded by
$9\!\to\!64\!\to\!32$ and $8\!\to\!64\!\to\!32$ MLPs. The luminance fusion
network maps $224\!\to\!256\!\to\!128$, while the chroma fusion network maps
$96\!\to\!128\!\to\!128$. All MLP hidden layers use SiLU activations.

The recurrent estimator contains separate luminance and chroma GRU cells.
Each cell has a 128-dimensional input and a 64-dimensional hidden state,
yielding a 128-dimensional combined recurrent state. The exposure head maps
$256\!\to\!128\!\to\!1$, the tone head maps
$256\!\to\!128\!\to\!7$, and the chroma head maps
$256\!\to\!128\!\to\!8$. The two single-layer GRU states are concatenated
and carried to the next policy observation.
\methodname{} has no U-Net-style image decoder or spatial residual branch.
The network predicts only explicit ISP controls, and the RGB output is
generated analytically by burst fusion, exposure, white balance,
color correction, and a shared monotonic Bernstein tone curve. The tone head
predicts seven relative logits and appends one fixed-zero reference logit,
producing eight monotonic intervals. Local color and local tone residuals are
disabled.

We use $E_{\max}=6$ EV, which bounds the achromatic exposure gain to
$[2^{-6},2^6]$. White balance is represented by two independent coordinates
that produce zero-sum channel offsets, with each residual bounded to
$[-1,1]$ EV. The chroma head predicts the six off-diagonal entries of the
color-correction residual as $0.25\tanh(\cdot)$. In the default
parameterization, the diagonal residuals are zero, so the diagonal of the
color-correction matrix remains one. The optional gray-preserving variant sets
each diagonal residual to the negative sum of the off-diagonal residuals in
its row, enforcing unit row sums. Neither variant applies determinant,
orthogonality, non-negativity, or additional projection constraints.
For burst processing, we use patch size 32, stride 16, a two-dimensional Hann
window, reliability scale $\kappa=1.8$, and temporal weights
$(0.05,0.075,0.125,0.1875,0.375)$ for the five historical frames of the
six-frame causal burst.

\subsection{Burst Denoising}
\label{app:burst_denoising}

\begin{figure}[t]
\centering
\includegraphics[width=\textwidth]{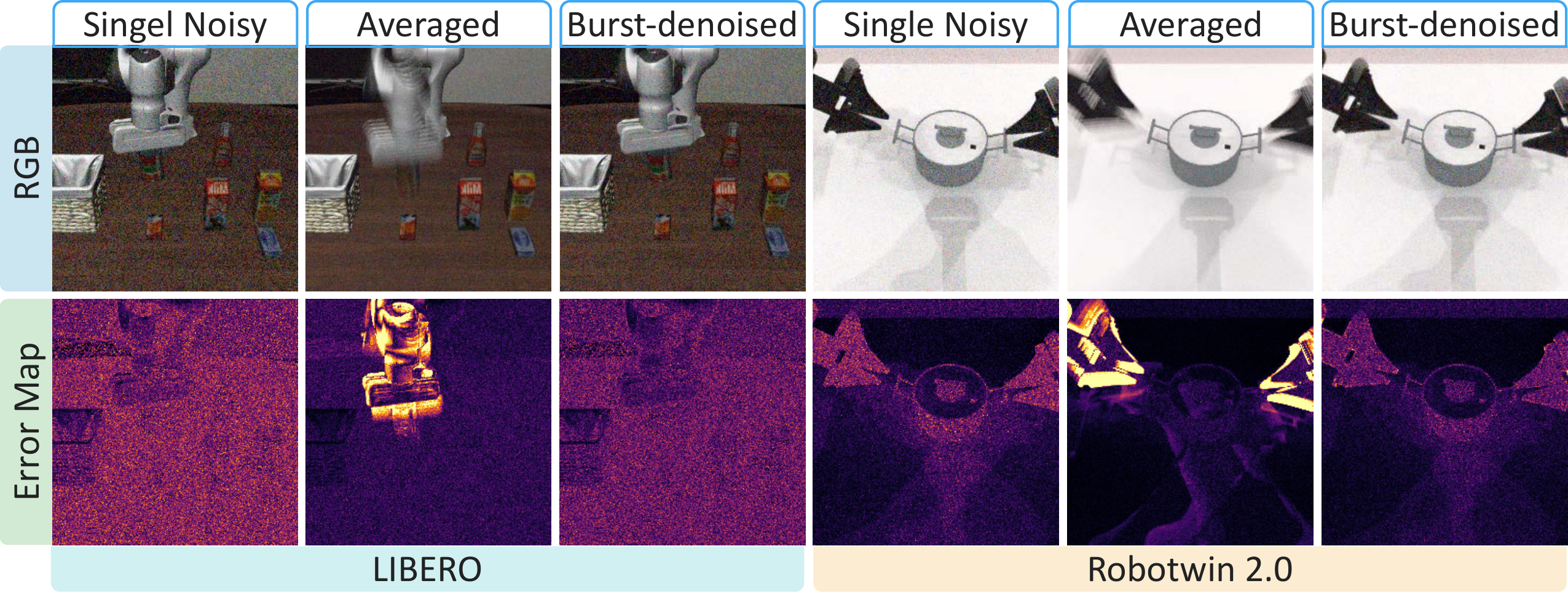}
\caption{Burst-denoising comparison on agent-view observations from
LIBERO~\citep{liu2023libero} (left) and
RoboTwin~2.0~\citep{chen2025robotwin} (right). The top row compares a single noisy frame,
direct averaging of $K=6$ consecutive frames, and our frequency-domain burst
denoising with $K=6$. The bottom row shows the corresponding error maps relative to the noise-free clean image. Direct averaging removes substantial sensor noise but blurs moving arms and objects, whereas burst denoising better preserves dynamic scene content while suppressing noise.}
\label{fig:burst_denoising_comparison}
\end{figure}

At policy step $t$, burst denoising uses the causal RAW sequence
\begin{equation}
    R_{t-K+1:t}
    =
    \{R_{t-K+1},\ldots,R_{t-1},R_t\},
    \qquad
    R_i\in[0,1]^{3\times H\times W},
\end{equation}
where $R_t$ is the current input frame. We use $K=6$, square patches of size
$N=32$, stride 16, and a separable two-dimensional Hann window. For a
historical-frame spectrum $F_i$ and the current reference spectrum $F_t$, let
$\Delta F_i=F_i-F_t$. The frequency-wise reliability and burst merge are
\begin{equation}
    \begin{aligned}
        \sigma_i^2&=\kappa\operatorname{mean}_{f}|\Delta F_i|^2,
        &\qquad
        r_i(f)&=\frac{\sigma_i^2}{|\Delta F_i(f)|^2+\sigma_i^2+10^{-8}},\\
        F_{\mathrm{out}}&=F_t+\gamma_{\mathrm{burst}}
        \sum_{i=1}^{K-1}\omega_i r_i\Delta F_i,
        &\qquad
        \gamma_{\mathrm{burst}}&=0.5.
    \end{aligned}
\end{equation}
The index $f$ spans the frequencies in each patch. We set $\kappa=1.8$ and
use historical weights
$\boldsymbol{\omega}=(0.05,0.075,0.125,0.1875,0.375)$ from the oldest frame
to the most recent historical frame. The current frame remains the reference
anchor with coefficient one. We fix the fusion strength to
$\gamma_{\mathrm{burst}}=0.5$ in the full model and evaluate
$\gamma_{\mathrm{burst}}\in\{0.2,0.5,0.8\}$ in
Table~\ref{tab:ablation_burst_length}. The reliability is computed
independently for each patch, channel, and frequency. Small spectral
differences are fused strongly,
while motion and misalignment are down-weighted. Reflection padding is used
when supported by the input dimensions and replication padding otherwise.
Inverse Fourier transformation and Hann-weighted overlap-add reconstruct the
denoised linear RAW frame $X_t$ before exposure and tone mapping to avoid
subsequent noise amplification. Figure~\ref{fig:burst_denoising_comparison}
compares the resulting observations with a single noisy frame and direct
averaging of the same $K=6$ frames. Direct averaging substantially suppresses
random sensor noise, but it also blurs the moving robot arm and manipulated
objects. Preserving these dynamic structures is important for accurate
grasping and for estimating the current manipulation state. In contrast, the
frequency-domain reliability weighting attenuates inconsistent motion and
misalignment during fusion, reducing noise while retaining sharper dynamic
content and lower error relative to the noise-free observation.

\subsection{Luminance Descriptor}
\label{app:luminance_descriptor}

The absolute luminance descriptor combines normalized linear- and log-luminance histograms with distribution quantiles and boundary-occupancy
statistics. Specifically,
\begin{equation}
    h_t^L
    =
    \left[
        \operatorname{Hist}_{64}(L_t),
        \operatorname{Hist}_{64}(\log(L_t+\epsilon)),
        \operatorname{Quant}(L_t),
        \rho_{\mathrm{dark}}(L_t),
        \rho_{\mathrm{clip}}(L_t)
    \right],
\end{equation}
where each histogram contains 64 normalized bins over a fixed range and
$\operatorname{Quant}(L_t)=[\pi_{0.01},\pi_{0.05},\pi_{0.50},\pi_{0.95},\pi_{0.99}]$ contains five intensity quantiles. For pixel index $u$, the boundary ratios are
\begin{equation}
    \rho_{\mathrm{dark}}(L_t)
    =
    \frac{1}{HW}\sum_u \mathbf{1}[L_t(u)<0.05],
    \qquad
    \rho_{\mathrm{clip}}(L_t)
    =
    \frac{1}{HW}\sum_u \mathbf{1}[L_t(u)>0.98].
\end{equation}
$h_t^L\in\mathbb{R}^{135}$ comprises two 64-bin histograms, five
quantiles, and two boundary ratios.

\subsection{Chroma Descriptor}
\label{app:chroma_descriptor}

Let
$\mathcal{V}_t=([\boldsymbol{\chi}_t]^{\mathrm R},
[\boldsymbol{\chi}_t]^{\mathrm G},[\boldsymbol{\chi}_t]^{\mathrm B},
[\mathbf q_t]_1,[\mathbf q_t]_2)$ contain the five
component maps of $\boldsymbol{\chi}_t$ and ${\mathbf q}_t$. The chroma descriptor is
\begin{equation}
    h_t^C
    =
    \operatorname{Concat}_{\psi\in\mathcal{V}_t}
    \left[
        \operatorname{Hist}_{B_C}(\psi),
        \mu(\psi),
        \sigma^2(\psi)
    \right],
\end{equation}
where $\operatorname{Hist}_{B_C}$ is a normalized $B_C$-bin histogram over a
fixed component-specific range. For pixel index $u$, the moments are
\begin{equation}
    \mu(\psi)=\frac{1}{HW}\sum_u \psi(u),
    \qquad
    \sigma^2(\psi)=\frac{1}{HW}\sum_u\left(\psi(u)-\mu(\psi)\right)^2.
\end{equation}
The resulting $h_t^C\in\mathbb{R}^{5(B_C+2)}$ summarizes the distributions of the
three chromaticity maps and two log-chrominance maps.

\subsection{Structured ISP Parameterization}
\label{app:isp_parameterization}

White balance has two degrees of freedom. The network predicts the red and
green offsets and constructs the zero-sum channel offsets as
\begin{equation}
    w_t=(w_t^{\mathrm R},w_t^{\mathrm G},-w_t^{\mathrm R}-w_t^{\mathrm G}).
\end{equation}
This parameterization separates relative channel gains from the global
exposure parameter $e_t$.

\section{Neural ISP Training Protocol}
\label{app:training_details}

This section specifies the shared frozen-policy training protocol in
Section~\ref{app:frozen_policy_training}, the baseline ISP configurations in
Section~\ref{app:baseline_isp_training}, and the optimization and temporal
sampling used by \methodname{} in Section~\ref{app:rawvla_optimization}.

\subsection{Frozen-Policy Training Protocol}
\label{app:frozen_policy_training}

We train an independent ISP module for every policy backbone and optimize all ISP modules for 2,000 optimizer steps. The pretrained vision or world-model backbone and action head are frozen and kept in evaluation mode. Their forward passes are not enclosed by \texttt{no\_grad}, so the native action loss remains differentiable with respect to the processed image while no policy parameter is updated. Each method receives the same three-channel pseudo-RAW signal in the linear RAW representation and produces RGB observations for the policy. In the main benchmark experiments, no ISP module is trained
with paired RGB reconstruction supervision. All methods use one NVIDIA A100
80GB GPU, per-device batch size one, and eight gradient-accumulation
micro-steps. The shared optimizer is AdamW with learning rate $10^{-4}$,
$\beta=(0.9,0.95)$, numerical epsilon $10^{-8}$, 100 warmup updates, cosine
decay to $10^{-6}$, and gradient-norm clipping at 1.0. Weight decay is
$10^{-8}$ for Qwen3~\citep{starvla2026} and WM4A~\citep{starvla2026}
backbones and $10^{-10}$ for PI0~\citep{black2024pi0} and
PI0.5~\citep{physicalintelligence2025pi05}.

\subsection{Baseline Neural ISP Configurations}
\label{app:baseline_isp_training}

We evaluate DarkISP~\citep{guo2025darkisp}, RAM~\citep{gamrian2025ram},
RAW-Adapter~\citep{cui2024rawadapter}, and RAWild~\citep{liu2026rawild} under
the common 2k-step budget. Methods originally designed for Bayer inputs use
deterministic channel adapters. DarkISP~\citep{guo2025darkisp} receives
$[R,G,B,G]$, while RAM~\citep{gamrian2025ram} reduces its packed green
channels to $[R,(G_r+G_b)/2,B]$. RAW-Adapter~\citep{cui2024rawadapter}
applies its learned exposure, denoising and sharpening, white-balance,
color-matrix, and optional LUT stages. RAWild~\citep{liu2026rawild} uses the
same pseudo-RAW observation as both its guide and apply image. These
adaptations preserve a common camera signal while retaining each method's ISP
parameterization.

To alleviate temporal inconsistency in methods designed for individual images, training samples trajectory-local windows of eight observations, applies the shared ISP independently to every observation, and averages the resulting action losses. This introduces no recurrence, cross-frame fusion, or explicit temporal-consistency loss. Direct-regression action heads use
their native L1 objective, while diffusion and flow-matching heads retain their native velocity-prediction objective.

For LIBERO~\citep{liu2023libero}, the action horizon is eight for the comparison
runs. For RoboTwin~2.0~\citep{chen2025robotwin}, the head, left-wrist, and
right-wrist views are processed by the same ISP module.
Qwen3-OFT~\citep{starvla2026} predicts a 50-step normalized action chunk, while
FastWAM~\citep{starvla2026} uses its released 32-step flow-matching objective.
The optimization budget remains 2,000 steps for both environments and all four
comparison ISP modules.

\subsection{\methodname{} Training and Temporal Sampling}
\label{app:rawvla_optimization}

\methodname{} follows the common optimization protocol above. Forward passes use \textit{bfloat16} where supported, with numerically sensitive action computations promoted to \textit{float32}.

Each sample contains $N_{\mathrm{rec}}=6$ recurrent endpoints and a $K=6$ causal camera burst at every endpoint. Following the native policy observation intervals, the endpoint stride is eight environment steps for Qwen3~\citep{starvla2026} and WM4A~\citep{starvla2026} policies and five for PI0~\citep{black2024pi0} and PI0.5~\citep{physicalintelligence2025pi05}. The recurrent operating-point estimator reads the endpoint of each burst. The final policy image is produced from the six adjacent frames in the terminal burst. Earlier calls contribute through the recurrent state, and the complete six-update sequence is trained with backpropagation through time without detaching the hidden state or the previous ISP parameters. At episode boundaries, invalid history indices are clipped to the nearest valid frame.

Qwen3-OFT~\citep{starvla2026} and WM4A-Wan-OFT~\citep{starvla2026} use their
mean-reduced action L1 losses. Qwen3-PI~\citep{starvla2026} and
WM4A-Cosmos-GR00T~\citep{starvla2026} use eight independently sampled
diffusion or flow-matching conditions per trajectory window.
PI0~\citep{black2024pi0} and PI0.5~\citep{physicalintelligence2025pi05} use
their native mean-reduced flow-matching losses. These stochastic repeats are
Monte Carlo samples of the action objective and are distinct from the six
recurrent observations and the eight optimizer-accumulation micro-steps.

The brightness and chroma weights are fixed to $\lambda_b=0.001$ and
$\lambda_{\mathrm{chroma}}=0.001$. The frozen VLA policies retain their native regression, diffusion, or flow-matching objectives, whose numerical scales and image-gradient magnitudes differ substantially. We therefore use the policy-specific coefficients in Table~\ref{tab:policy_action_loss_weights} to place action supervision on a comparable scale relative to the two fixed photometric regularizers.

\begin{table}[!t]
\centering
\caption{Action-loss coefficients used to train \methodname{} with each frozen VLA policy.}
\label{tab:policy_action_loss_weights}
\small
\begin{tabularx}{0.55\textwidth}{@{}>{\raggedright\arraybackslash}X c@{}}
\toprule
Frozen VLA policy & $\lambda_a$ \\
\midrule
Qwen3-PI & 0.0015 \\
Qwen3-OFT & 0.20 \\
WM4A-Cosmos-GR00T & 0.01 \\
WM4A-Wan-OFT & 0.02 \\
PI0 & 0.15 \\
PI0.5 & 0.10 \\
\bottomrule
\end{tabularx}
\end{table}

The brightness auxiliary uses a two-sided L1 penalty with target mean 0.48.
The chroma auxiliary is zero within an expanded unpaired RGB envelope and
penalizes log-chroma descriptors outside it. Auxiliary renderings use
straight-through hard clipping and branch-specific parameter detachment, whereas
the deployed policy operator uses ordinary hard clipping. Paired default-lighting
RGB is used only for no-gradient checkpoint selection and never contributes to
the training loss.

\section{\benchmarkname{} Rendering Visualizations}
\label{app:rawvla_bench_visualizations}

\begin{figure}[t]
\centering
\includegraphics[width=\textwidth]{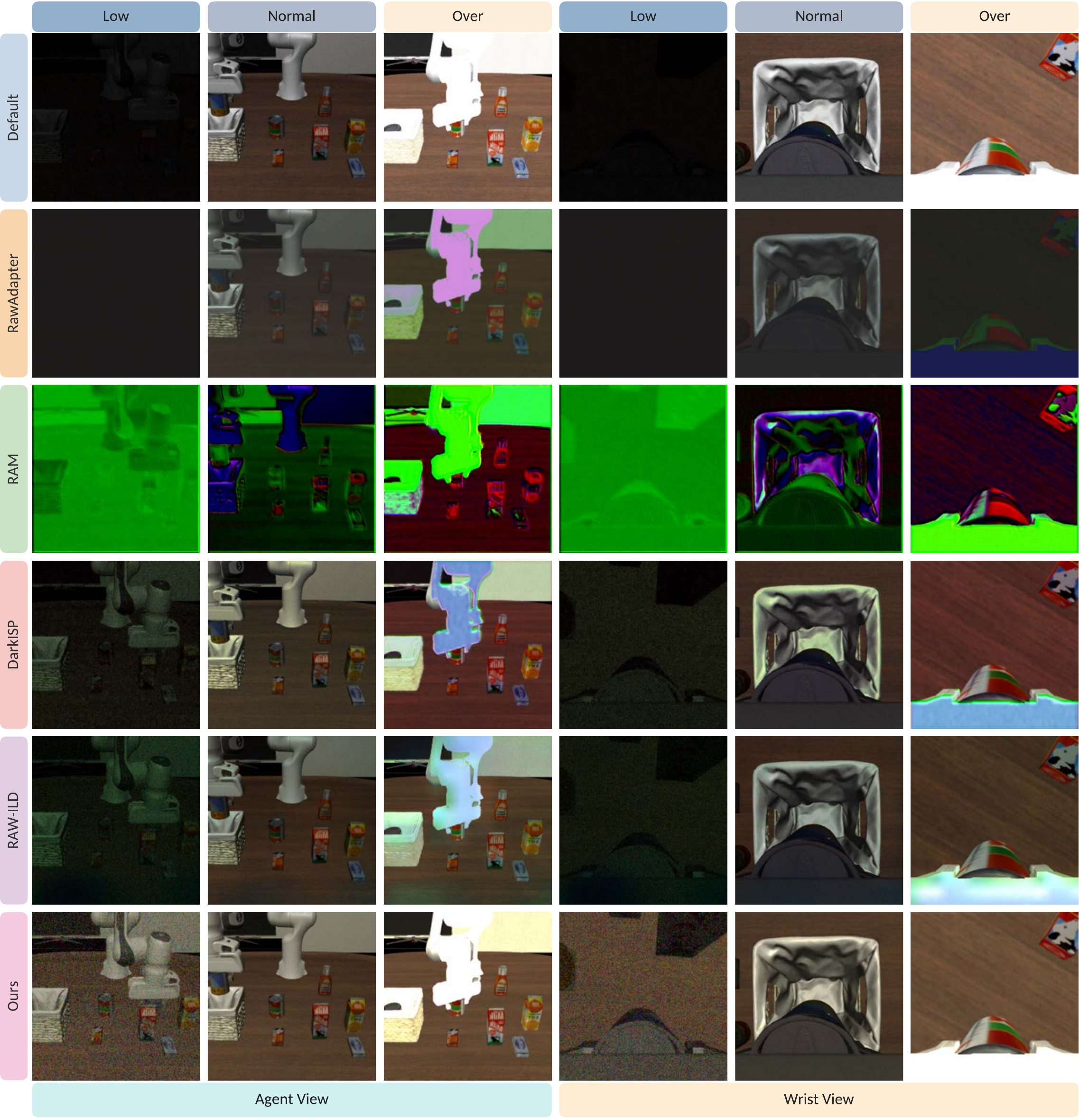}
\caption{Processed agent-view and wrist-view observations from the LIBERO
portion of \benchmarkname{} under low, normal, and overexposed illumination.
The rows compare the Default ISP with neural ISP methods using Qwen3-OFT as
the policy backbone.}
\label{fig:rawvla_bench_libero_visualization}
\end{figure}

\begin{figure}[t]
\centering
\includegraphics[width=\textwidth]{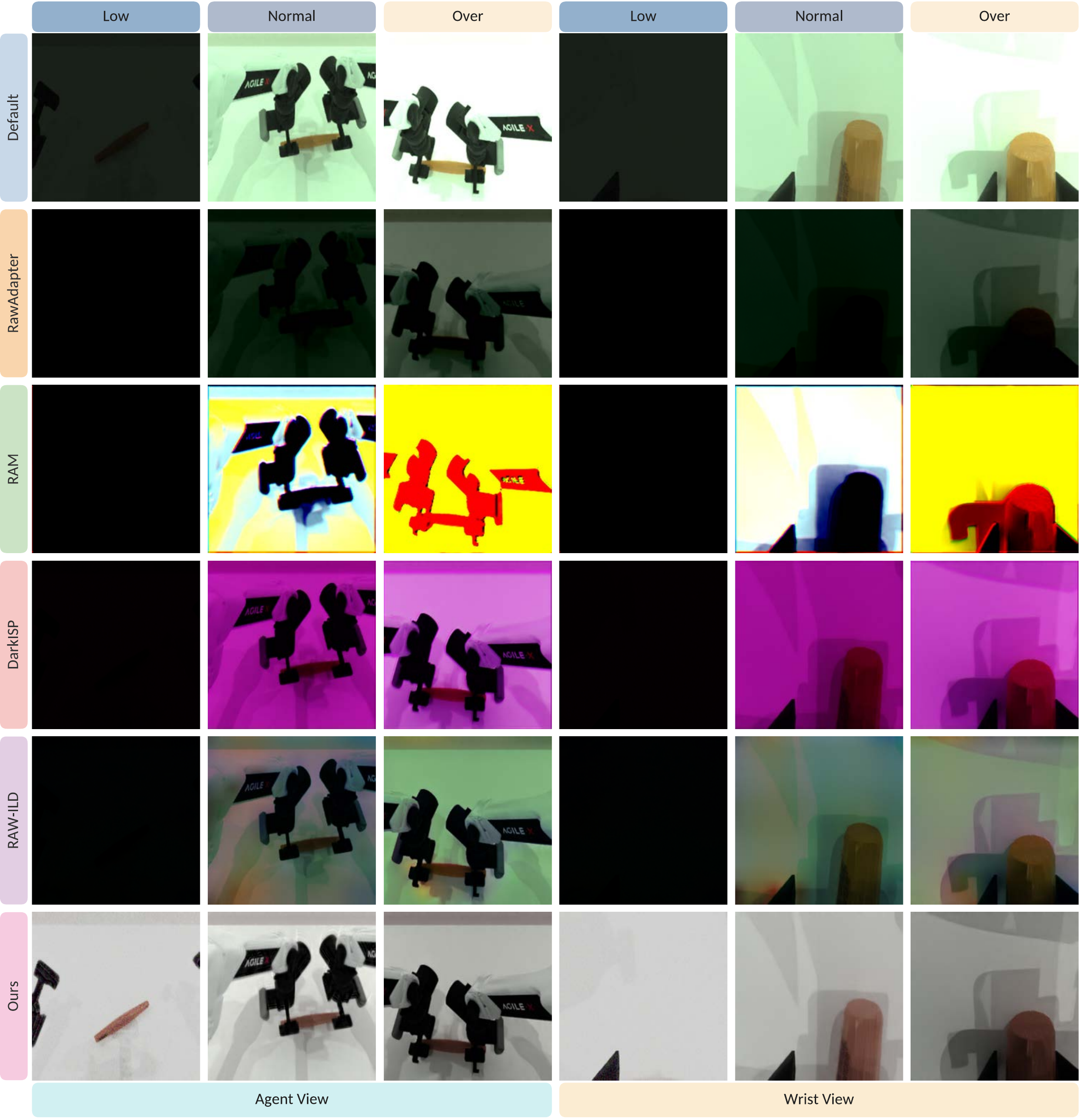}
\caption{Processed agent-view and wrist-view observations from the RoboTwin~2.0
portion of \benchmarkname{} under low, normal, and overexposed illumination.
The rows compare the Default ISP with neural ISP methods using PI-0.5 as the
policy backbone.}
\label{fig:rawvla_bench_robotwin_visualization}
\end{figure}

Figures~\ref{fig:rawvla_bench_libero_visualization}
and~\ref{fig:rawvla_bench_robotwin_visualization} compare the observations
rendered by the Default ISP, four neural ISP baselines, and \methodname{} on
the LIBERO and RoboTwin~2.0 portions of \benchmarkname{}. Each comparison uses
matched RAW observations from the same scene and robot state under low,
normal, and overexposed illumination. Agent and wrist views are shown together
to reveal whether each ISP produces a consistent multi-view representation for
the downstream policy.

On LIBERO, the methods produce substantially different brightness, color, and
local contrast from identical RAW observations. Several baseline renderings
remain attenuated under low light or introduce strong chromatic and tonal
changes, while \methodname{} preserves task-relevant scene structure across
both camera views.

The RoboTwin~2.0 examples show the same variation across three cameras and a
different renderer. The qualitative comparison illustrates the observations
received by the policy, while the closed-loop results in the main paper measure
their utility for manipulation.

\section{Additional Robustness Evaluation under Severe Haze}
\label{app:libero_haze}

An adaptive ISP can improve the visibility and task relevance of observations
by adjusting their photometric rendering before they reach the policy. This
capability may also benefit adverse visual conditions beyond the illumination
and sensor degradations included in \benchmarkname{}. To test this broader
robustness, we draw on the adverse environments introduced by
LIBERO-Plus~\citep{fei2025liberoplus} and construct a substantially stronger
haze condition. We then evaluate whether \methodname{} can enhance these
severely degraded observations and recover the closed-loop performance of a
frozen PI-0 policy~\citep{black2024pi0}.

\paragraph{Experimental Setting.}
We follow the image-space haze construction of
LIBERO-Plus~\citep{fei2025liberoplus}, but increase the haze coefficient to
$\alpha_{\mathrm h}=0.95$. For every pixel and color channel of a clean
simulator RGB image
$I_{\mathrm{clean}}\in[0,255]$, we compute
\begin{equation}
    I_{\mathrm{haze}}
    =\operatorname{clip}\!\left(
        (1-\alpha_{\mathrm h})I_{\mathrm{clean}}
        +\alpha_{\mathrm h}I_{\mathrm{air}},0,255
    \right),
    \qquad \alpha_{\mathrm h}=0.95,\quad I_{\mathrm{air}}=235.
    \label{eq:libero_severe_haze}
\end{equation}
The operation is performed in \textit{float32} and the result is converted to
\textit{uint8}. Thus, only $5\%$ of the original RGB signal is retained and
$95\%$ gray-white airlight is mixed into the observation. We add no sensor
noise, blur, random image noise, or additional quantization. This experiment
uses hazy RGB observations rather than sensor RAW. In this evaluation,
``RAW'' in \methodname{} refers only to the model name.

\begin{figure}[t]
\centering
\includegraphics[width=\textwidth]{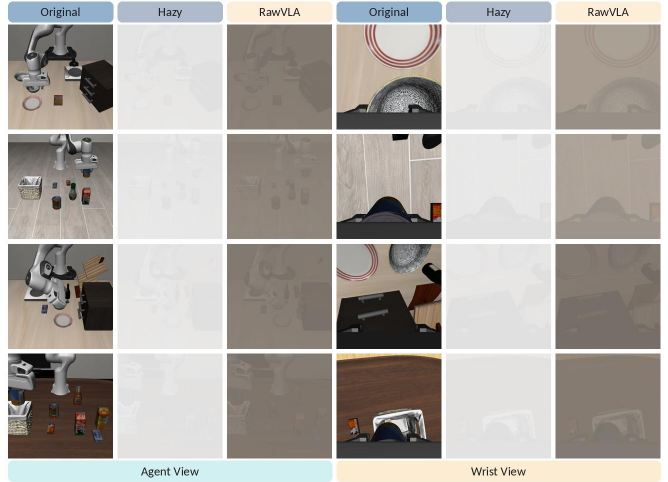}
\caption{Representative agent-view and wrist-view observations from LIBERO
under severe haze. Each triplet shows the original simulator RGB image, the
hazy input generated with $\alpha_{\mathrm h}=0.95$ and
$I_{\mathrm{air}}=235$, and the corresponding RGB output produced by
\methodname{}. The same enhanced observations are provided to the frozen PI-0
policy during closed-loop evaluation.}
\label{fig:libero_haze_pi0}
\end{figure}

At every simulator step, Equation~\ref{eq:libero_severe_haze} is applied
independently to the agent and wrist views. The PI-0 baseline receives the two
hazy views directly. In the enhanced condition, each view is first processed
by an $\alpha_{\mathrm h}=0.95$-specific \methodname{} checkpoint and the
resulting RGB images
are passed to the same frozen PI-0 policy. Enhancement runs at the simulator
resolution, before the native PI-0 pipeline resizes each view to
$224\times224$. We use the standard causal \methodname{} inference procedure
without access to future observations or clean reference frames. Language
instructions, robot states, control frequency, action horizon, and all other
policy settings are held fixed.
Figure~\ref{fig:libero_haze_pi0} shows representative observations from both
camera views before and after haze enhancement.

We train a haze-specific instance of the architecture described in
Section~\ref{sec:rawvla} using paired reconstruction supervision rather than
the frozen-policy objective in Section~\ref{app:training_details}. Training
runs for 1,500 updates on paired LIBERO replay images from both camera views.
Each target is a clean simulator RGB frame, and its input is generated online
from that same frame using the exact $\alpha_{\mathrm h}=0.95$ and
$I_{\mathrm{air}}=235$ transform, without strength jitter. The clean
reconstruction targets are used only during training and are unavailable
during closed-loop evaluation.

The two policy conditions use identical tasks, official initial states, and
episode indices. PI-0 diffusion noise is generated deterministically from the
language instruction and robot state to reduce sampling variation between the
paired visual conditions. Success is determined by the standard LIBERO task
predicate. Each suite includes ten tasks with 50 distinct rollouts per task.

\paragraph{Evaluation on Hazy LIBERO.}
On held-out image pairs, enhancement reduces mean absolute error from $0.4271$
to $0.1688$ and raises PSNR from $6.56$ to $13.60$ dB.
Table~\ref{tab:libero_haze_pi0} reports closed-loop suite success rates and the
macro-average that weights the four suites equally.

\begin{table}[t]
\centering
\caption{PI-0 success rates (\%) on LIBERO under severe haze
($\alpha_{\mathrm h}=0.95$, $I_{\mathrm{air}}=235$), evaluated with 50
rollouts per task. Avg. is the macro-average over the four suites.}
\label{tab:libero_haze_pi0}
\setlength{\tabcolsep}{7pt}
\begin{tabular}{@{}lccccc@{}}
\toprule
Method & Spatial & Object & Goal & LIBERO-10 & Avg. \\
\midrule
PI-0 & 78.0 & 84.0 & 82.0 & 22.0 & 66.5 \\
PI-0 with \methodname{} & \firstbest{98.0} & \firstbest{96.0} & \firstbest{96.0} & \firstbest{63.6} & \firstbest{88.4} \\
\midrule
Improvement (pp) & \textcolor{improvementgreen}{+20.0} & \textcolor{improvementgreen}{+12.0} & \textcolor{improvementgreen}{+14.0} & \textcolor{improvementgreen}{+41.6} & \textcolor{improvementgreen}{+21.9} \\
\bottomrule
\end{tabular}
\end{table}

Across the four equally sized suites, PI-0 achieves an average success rate of
66.5\%. Processing the hazy observations with \methodname{} before PI-0 raises
this average to 88.4\%, an improvement of 21.9 percentage points. On
LIBERO-10, \methodname{} enables PI-0 to succeed in 230 rollouts that fail
without enhancement, while 22 rollouts show the opposite outcome. The paired
comparison therefore yields 208 additional successful rollouts. A two-sided
exact McNemar test confirms that this improvement is statistically significant,
with $p=7.16\times10^{-45}$.

\begin{figure}[t]
\centering
\includegraphics[width=\textwidth]{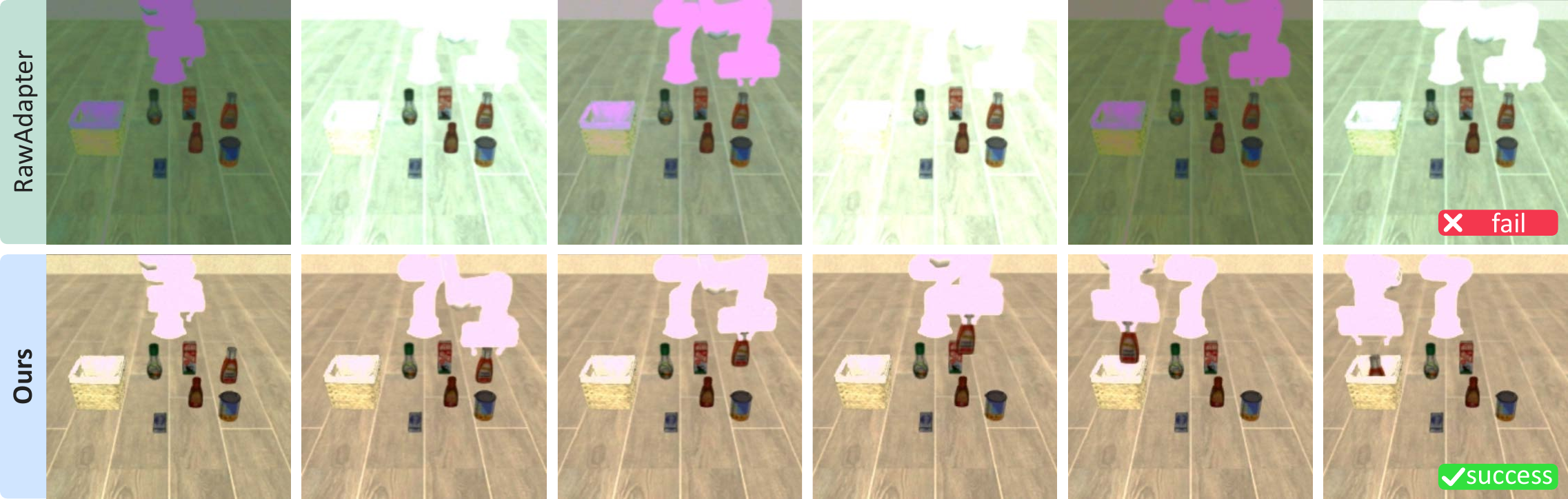}
\caption{Temporal rendering stability on an overexposed
LIBERO~\citep{liu2023libero} trajectory with
Qwen3-OFT~\citep{starvla2026}. RAW-Adapter~\citep{cui2024rawadapter} exhibits pronounced frame-to-frame appearance
fluctuations that destabilize the policy and lead to task failure, whereas
\methodname{} maintains temporally consistent observations and successfully completes
the task.}
\label{fig:libero_temporal_jitter}
\end{figure}

\section{Additional Ablation Studies}
\label{app:additional_ablations}

The following tables separate structured ISP operators from burst-length design
choices. LIBERO~\citep{liu2023libero} uses Qwen3-OFT~\citep{starvla2026}, and
RoboTwin~2.0~\citep{chen2025robotwin} uses
PI-0.5~\citep{physicalintelligence2025pi05}, matching the main
ablation protocol.

Because exhaustive evaluation of every additional ablation is computationally
expensive, we use a reduced evaluation budget for LIBERO~\citep{liu2023libero}
in this appendix. We retain all 40 tasks across the four
LIBERO~\citep{liu2023libero} suites, but evaluate each task from
10 distinct initial states, rather than the 50 initial states used for the main
results. For RoboTwin~2.0~\citep{chen2025robotwin}, we retain the full evaluation protocol with 13 tasks
and 50 distinct initial states per task.

\paragraph{Training Objectives.}
Table~\ref{tab:ablation_objective} isolates the action, brightness, and chroma
objectives. The action objective alone preserves normal-light performance but
fails under severe illumination changes, whereas the two photometric
auxiliaries without action supervision improve robustness but do not fully
align the rendered observations with the frozen policy. Combining all three
objectives gives the strongest overall performance, reaching 51.32\% on
LIBERO~\citep{liu2023libero} and 57.12\% on
RoboTwin~2.0~\citep{chen2025robotwin}. These results show that the action loss provides
policy alignment, while the brightness and chroma terms stabilize adaptation in
the adverse regimes where action supervision alone is difficult to optimize.

\begin{table}[!tp]
\centering
\caption{Ablation of the three training objectives. The full objective is compared with action-free auxiliary supervision and each loss used in isolation. Loss weights are fixed and are not swept.}
\label{tab:ablation_objective}
\scriptsize
\setlength{\tabcolsep}{2.1pt}
\renewcommand{\arraystretch}{1.05}
\resizebox{\textwidth}{!}{%
\begin{tabular}{@{}lccc*{6}{c}@{\hspace{7pt}}*{6}{c}@{}}
\toprule
& \multicolumn{3}{c}{Training objective}
& \multicolumn{6}{c}{LIBERO / Qwen3-OFT}
& \multicolumn{6}{c}{RoboTwin~2.0 / PI-0.5} \\
\cmidrule(lr){2-4}\cmidrule(lr){5-10}\cmidrule(lr){11-16}
Variant & $\mathcal{L}_{\rm action}$ & $\mathcal{L}_{\rm bright}$ & $\mathcal{L}_{\rm color}$
& Ext.Low & Low & Normal & Over & Ext.Over & Avg.
& Ext.Low & Low & Normal & Over & Ext.Over & Avg. \\
\midrule
Action loss only    & \checkmark &            &            & 0.00 & 0.00 & \firstbest{97.75} & \firstbest{44.50} & \firstbest{29.25} & 34.30 & 0.00 & 0.00 & \firstbest{67.72} & \thirdbest{54.76} & \thirdbest{46.40} & 33.70 \\
w/o action loss     &            & \checkmark & \checkmark & \secondbest{9.50} & \secondbest{72.75} & \secondbest{97.50} & \secondbest{44.25} & \secondbest{25.50} & \secondbest{49.90} & \secondbest{11.81} & \secondbest{55.12} & \thirdbest{62.20} & \secondbest{59.52} & \secondbest{51.20} & \secondbest{47.94} \\
Brightness only     &            & \checkmark &            & \thirdbest{4.00} & \thirdbest{39.50} & \thirdbest{97.25} & 41.25 & \thirdbest{21.50} & \thirdbest{40.70} & \thirdbest{7.87} & \thirdbest{41.73} & \secondbest{65.35} & 53.17 & \secondbest{51.20} & \thirdbest{43.83} \\
Color only          &            &            & \checkmark & 0.00 & 0.00 & \secondbest{97.50} & \thirdbest{43.50} & 20.00 & 32.20 & 0.00 & 12.60 & 53.54 & 52.38 & 26.40 & 23.73 \\
\midrule
Full \methodname{}         & \checkmark & \checkmark & \checkmark & \firstbest{14.05} & \firstbest{81.35} & 96.95 & 43.40 & 20.85 & \firstbest{51.32} & \firstbest{25.98} & \firstbest{63.78} & \firstbest{67.72} & \firstbest{65.08} & \firstbest{63.20} & \firstbest{57.12} \\
\bottomrule
\end{tabular}%
}
\end{table}
\begin{table}[!tp]
\centering
\caption{Ablation study of \methodname{}'s modules on LIBERO with Qwen3-OFT~\citep{starvla2026}. We report success rates (\%) across five illumination levels and overall. Each ablated variant removes one ISP module while keeping the remaining architecture unchanged. Shared tone denotes the shared-head equivalence control.}
\label{tab:ablation_isp_libero}
\footnotesize
\setlength{\tabcolsep}{0.4pt}
\renewcommand{\arraystretch}{0.95}
\resizebox{\textwidth}{!}{%
\begin{tabular}{@{}l*{18}{>{\centering\arraybackslash}p{2.65em}}@{}}
\toprule
\multirow{2}{*}{Variant}
& \multicolumn{3}{c}{Ext.Low} & \multicolumn{3}{c}{Low}
& \multicolumn{3}{c}{Normal} & \multicolumn{3}{c}{Over}
& \multicolumn{3}{c}{Ext.Over} & \multicolumn{3}{c}{Overall} \\
\cmidrule(lr){2-4}\cmidrule(lr){5-7}\cmidrule(lr){8-10}\cmidrule(lr){11-13}\cmidrule(lr){14-16}\cmidrule(lr){17-19}
& SOG & Long & Avg & SOG & Long & Avg & SOG & Long & Avg
& SOG & Long & Avg & SOG & Long & Avg & SOG & Long & Avg \\
\midrule
w/o Exposure & 6.87 & 3.60 & 6.05 & 42.80 & 22.00 & 37.60 & \firstbest{98.53} & \thirdbest{92.20} & \thirdbest{96.95} & \secondbest{92.34} & \firstbest{60.00} & \firstbest{84.25} & 38.47 & \secondbest{1.00} & 29.10 & 55.80 & 35.76 & 50.79 \\
w/o White-balance & \firstbest{15.87} & \firstbest{12.60} & \firstbest{15.05} & \firstbest{87.46} & \thirdbest{62.00} & \secondbest{81.10} & \firstbest{98.53} & \secondbest{94.20} & \firstbest{97.45} & 91.67 & 53.00 & 82.00 & 42.80 & \firstbest{2.00} & 32.60 & \thirdbest{67.27} & \secondbest{44.76} & \thirdbest{61.64} \\
w/o CCM & \thirdbest{12.20} & \firstbest{12.60} & \thirdbest{12.30} & 86.46 & \thirdbest{62.00} & 80.35 & \secondbest{97.86} & \firstbest{95.20} & \secondbest{97.20} & \firstbest{92.67} & \secondbest{58.00} & \secondbest{84.00} & 43.13 & \firstbest{2.00} & 32.85 & 66.47 & \firstbest{45.96} & 61.34 \\
w/o Tone-mapping & 7.53 & \thirdbest{5.60} & 7.05 & 73.46 & 46.00 & 66.60 & \thirdbest{97.53} & 91.20 & 95.95 & 90.67 & 54.00 & 81.50 & \thirdbest{45.13} & \firstbest{2.00} & \thirdbest{34.35} & 62.87 & 39.76 & 57.09 \\
\hspace{0.6em}$\llcorner$\, Shared tone & \secondbest{15.20} & \secondbest{10.60} & \secondbest{14.05} & \thirdbest{86.80} & \secondbest{63.00} & \thirdbest{80.85} & \firstbest{98.53} & \thirdbest{92.20} & \thirdbest{96.95} & \thirdbest{92.00} & 54.00 & \thirdbest{82.50} & \secondbest{47.47} & \secondbest{1.00} & \secondbest{35.85} & \secondbest{68.00} & \thirdbest{44.16} & \secondbest{62.04} \\
\midrule
Full \methodname{} & \secondbest{15.20} & \secondbest{10.60} & \secondbest{14.05} & \secondbest{87.13} & \firstbest{64.00} & \firstbest{81.35} & \firstbest{98.53} & \thirdbest{92.20} & \thirdbest{96.95} & 91.67 & \thirdbest{55.00} & \thirdbest{82.50} & \firstbest{47.80} & \firstbest{2.00} & \firstbest{36.35} & \firstbest{68.07} & \secondbest{44.76} & \firstbest{62.24} \\
\bottomrule
\end{tabular}%
}
\end{table}

\paragraph{Structured ISP Operators.}
Table~\ref{tab:ablation_isp_libero} evaluates exposure, white balance, color
correction, and tone mapping on LIBERO~\citep{liu2023libero}. Removing exposure causes the largest
overall degradation, reducing average success from 62.24\% to 50.79\%.
Removing tone mapping also lowers robustness, while white balance and the
color-correction matrix provide complementary gains across illumination
conditions. The shared-tone control nearly matches the full model,
supporting the use of a single achromatic curve across channels while leaving
chromatic adaptation to white balance and color correction.

\paragraph{Cross-Benchmark Operator Effects.}
Table~\ref{tab:ablation_isp_robotwin} repeats the operator analysis on
RoboTwin~2.0~\citep{chen2025robotwin}. Exposure remains the most consequential component. Disabling it
reduces average success from 57.12\% to 43.35\%. The remaining operator
ablations also reduce the overall result, confirming that exposure, chromatic
correction, and tone mapping remain complementary across environments and
policy backbones.

\begin{table}[!tp]
\centering
\caption{Inference-only ISP component ablations on RoboTwin~2.0 using PI-0.5. All ablated variants use the same frozen checkpoint and force one ISP component to identity. Full (inference-only) keeps all components enabled, while Full \methodname{} reports the main benchmark result.}
\label{tab:ablation_isp_robotwin}
\small
\setlength{\tabcolsep}{6pt}
\renewcommand{\arraystretch}{1.05}
\begin{tabular*}{\textwidth}{@{\extracolsep{\fill}}lcccccc@{}}
\toprule
Variant & Ext.Low & Low & Normal & Over & Ext.Over & Avg. \\
\midrule
w/o Exposure        & 12.60 & 18.11 & 65.35 & \secondbest{61.11} & \thirdbest{60.00} & 43.35 \\
w/o White-balance   & \firstbest{25.98} & \thirdbest{58.27} & \thirdbest{66.14} & \firstbest{65.08} & 54.40 & \thirdbest{53.96} \\
w/o CCM             & \secondbest{25.20} & \secondbest{61.42} & \firstbest{69.29} & \thirdbest{57.14} & \secondbest{62.40} & \secondbest{55.06} \\
w/o Tone-mapping    & \thirdbest{22.05} & 50.39 & 64.57 & \firstbest{65.08} & \thirdbest{60.00} & 52.37 \\
\midrule
Full \methodname{}          & \firstbest{25.98} & \firstbest{63.78} & \secondbest{67.72} & \firstbest{65.08} & \firstbest{63.20} & \firstbest{57.12} \\
\bottomrule
\end{tabular*}
\end{table}

\begin{table}[!tp]
\centering
\caption{Burst-denoising ablations. We study the burst length $K$, global fusion weight $\eta$, and frequency-wise reliability decomposition. We compare the rendered output $\hat{I}$ against the corresponding noise-free simulation ground truth $I_{\rm GT}$ using PSNR (dB) and SSIM. Average is computed over the ExtremeLow, Low, and Normal illumination regimes. Underlined settings denote the default configuration.}
\label{tab:ablation_burst_length}
\scriptsize
\setlength{\tabcolsep}{2.8pt}
\renewcommand{\arraystretch}{1.05}
\resizebox{\textwidth}{!}{%
\begin{tabular}{@{}lll*{4}{cc}@{}}
\toprule
\multirow{2}{*}{Benchmark} & \multirow{2}{*}{Ablation} & \multirow{2}{*}{Setting}
& \multicolumn{2}{c}{Ext.Low} & \multicolumn{2}{c}{Low}
& \multicolumn{2}{c}{Normal} & \multicolumn{2}{c}{Average} \\
\cmidrule(lr){4-5}\cmidrule(lr){6-7}\cmidrule(lr){8-9}\cmidrule(lr){10-11}
& & & PSNR $\uparrow$ & SSIM $\uparrow$
& PSNR $\uparrow$ & SSIM $\uparrow$
& PSNR $\uparrow$ & SSIM $\uparrow$
& PSNR $\uparrow$ & SSIM $\uparrow$ \\
\midrule
\multirow[c]{7}{*}{\begin{tabular}[c]{@{}c@{}}LIBERO\\Qwen3-OFT\end{tabular}}
& \multirow{3}{*}{Burst length} & $K=1$ & 18.93 & 0.117 & 19.99 & 0.261 & 36.69 & 0.863 & 25.21 & 0.414 \\
& & $K=3$ & 20.15 & 0.144 & \thirdbest{21.34} & \thirdbest{0.304} & \secondbest{37.66} & 0.893 & \thirdbest{26.38} & 0.447 \\
& & \underline{$K=6$} & \firstbest{21.98} & \firstbest{0.195} & \firstbest{23.27} & \firstbest{0.373} & \thirdbest{37.57} & \firstbest{0.921} & \firstbest{27.61} & \firstbest{0.497} \\
\cmidrule(lr){2-11}
& \multirow{3}{*}{Fusion strength} & $\eta=0.2$ & 19.63 & 0.132 & 20.79 & 0.286 & 37.44 & 0.882 & 25.95 & 0.433 \\
& & \underline{$\eta=0.5$} & \firstbest{21.98} & \firstbest{0.195} & \firstbest{23.27} & \firstbest{0.373} & \thirdbest{37.57} & \firstbest{0.921} & \firstbest{27.61} & \firstbest{0.497} \\
& & $\eta=0.8$ & \secondbest{20.78} & \secondbest{0.160} & \secondbest{22.04} & \secondbest{0.328} & \firstbest{37.89} & \thirdbest{0.905} & \secondbest{26.90} & \secondbest{0.465} \\
\cmidrule(lr){2-11}
& Frequency & w/o decomposition & \thirdbest{20.54} & \thirdbest{0.152} & 20.43 & 0.295 & 33.21 & \secondbest{0.908} & 24.73 & \thirdbest{0.452} \\
\midrule
\multirow[c]{7}{*}{\begin{tabular}[c]{@{}c@{}}RoboTwin~2.0\\PI-0.5\end{tabular}}
& \multirow{3}{*}{Burst length} & $K=1$ & \secondbest{18.16} & \firstbest{0.082} & \firstbest{27.30} & \firstbest{0.636} & \firstbest{48.45} & \firstbest{0.986} & \firstbest{31.30} & \firstbest{0.568} \\
& & $K=3$ & \secondbest{18.16} & \firstbest{0.082} & \thirdbest{27.27} & \secondbest{0.635} & \thirdbest{46.35} & \thirdbest{0.984} & \thirdbest{30.59} & \secondbest{0.567} \\
& & \underline{$K=6$} & \firstbest{18.18} & \firstbest{0.082} & \secondbest{27.28} & \thirdbest{0.634} & \secondbest{47.35} & \secondbest{0.985} & \secondbest{30.94} & \secondbest{0.567} \\
\cmidrule(lr){2-11}
& \multirow{3}{*}{Fusion strength} & $\eta=0.2$ & \secondbest{18.16} & \firstbest{0.082} & 27.22 & \secondbest{0.635} & 45.20 & 0.982 & 30.20 & \thirdbest{0.566} \\
& & \underline{$\eta=0.5$} & \firstbest{18.18} & \firstbest{0.082} & \secondbest{27.28} & \thirdbest{0.634} & \secondbest{47.35} & \secondbest{0.985} & \secondbest{30.94} & \secondbest{0.567} \\
& & $\eta=0.8$ & \thirdbest{18.14} & \firstbest{0.082} & 27.14 & \secondbest{0.635} & 43.53 & 0.978 & 29.60 & 0.565 \\
\cmidrule(lr){2-11}
& Frequency & w/o decomposition & 18.08 & 0.079 & 26.35 & 0.628 & 38.47 & 0.967 & 27.63 & 0.558 \\
\bottomrule
\end{tabular}
}
\end{table}

\paragraph{Burst-Denoising Design.}
Table~\ref{tab:ablation_burst_length} studies burst length, fusion strength, and
frequency-wise reliability decomposition using reconstruction quality relative
to noise-free observations. Longer bursts substantially improve
LIBERO~\citep{liu2023libero}, with $K=6$ outperforming single-frame processing,
whereas RoboTwin~2.0~\citep{chen2025robotwin} shows
diminishing returns because its observations contain less recoverable temporal
noise. A moderate fusion strength provides the best average trade-off on both
benchmarks. Removing the frequency-wise decomposition produces the clearest
degradation, confirming that motion-aware spectral reliability is important for
suppressing noise without indiscriminately averaging dynamic content.

\paragraph{Temporal Stability.}
Figure~\ref{fig:libero_temporal_jitter} complements the quantitative ablations
with an overexposed LIBERO~\citep{liu2023libero} trajectory. Compared with the
frame-wise RAW-Adapter~\citep{cui2024rawadapter}, the recurrent formulation of
\methodname{} avoids abrupt rendering
changes and maintains observations that support successful task execution.

\begin{figure}[t]
\centering
\begin{minipage}[t]{0.235\textwidth}
\centering
\includegraphics[width=\linewidth]{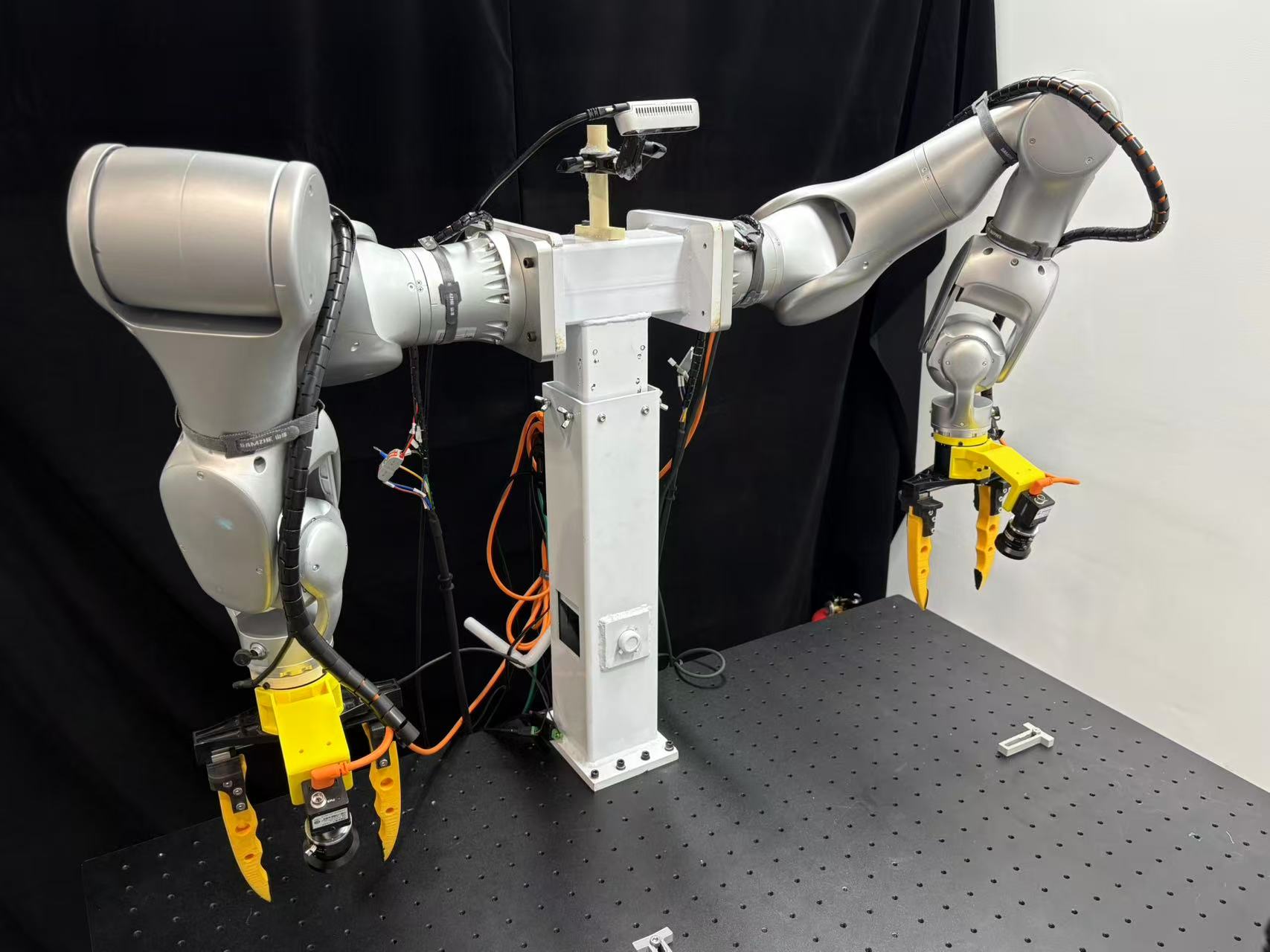}\\[-0.2ex]
\small (a) ROKAE AR5
\end{minipage}\hspace{0.012\textwidth}%
\begin{minipage}[t]{0.235\textwidth}
\centering
\includegraphics[width=\linewidth]{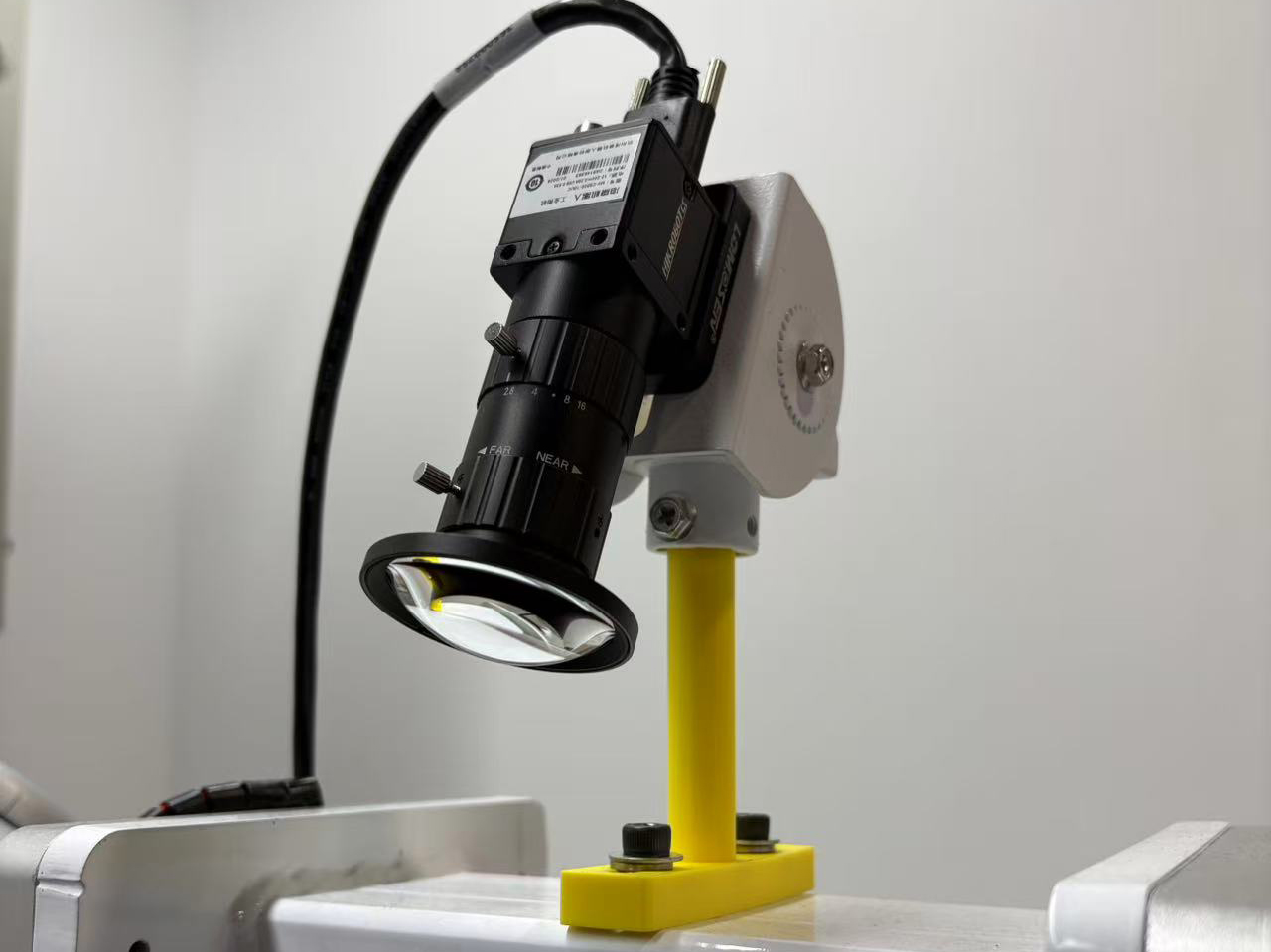}\\[-0.2ex]
\small (b) Agent-view camera
\end{minipage}\hspace{0.012\textwidth}%
\begin{minipage}[t]{0.235\textwidth}
\centering
\includegraphics[width=\linewidth]{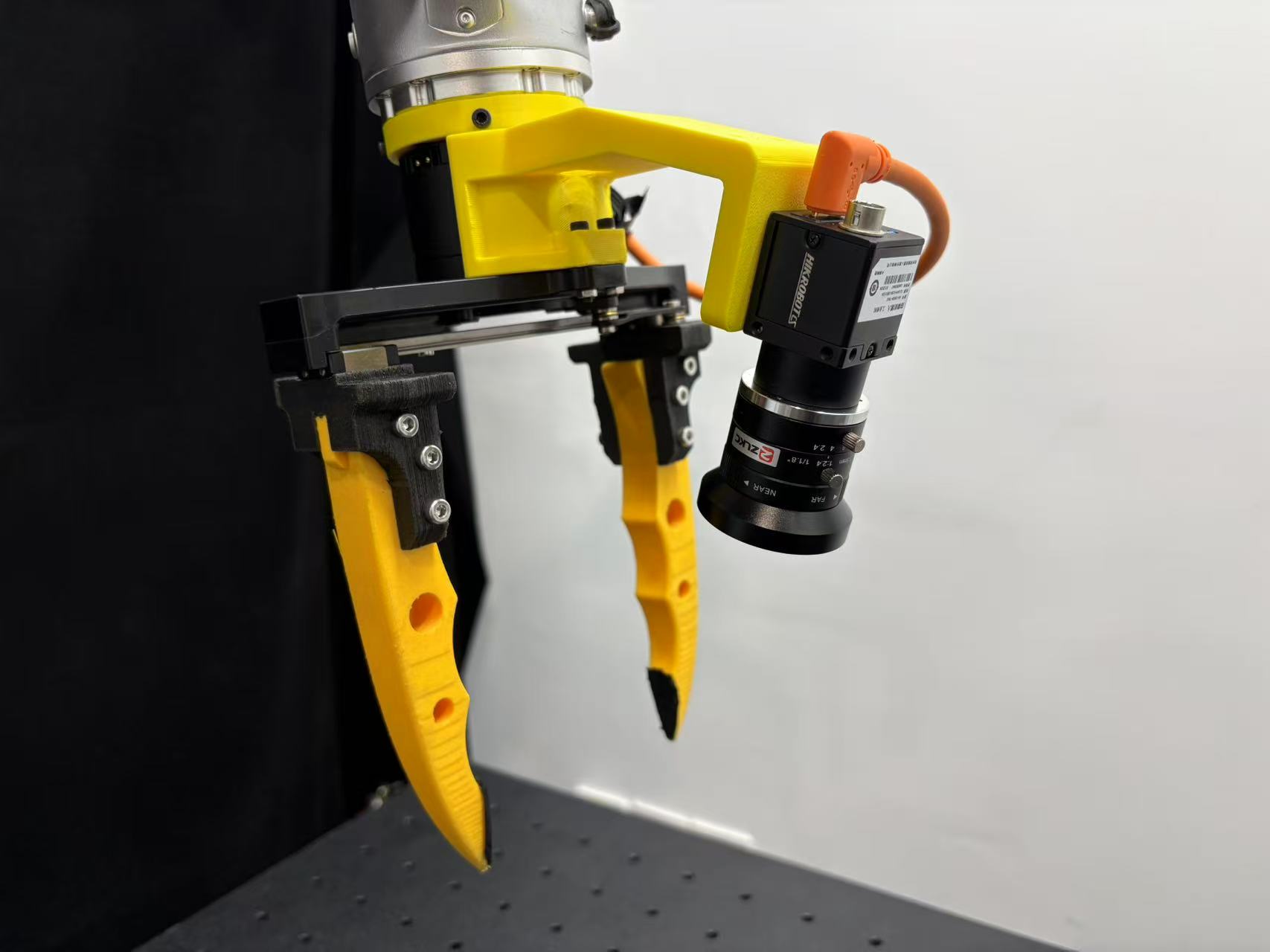}\\[-0.2ex]
\small (c) Left wrist \& camera
\end{minipage}\hspace{0.012\textwidth}%
\begin{minipage}[t]{0.235\textwidth}
\centering
\includegraphics[width=\linewidth]{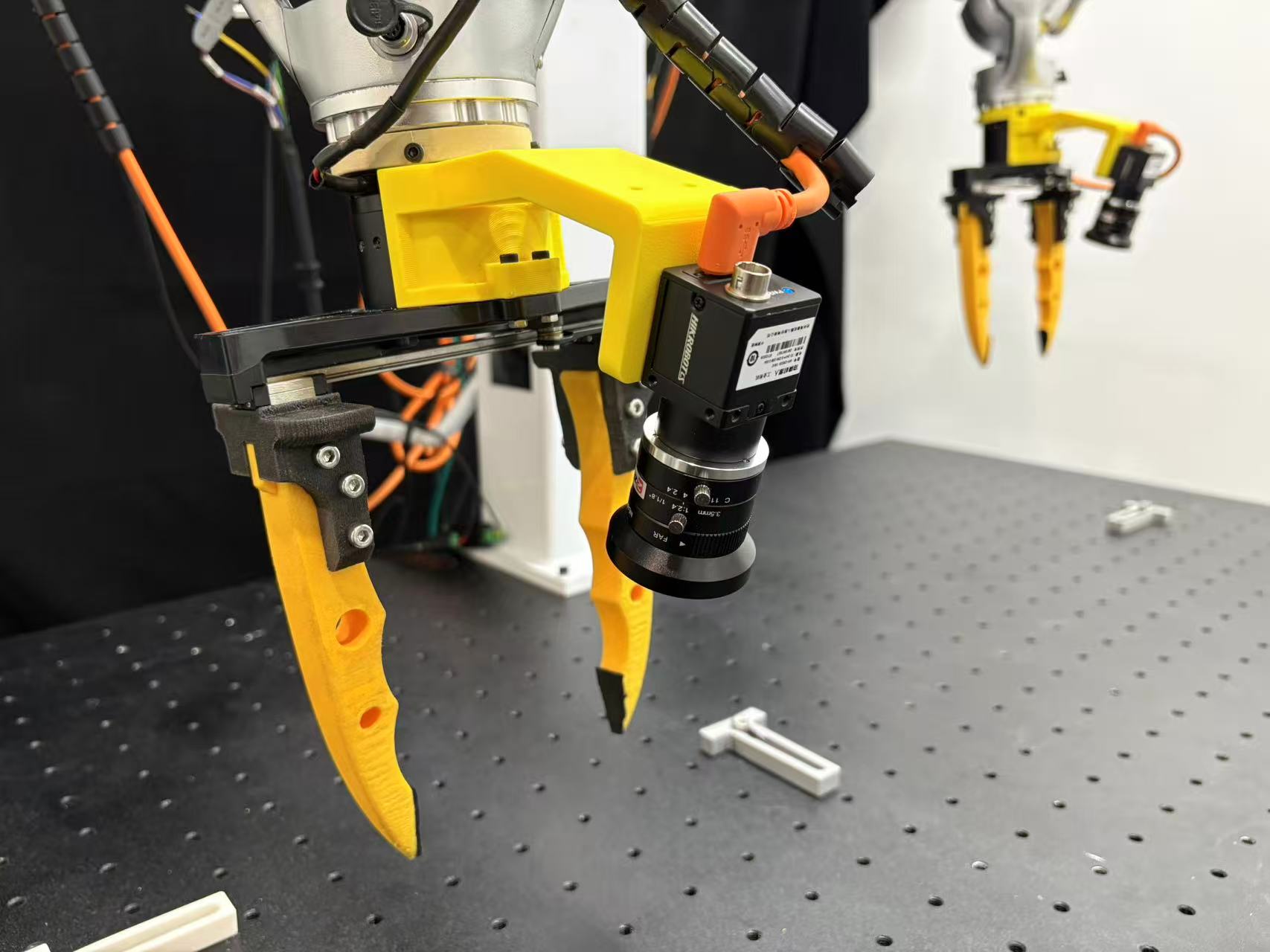}\\[-0.2ex]
\small (d) Right wrist \& camera
\end{minipage}
\caption{Real-world dual-arm platform and camera setup.}
\label{fig:real_world_equipment}
\end{figure}

\begin{figure}[t]
\centering
\begin{minipage}[t]{0.235\textwidth}
\centering
\includegraphics[width=\linewidth]{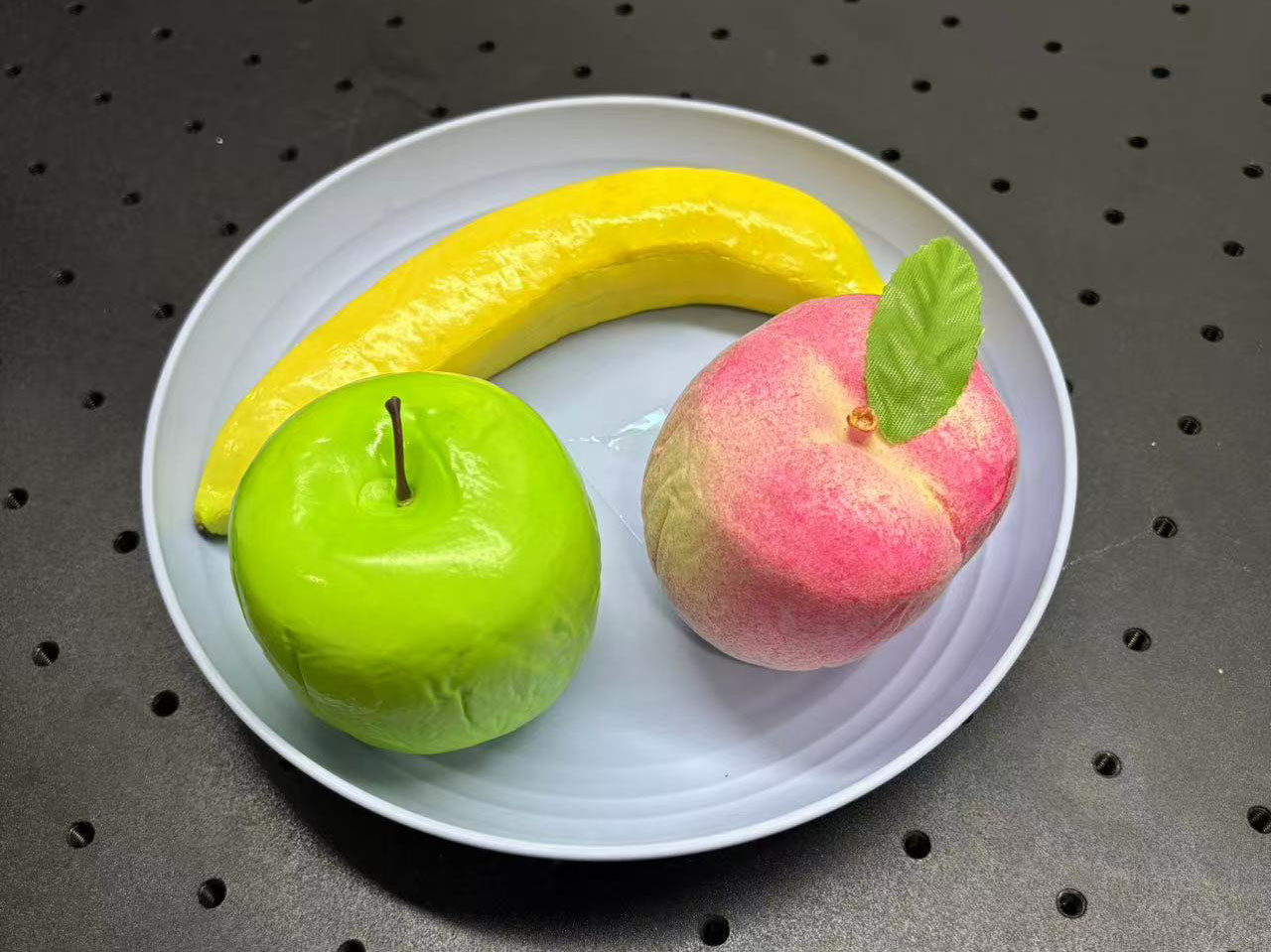}\\[-0.2ex]
\small (a) Fruits
\end{minipage}\hspace{0.012\textwidth}%
\begin{minipage}[t]{0.235\textwidth}
\centering
\includegraphics[width=\linewidth]{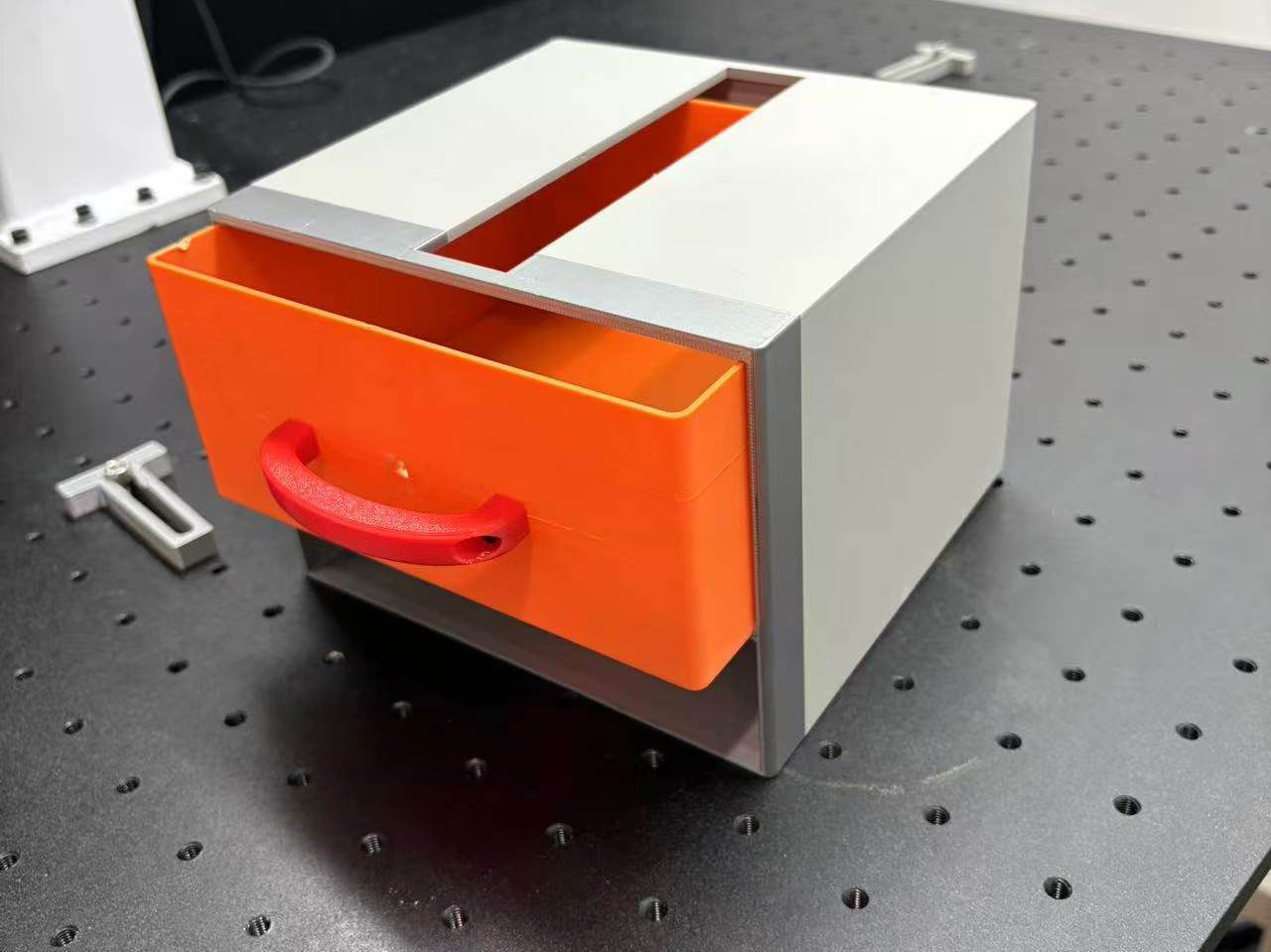}\\[-0.2ex]
\small (b) Drawer
\end{minipage}\hspace{0.012\textwidth}%
\begin{minipage}[t]{0.235\textwidth}
\centering
\includegraphics[width=\linewidth]{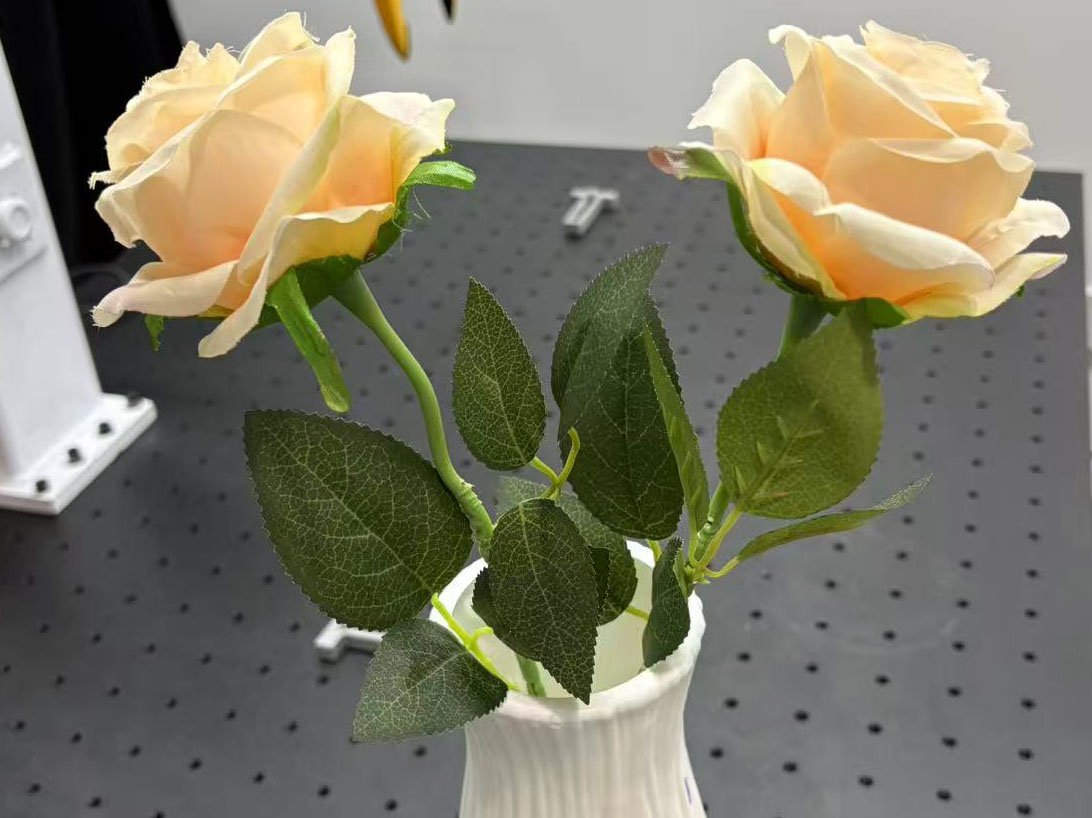}\\[-0.2ex]
\small (c) Flowers
\end{minipage}\hspace{0.012\textwidth}%
\begin{minipage}[t]{0.235\textwidth}
\centering
\includegraphics[width=\linewidth]{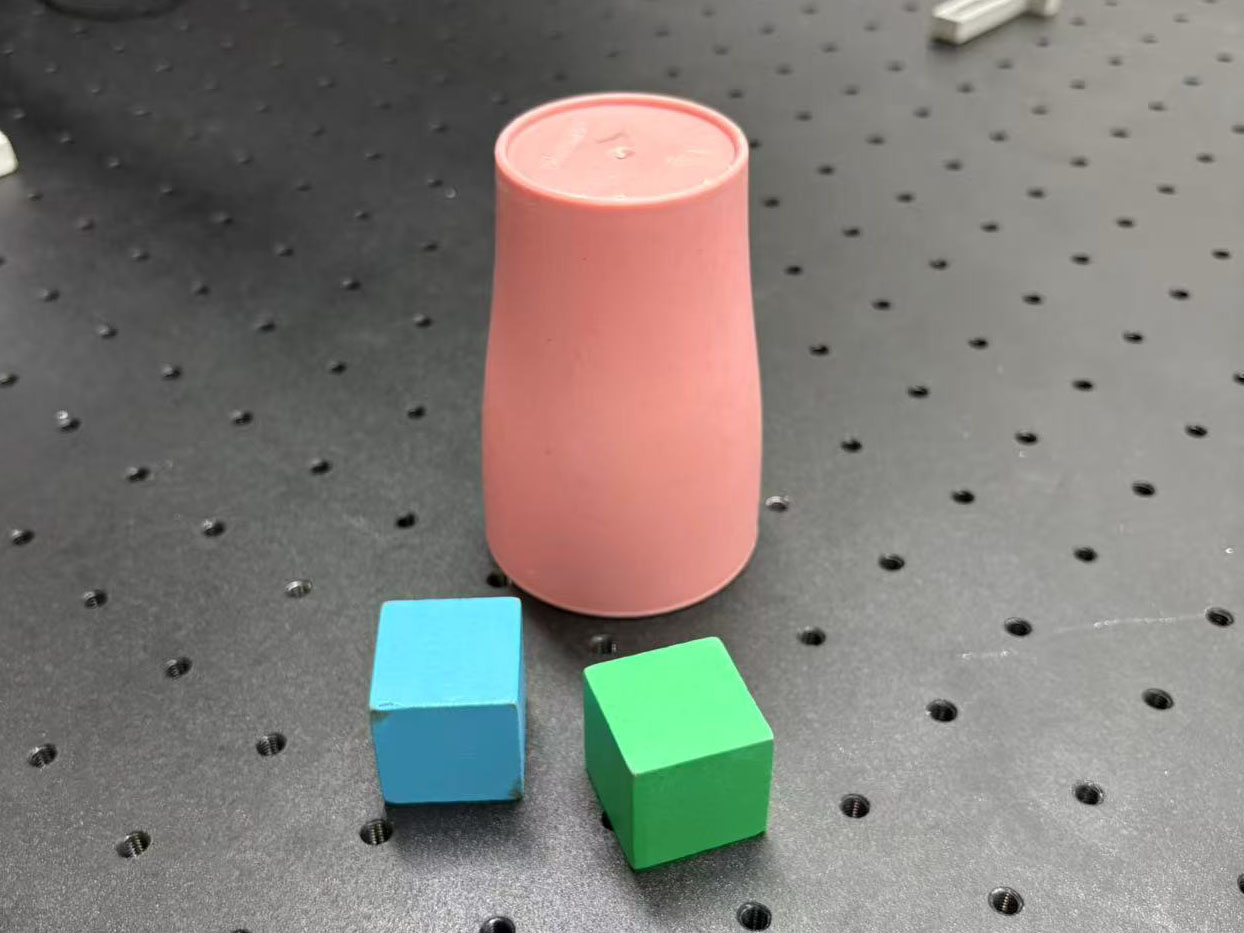}\\[-0.2ex]
\small (d) Cup \& blocks
\end{minipage}
\caption{Objects used in the four real-world tasks.}
\label{fig:real_world_task_objects}
\end{figure}

\section{Real-World Experimental Setup}
\label{app:real_world_experiments}

This section describes the robotic platform and cameras in
Section~\ref{app:realworld_platform}, data acquisition and calibration in
Section~\ref{app:realworld_data}, policy training and deployment in
Section~\ref{app:realworld_deployment}, and qualitative executions in
Section~\ref{app:realworld_qualitative}.

\subsection{Robotic Platform and Camera Configuration}
\label{app:realworld_platform}

Our real-world platform comprises two ROKAE AR5 robot arms and three synchronized
HIKROBOT MV-CS020-10UC industrial cameras, as shown in
Figure~\ref{fig:real_world_equipment}. Each arm is equipped with a Songling
gripper fitted with custom 3D-printed finger extensions to increase its usable
reach for the selected manipulation tasks. The agent-view and two wrist-view streams
are captured by the same camera model. The agent-view camera uses a 4-mm
fixed-focal-length lens at $f/2.8$, while each wrist camera uses a 3.5-mm lens at
$f/4$.

\subsection{Data Acquisition and Camera Calibration}
\label{app:realworld_data}

Figure~\ref{fig:real_world_task_objects} shows the task-specific objects used
for pick-and-place, drug-drawer manipulation, flower placement, and
stack-and-cover. 
The cameras continuously stream 12-bit mosaiced Bayer RAW measurements at 30
FPS. We subtract the camera-specific black level, normalize by the usable
white level, and apply the cameras' default demosaicing to obtain the
three-channel linear RAW representation defined in
Appendix~\ref{app:unprocessing}. We then downsample each view to
$228\times171$. For each task, we collect over 200
trajectories with randomized initial arm configurations, recording all three
views, robot states, and expert actions at 30 FPS. The demonstrations retain
the synchronized linear RAW observations and expert trajectories. This
sensor-linear representation allows exposure and noise to be modified directly
without approximately inverting the nonlinear tone mapping and display
encoding already applied to RGB images.

Because the agent-view and wrist-view cameras use different lenses and
apertures, their uncalibrated RAW streams exhibit different brightness levels.
Before data collection, we perform a three-camera photometric calibration and
increase the agent-view sensor gain so that the three RAW streams have matched
illumination levels under the same scene lighting. This calibration provides a
consistent multi-view input while retaining RAW-domain measurements for the
learned ISP.

\begin{figure}[t]
\centering
\includegraphics[width=0.8\textwidth]{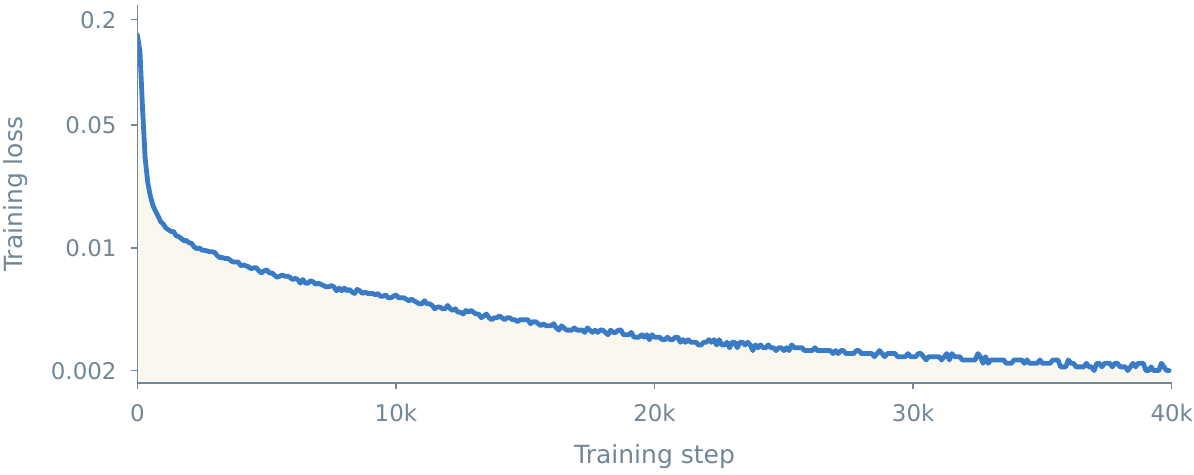}
\caption{Action-loss curves for PI0.5~\citep{physicalintelligence2025pi05} fine-tuning on the four real-world tasks,
using 909 demonstrations and 263,863 frames in total.}
\label{fig:real_world_pi05_finetune}
\end{figure}

\begin{table}[!tp]
\centering
\small
\setlength{\tabcolsep}{5pt}
\begin{tabularx}{\textwidth}{@{}l>{\raggedright\arraybackslash}X@{}}
\toprule
Task & Language prompt \\
\midrule
\texttt{pick\_and\_place}
& Place the green apple, pink object, and yellow banana in the blue bowl. \\
\texttt{drag\_the\_drawer}
& Open the drawer and place the yellow block inside. \\
\texttt{pick\_flower}
& Pick up the two flowers and place them in the white vase. \\
\texttt{stack\_and\_cover}
& Stack the green and blue blocks and cover them with the pink cup. \\
\bottomrule
\end{tabularx}
\caption{Language prompts for the four real-world tasks.}
\label{tab:real_world_language_prompts}
\end{table}

\subsection{Policy and Neural ISP Training and Deployment}
\label{app:realworld_deployment}

We initialize PI-0.5~\citep{physicalintelligence2025pi05} from its pretrained base checkpoint and fully fine-tune
the policy on 909 demonstrations comprising 263,863 frames across the four
real-world tasks under normal illumination. Each training sample contains the current observations from
the three cameras, the 16-dimensional robot state, and the task instruction.
The target is a 16-step chunk of absolute 16-dimensional actions sampled from
the 15 FPS demonstrations. We use four-task quantile statistics to normalize
states and actions and apply state-text dropout with probability 0.2.

Training runs for 40,000 updates on eight NVIDIA B300 GPUs with a global batch
size of 256. We use AdamW with a peak learning rate of $5\times10^{-5}$, a
10,000-step linear warmup, global gradient clipping at 1.0, and an exponential
moving average decay of 0.999. Figure~\ref{fig:real_world_pi05_finetune} shows
the action-loss trajectory over the four-task fine-tuning process.

After obtaining the fine-tuned PI-0.5 policy, we freeze its parameters and
train the neural ISP modules. Each learned ISP is trained on both the recorded
normal-light linear RAW observations and synthetic low-light counterparts. We
generate the low-light inputs directly from the recorded linear RAW images by
applying an exposure offset of $-6$ EV and adding simulated sensor noise. The
synchronized expert trajectories remain unchanged and provide supervision for
the ISP training. Each module is trained for 2,000 steps, yielding one ISP that
is shared across both lighting conditions.

At deployment, all evaluations use a single PI-0.5~\citep{physicalintelligence2025pi05}
checkpoint fine-tuned only under normal illumination. The Default ISP uses the
same fixed camera pipeline under both normal and low light. DarkISP and
\methodname{} each deploy one learned ISP checkpoint under both conditions,
without condition-specific retraining or model switching. This protocol
measures cross-illumination robustness while holding the downstream policy and
ISP model fixed. The policy runs in streaming mode and predicts a 16-step
action chunk from the latest multi-view observation
at 4 Hz while the robot continues executing the previously committed
trajectory. When a new chunk becomes available, a quadratic programming
trajectory optimizer interpolates the remaining previous trajectory toward the
new prediction before execution. This transition accounts for robot inertia
and avoids abrupt changes between consecutive chunks, preserving smooth and
continuous motion. For every ISP method and illumination condition, we conduct
50 independent trials per task. The resulting success rates are reported in
Table~\ref{tab:realworld_illumination}, and the ISP throughput comparison is
given in Table~\ref{tab:realworld_fps}. We condition the policy on the
task-specific language prompts listed in
Table~\ref{tab:real_world_language_prompts}.

\subsection{Qualitative Task Executions}
\label{app:realworld_qualitative}

\begin{figure}[t]
\centering
\includegraphics[width=\textwidth]{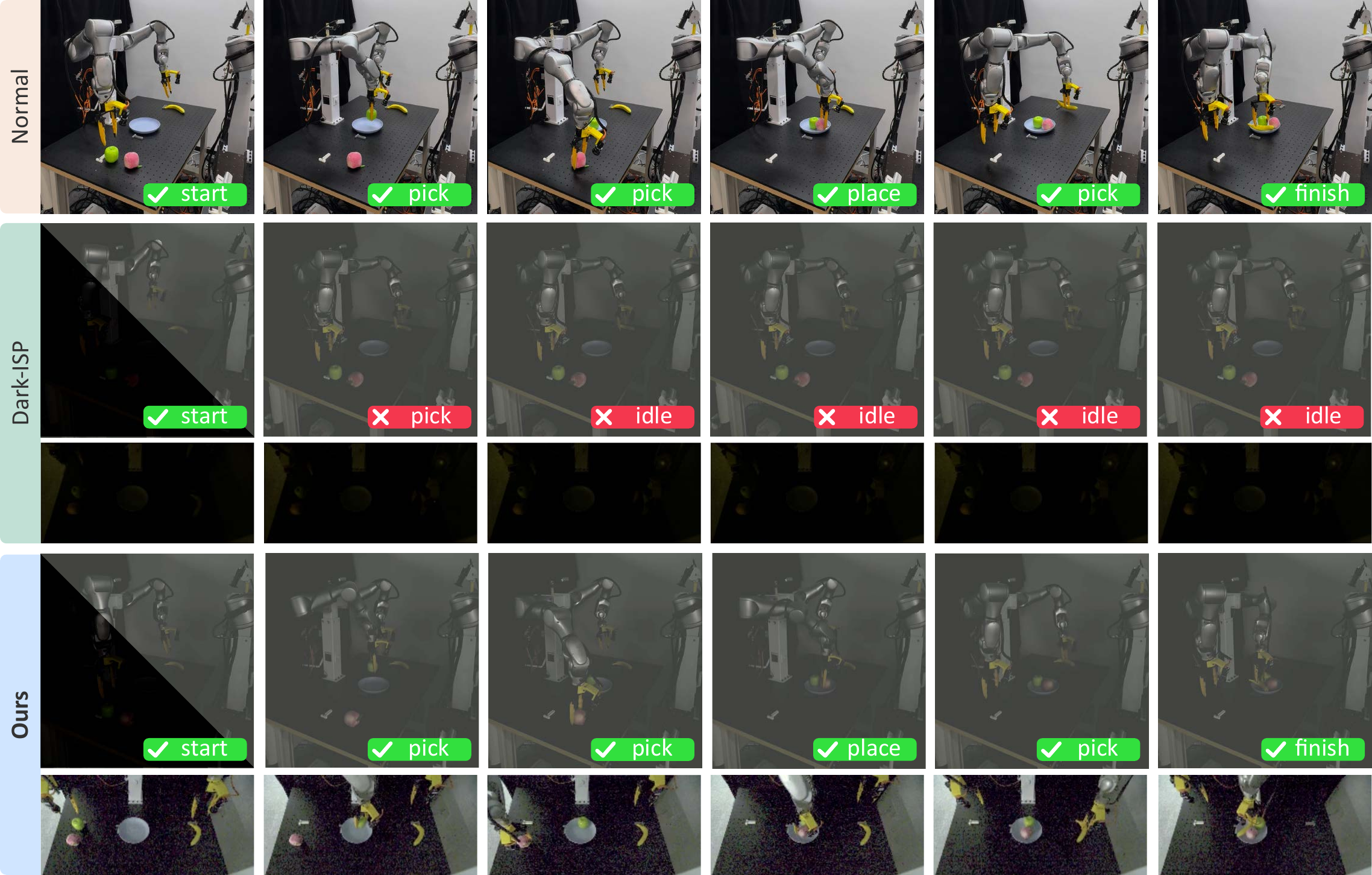}
\caption{\textit{Pick \& place} executions under normal and low-light
illumination. The third-person frames receive postprocessing enhancement only
for visualization and are not provided to the policy. The lower strips show
the rendered agent views produced by Dark-ISP~\citep{guo2025darkisp} and
\methodname{} under low light.}
\label{fig:real_world_pick_place}
\end{figure}

\begin{figure}[t]
\centering
\includegraphics[width=\textwidth]{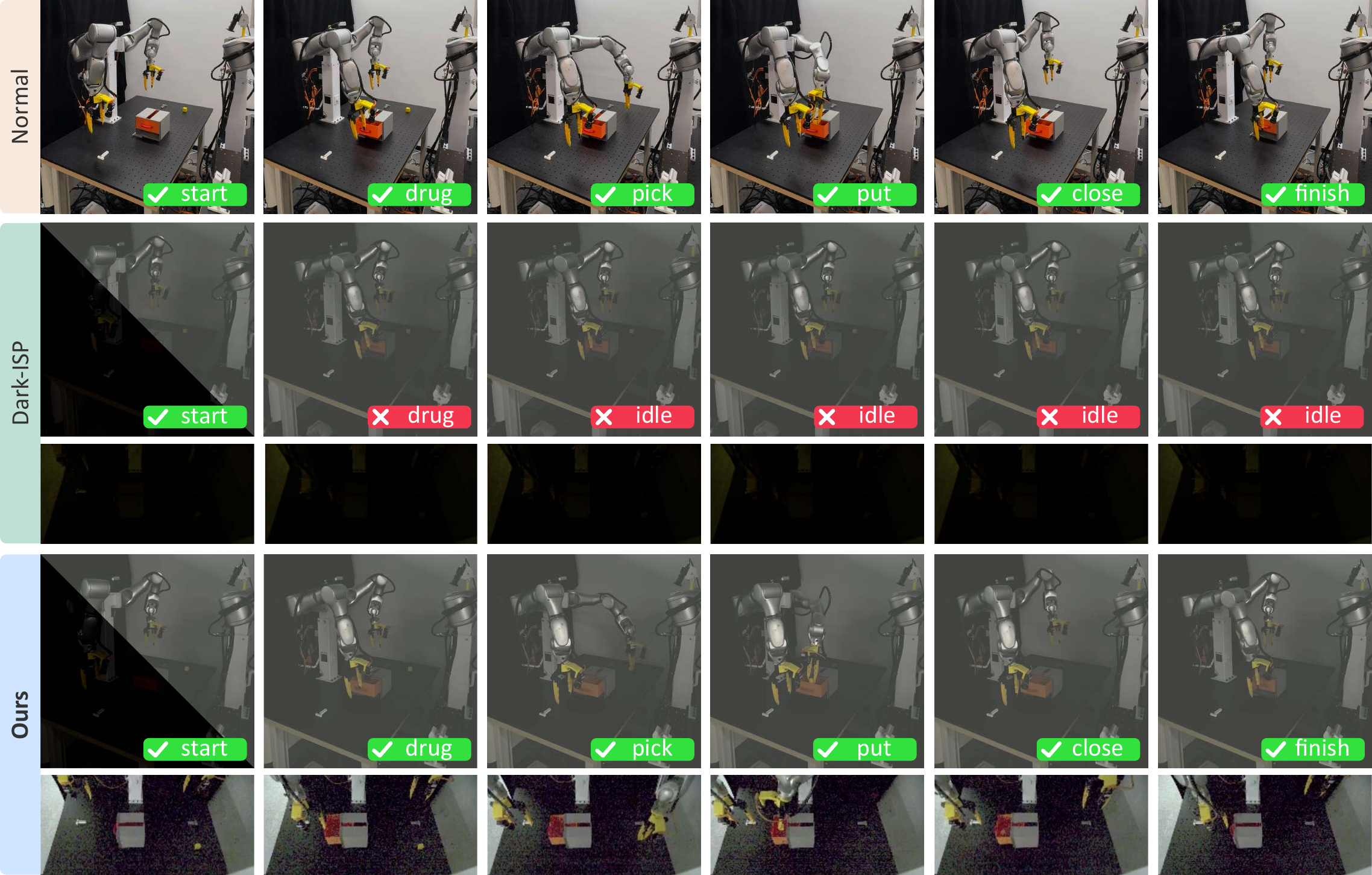}
\caption{\textit{Drug drawer} executions under normal and low-light
illumination. The third-person frames receive postprocessing enhancement only
for visualization and are not provided to the policy. The lower strips show
the rendered agent views produced by Dark-ISP~\citep{guo2025darkisp} and
\methodname{} under low light.}
\label{fig:real_world_drug_drawer}
\end{figure}

\begin{figure}[t]
\centering
\includegraphics[width=\textwidth]{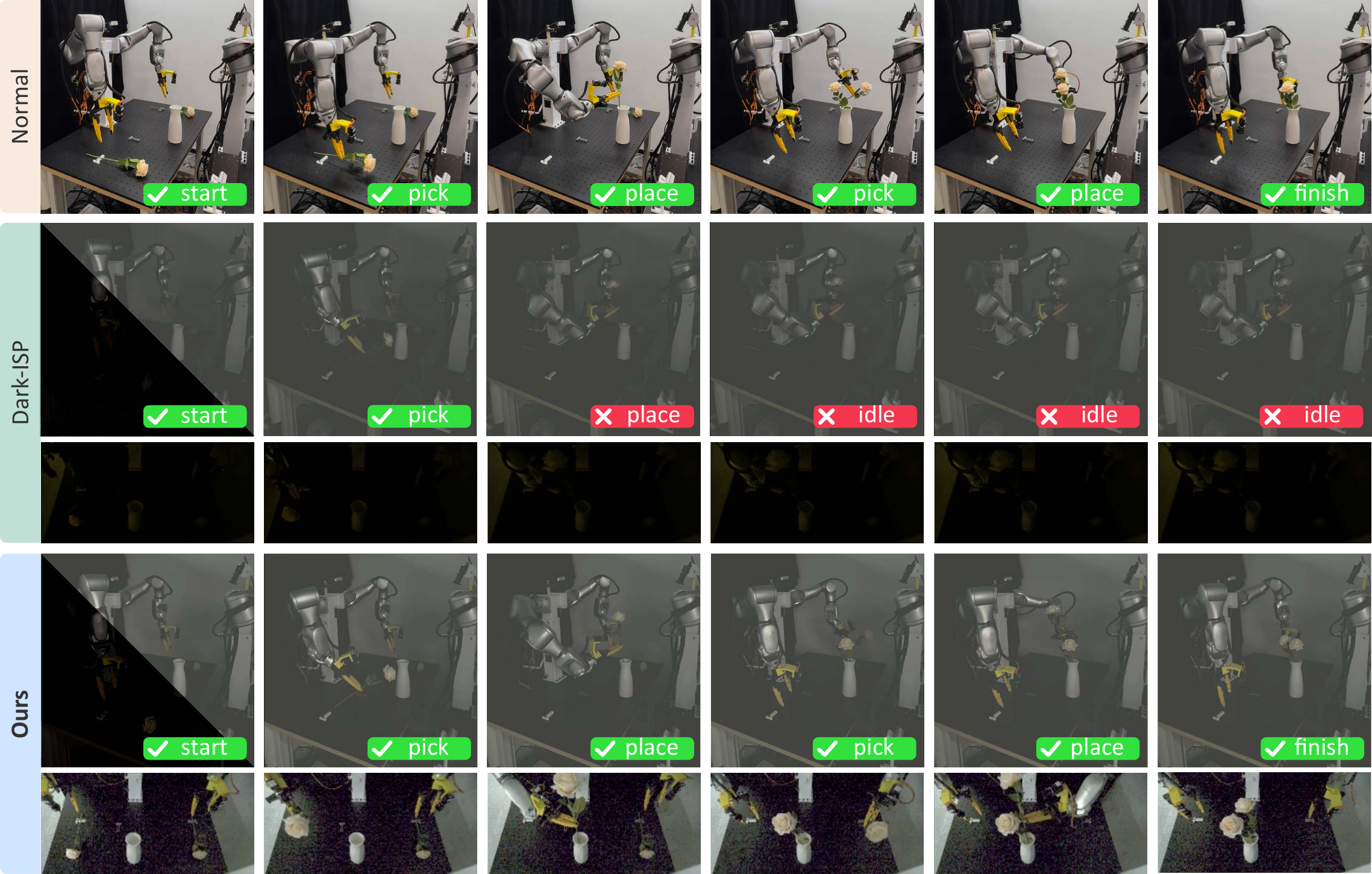}
\caption{\textit{Pick flower} executions under normal and low-light
illumination. The third-person frames receive postprocessing enhancement only
for visualization and are not provided to the policy. The lower strips show
the rendered agent views produced by Dark-ISP~\citep{guo2025darkisp} and
\methodname{} under low light.}
\label{fig:real_world_pick_flower}
\end{figure}

\begin{figure}[t]
\centering
\includegraphics[width=\textwidth]{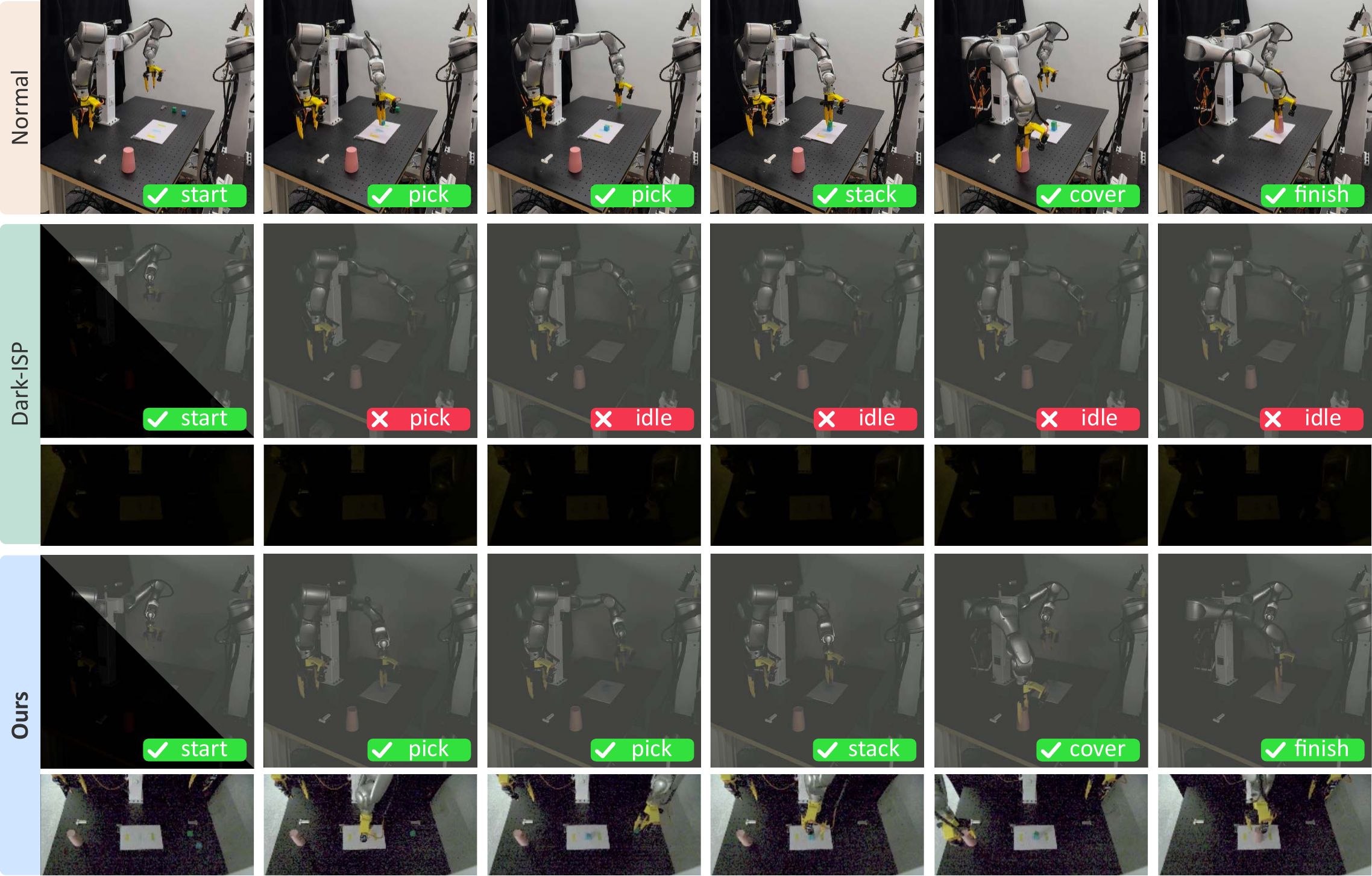}
\caption{\textit{Stack \& cover} executions under normal and low-light
illumination. The third-person frames receive postprocessing enhancement only
for visualization and are not provided to the policy. The lower strips show
the rendered agent views produced by Dark-ISP~\citep{guo2025darkisp} and
\methodname{} under low light.}
\label{fig:real_world_stack_cover}
\end{figure}

Figures~\ref{fig:real_world_pick_place},
\ref{fig:real_world_drug_drawer}, \ref{fig:real_world_pick_flower},
and~\ref{fig:real_world_stack_cover} visualize complete executions of
\textit{pick \& place}, \textit{drug drawer}, \textit{pick flower}, and
\textit{stack \& cover}. The normal illumination sequences demonstrate the
nominal behavior of \methodname{}. Under low light, the Default ISP produces
nearly zero-valued observations and the policy does not initiate motion.
Dark-ISP~\citep{guo2025darkisp} recovers some scene structure, but the policy
stalls during the first stage of each multi-step task. \methodname{} instead
preserves usable agent and wrist observations and completes every sequence.
Low-light execution takes longer than execution under clear normal
illumination because the policy repeatedly observes the scene and refines the
end-effector pose during precise steps such as grasping the blocks and aligning
a flower with the vase opening. Despite these additional adjustments, the
policy completes each fine-grained operation successfully. Videos of all
executions are included in the supplementary material.

\clearpage

\clearpage

\section{WAM Illumination Robustness on LIBERO}
\label{app:wam_libero}

We additionally evaluate \methodname{} with the WM4A-Wan and WM4A-Cosmos
WAM~\citep{starvla2026} backbones in
Table~\ref{tab:libero_illumination_wam}. The results remain uniformly low, and
none of the learned ISP methods produces a substantial or consistent
improvement across the two backbones. Even the strongest averages differ by
only 1.19 percentage points for WM4A-Wan and 0.14 percentage points for
WM4A-Cosmos, which is insufficient to establish a meaningful robustness gain.

This limited effect is consistent with the sensitivity analysis in
Section~\ref{sec:isp_sensitivity}. WAM policies are highly sensitive to
pixel-level appearance changes and operate within a narrower stable imaging
range than the VLA backbones. Consequently, an ISP transformation that improves
visibility may still move the observation away from the image distribution
expected by the generative world model. Optimization is also less direct
because the action objective must propagate through the VAE-based visual
generation pathway before reaching the neural ISP. This long gradient path can
weaken or distort the supervision available for learning task-relevant image
processing. These results suggest that conventional action-driven ISP training
is not sufficient for the evaluated WAM backbones. More direct supervision and
WAM-specific imaging interfaces are left for future study.

\begin{table}[H]
\centering
\caption{Success rates (\%) of ISP modules across five illumination levels on LIBERO using WAM backbones. Avg reports the mean across illumination levels for each backbone.}
\label{tab:libero_illumination_wam}
\small
\setlength{\tabcolsep}{2.5pt}
\renewcommand{\arraystretch}{0.95}
\begin{tabularx}{\textwidth}{@{}ll*{6}{>{\centering\arraybackslash}X}@{}}
\toprule
Illumin. & Backbone & Default & DarkISP & RAM & \makebox[0pt][c]{RAWAdapter} & RAWild & Ours \\
\midrule
\multirow{2}{*}{\shortstack[l]{Extreme\\Lowlight}}
 & WM4A-Wan & 0.00 & 0.05 & 0.00 & 0.00 & 0.00 & 0.75 \\
 & WM4A-Cosmos & 0.00 & 0.00 & 0.00 & 0.00 & 0.00 & 0.20 \\
\midrule
\multirow{2}{*}{Lowlight}
 & WM4A-Wan & 0.05 & 0.05 & 0.05 & 0.00 & 0.00 & 0.55 \\
 & WM4A-Cosmos & 0.05 & 0.00 & 0.00 & 0.00 & 0.00 & 0.75 \\
\midrule
\multirow{2}{*}{Normal}
 & WM4A-Wan & 76.10 & 76.50 & 0.00 & 38.05 & 75.15 & 78.20 \\
 & WM4A-Cosmos & 85.75 & 0.75 & 1.75 & 2.45 & 4.25 & 83.45 \\
\midrule
\multirow{2}{*}{OverExp}
 & WM4A-Wan & 0.50 & 1.45 & 0.00 & 0.00 & 0.55 & 4.55 \\
 & WM4A-Cosmos & 1.90 & 0.05 & 0.00 & 0.00 & 0.35 & 3.50 \\
\midrule
\multirow{2}{*}{\shortstack[l]{Extreme\\OverExp}}
 & WM4A-Wan & 0.00 & 0.45 & 0.00 & 0.00 & 0.00 & 0.40 \\
 & WM4A-Cosmos & 0.00 & 0.00 & 0.00 & 0.00 & 0.30 & 0.50 \\
\midrule
\multirow{2}{*}{Avg}
 & WM4A-Wan & 15.42 & 15.70 & 0.01 & 7.61 & 15.14 & 16.89 \\
 & WM4A-Cosmos & 17.54 & 0.16 & 0.35 & 0.49 & 0.98 & 17.68 \\
\bottomrule
\end{tabularx}
\end{table}

\section{Limitations and Future Studies}
\label{app:limitations}

Our study has several limitations that motivate future work. The simulated
experiments use three-channel linear pseudo-RAW observations rather than
camera-native mosaiced measurements and therefore do not capture the full
diversity of sensor spectral responses, demosaicing artifacts, lens effects,
or proprietary camera pipelines. Extending \benchmarkname{} with calibrated
physical sensors and multiple camera models would provide a more complete
evaluation of RAW-domain manipulation across diverse robotic platforms and
deployment settings.

The current training protocol learns a separate neural ISP for each downstream
policy. Although our evaluation covers multiple VLA backbones, a single ISP
that generalizes across policies, embodiments, and tasks remains an open
challenge. Promising directions include policy-agnostic pretraining,
lightweight online adaptation, and extending adaptive image processing to
world-action models through dedicated objectives and imaging interfaces. The
real-world experiments are also limited to a small set of indoor
manipulation tasks and lighting conditions. Broader evaluations could cover
outdoor scenes, spatially varying illumination, motion blur, weather effects,
and longer task horizons. The severe-haze experiment further relies on
synthesized RGB observations. Evaluating adverse weather with physically
captured RAW data and jointly modeling multiple degradations are important
directions for future study.

\section{Controlled ISP Perturbation Protocol}
\label{app:perturbation_settings}

The two evaluation environments expose different renderer products.
LIBERO~\citep{liu2023libero}
uses MuJoCo RGB8 camera observations resized to the policy input resolution,
whereas RoboTwin~2.0~\citep{chen2025robotwin} uses SAPIEN float RGB textures in $[0,1]$ without an
intermediate RGB8 quantization. In both cases, we apply the unprocessing in
Appendix~\ref{app:unprocessing} with channel gains
$(1.8,1.0,1.7)$ and inverse global gain $1.0$, then quantize the resulting
three-channel linear proxy as
$R_{10}=\operatorname{round}(1023R)$. No Bayer mosaic or camera-specific color
correction matrix is introduced. All policy inputs remain \textit{float32} in $[0,1]$,
and only one perturbation axis is enabled in each evaluation.
Section~\ref{app:perturbation_operators} defines the perturbation operators and
their ISP stages, while Sections~\ref{app:libero_perturbation_settings}
and~\ref{app:robotwin_perturbation_settings} provide the environment-specific
settings.

\subsection{Perturbation Operators and Processing Stages}
\label{app:perturbation_operators}

\paragraph{Exposure.}
Exposure acts on the linear sensor signal before nonlinear display rendering. In a
physical camera, exposure time and analog gain determine the signal collected
or amplified before digitization, while later digital gain adjusts brightness
inside the ISP. Our perturbation uses a single RAW-domain multiplier to capture
their shared effect on signal scale. Negative EV suppresses weak measurements
and makes subsequent quantization more destructive. Positive EV saturates
bright measurements at the sensor range. Applying the reciprocal gain can
restore brightness but cannot recover samples already clipped or collapsed to
the same quantization level. We therefore compare direct and recovered forms
to separate reversible rescaling from irreversible information loss.
For an offset $e$ measured in exposure values (EV), we define
\begin{equation}
    R_e=\operatorname{clip}(2^eR,0,1),
    \qquad
    Y^{\mathrm{exp}}_e=\mathcal{P}_0(R_e),
    \qquad
    Y^{\mathrm{exp}}_0=\mathcal{P}_0(R).
    \label{eq:exposure_perturbation}
\end{equation}
One EV doubles or halves the linear RAW signal, and
$\operatorname{clip}$ acts elementwise.

\paragraph{Sensor Noise.}
Photon arrival follows signal-dependent counting statistics, while readout
electronics add a signal-independent noise floor. Low illumination or short
exposure reduces the number of collected photoelectrons and therefore lowers
the signal-to-noise ratio even when an ISP later restores the mean brightness.
We reproduce this process by attenuating linear RAW with a negative capture EV,
sampling heteroscedastic Gaussian noise whose variance is the sum of shot and
read components, and applying the matching recovery gain. The noise is added
before RAW recovery and the default ISP, so recovery amplifies both the desired
signal and the acquisition noise. This differs from adding a fixed image-space
corruption after rendering.
After forming $R_e$ with Equation~\ref{eq:exposure_perturbation}, we sample
\begin{equation}
    \begin{aligned}
        \varepsilon_u&\sim\operatorname{Normal}\!\left(
            0,\sigma_{\mathrm{shot}}^2 [R_e]_u+\sigma_{\mathrm{read}}^2\right),\\
        \widetilde R_e&=
        \operatorname{clip}\!\left(
            2^{-e}\operatorname{clip}(R_e+\varepsilon,0,1),0,1
        \right),\\
        Y^{\mathrm{noise}}_e&=\mathcal{P}_0(\widetilde R_e).
    \end{aligned}
    \label{eq:sensor_noise_perturbation}
\end{equation}
Here $u$ indexes RAW samples, $\sigma_{\mathrm{shot}}^2$ and
$\sigma_{\mathrm{read}}^2$ control shot and read
noise, and $\varepsilon=\{\varepsilon_u\}_u$. Reciprocal gain restores the
mean exposure while preserving the reduced signal-to-noise ratio.

\paragraph{Chromatic Response.}
Chromatic processing primarily occurs through white balance and color
rendering. White balance applies channel-dependent gains so that the sensor
response to an assumed illuminant becomes neutral. Moving the target white
point therefore changes the absolute tint of the entire rendered observation.
Color rendering then maps the balanced sensor channels into a display color
space and controls relative hue and saturation. To probe this second role
without changing luminance, we rotate and scale the OKLab chroma plane after
the default ISP while preserving lightness. The white-point family measures
sensitivity to illuminant-dependent calibration, whereas the relative-color
family measures sensitivity to relationships among object and scene colors.
Let $\boldsymbol{\nu}_0$ be the reference white point in the $u'v'$ plane and let
$(a_{\mathrm{ok}},b_{\mathrm{ok}})$ be the OKLab chroma coordinates
\citep{ottosson2020oklab} of $Y_0$.
The two chromatic perturbation families are
\begin{equation}
    \begin{aligned}
        \boldsymbol{\nu}_{\delta_{\mathrm{wp}},\theta}
        &=\boldsymbol{\nu}_0+\delta_{\mathrm{wp}}(\cos\theta,\sin\theta),\\
        Y^{\mathrm{wp}}_{\delta_{\mathrm{wp}},\theta}
        &=\mathcal{P}^{\mathrm{wp}}_{\delta_{\mathrm{wp}},\theta}(R),\\
        \begin{bmatrix}a'_{\mathrm{ok}}&b'_{\mathrm{ok}}\end{bmatrix}^{\!\top}
        &=s\operatorname{Rot}(\theta)
        \begin{bmatrix}a_{\mathrm{ok}}&b_{\mathrm{ok}}\end{bmatrix}^{\!\top}.
    \end{aligned}
    \label{eq:chromatic_perturbation}
\end{equation}
Here $\delta_{\mathrm{wp}}$ and $s$ control magnitude, $\theta$ controls direction,
$\operatorname{Rot}(\theta)$ is a two-dimensional rotation, and OKLab lightness is
preserved in the relative-color family.

\paragraph{Tonal Response.}
Tone mapping compresses or expands linear scene luminance before the final
display transfer and determines how contrast is distributed across shadows,
mid-tones, and highlights. We compute BT.709 luminance after restoring the
default linear RGB gains, transform it around the middle-gray pivot in logit
space, and rescale all three channels by the same luminance ratio. Equal
rescaling approximately preserves hue while changing local visibility through
the global tone response. The standard sRGB transfer and smoothstep display
curve are applied afterward. Values below the identity flatten contrast and
values above it increasingly separate luminance levels around the pivot.
For BT.709 luminance $L$ \citep{iturbt709}, the mapping is
\begin{equation}
    L_{c_{\mathrm{tone}}}=\sigma\!\left(
    \operatorname{logit}(\tau)
    +c_{\mathrm{tone}}\bigl(\operatorname{logit}(L)-\operatorname{logit}(\tau)\bigr)
    \right),
    \label{eq:tonal_perturbation}
\end{equation}
where $\tau$ is the luminance pivot and $c_{\mathrm{tone}}$ controls contrast. Linear RGB is
rescaled by $L_{c_{\mathrm{tone}}}/\max(L,\epsilon)$ before the default display operations.
$c_{\mathrm{tone}}=1$ is the identity and $\epsilon>0$ prevents division by zero.

\paragraph{Bit Depth.}
Bit depth represents precision loss at digitization and image transport rather
than a change in photometric rendering. A sensor analog-to-digital converter
quantizes RAW measurements, and a later image representation may introduce a
second quantization after the ISP. Uniform quantization merges nearby values
into the same discrete level and may remove weak edges, texture, or color
differences used by a policy. Quantizing normalized RAW measures precision
loss before white balance, tone mapping, and display encoding. Quantizing
default-ISP RGB measures precision loss in the final rendered observation.
Keeping the transported tensor in \textit{float32} ensures that only the number of
distinct intensity levels changes.
For normalized $U\in[0,1]$, we use
\begin{equation}
    \mathcal{Q}_n(U)
    =\frac{\operatorname{round}((2^n-1)U)}{2^n-1},
    \label{eq:bit_depth_perturbation}
\end{equation}
where $n$ is the number of bits. We apply this operator at the RAW or RGB stage
as specified below.

\subsection{LIBERO Perturbation Settings}
\label{app:libero_perturbation_settings}

Both the agent and wrist cameras use the same parameters. For the
\emph{exposure} axis, exposure is applied before clipping and RAW10 quantization,
with
\begin{equation}
    e\in\{-8,-7,-6,-5,-4,-3,-2,-1,1,2,3,4\}.
\end{equation}
Let $\overline R_{10,e}=\operatorname{round}(1023R_e)/1023$ denote the
normalized exposed pseudo-RAW10 observation. We evaluate four observation
forms. The \emph{raw-direct} form exposes $\overline R_{10,e}$. The
\emph{raw-recovered} form multiplies it by $2^{-e}$ and clips. The
\emph{RGB-direct} form applies the default ISP to $\overline R_{10,e}$. The
\emph{RGB-recovered} form approximately inverts the display mapping, applies
$2^{-e}$, and reapplies gamma and tone. The additional $e\in\{-3,-2,-1\}$
settings are evaluated for the RGB-direct form. The current sweep records
$e=-8$ only for recovered inputs.

For the \emph{noise} axis, the six capture EVs and matching recovery gains are
\begin{equation}
    (e,2^{-e})\in
    \{(-1,2),(-2,4),(-3,8),(-4,16),(-5,32),(-6,64)\}.
\end{equation}
We use $\sigma_{\mathrm{shot}}^2=4.0\times10^{-4}$,
$\sigma_{\mathrm{read}}^2=1.0\times10^{-5}$, and base seed
$1695213855$ in the noise model from
Equation~\ref{eq:sensor_noise_perturbation}.
Static visualizations use frame index $0$, with view indices $0$ and $1$ for
the agent and wrist cameras. During deployment, the episode seed is derived
from the base seed, task index, episode index, and view index. The noise draw
also includes frame and view identity. The resulting draws are reproducible
but differ across episodes, views, and policy frames.
The completed noise experiment evaluates all six policy checkpoints across the
four LIBERO suites and six noise levels. Each model, suite, and level contains
500 episodes from ten tasks with 50 trials per task, producing 72,000 episodes
overall. At every level, noise is sampled after capture attenuation and before
the matching recovery gain and default ISP. Table~\ref{tab:appendix_noise}
reports the full results.

For the \emph{chromatic} axis, white-point radii are
$\delta_{\mathrm{wp}}\in\{0.060,0.075\}$ around the D65 coordinate
$\boldsymbol{\nu}_0=(0.1978,0.4683)$, and relative OKLab chroma scales are
$s\in\{4.5,5.5\}$. Both families use
$\theta\in\{0,45,90,135,180,225,270,315\}^{\circ}$. The displaced white point
is converted through XYZ-to-sRGB, clipped, normalized by luminance, and used
as additional RAW-to-RGB gains, allowing neutral gray to become tinted. The
relative-color transform instead preserves OKLab lightness and adds no chroma
offset.

\begin{figure}[t]
\centering
\includegraphics[width=\textwidth]{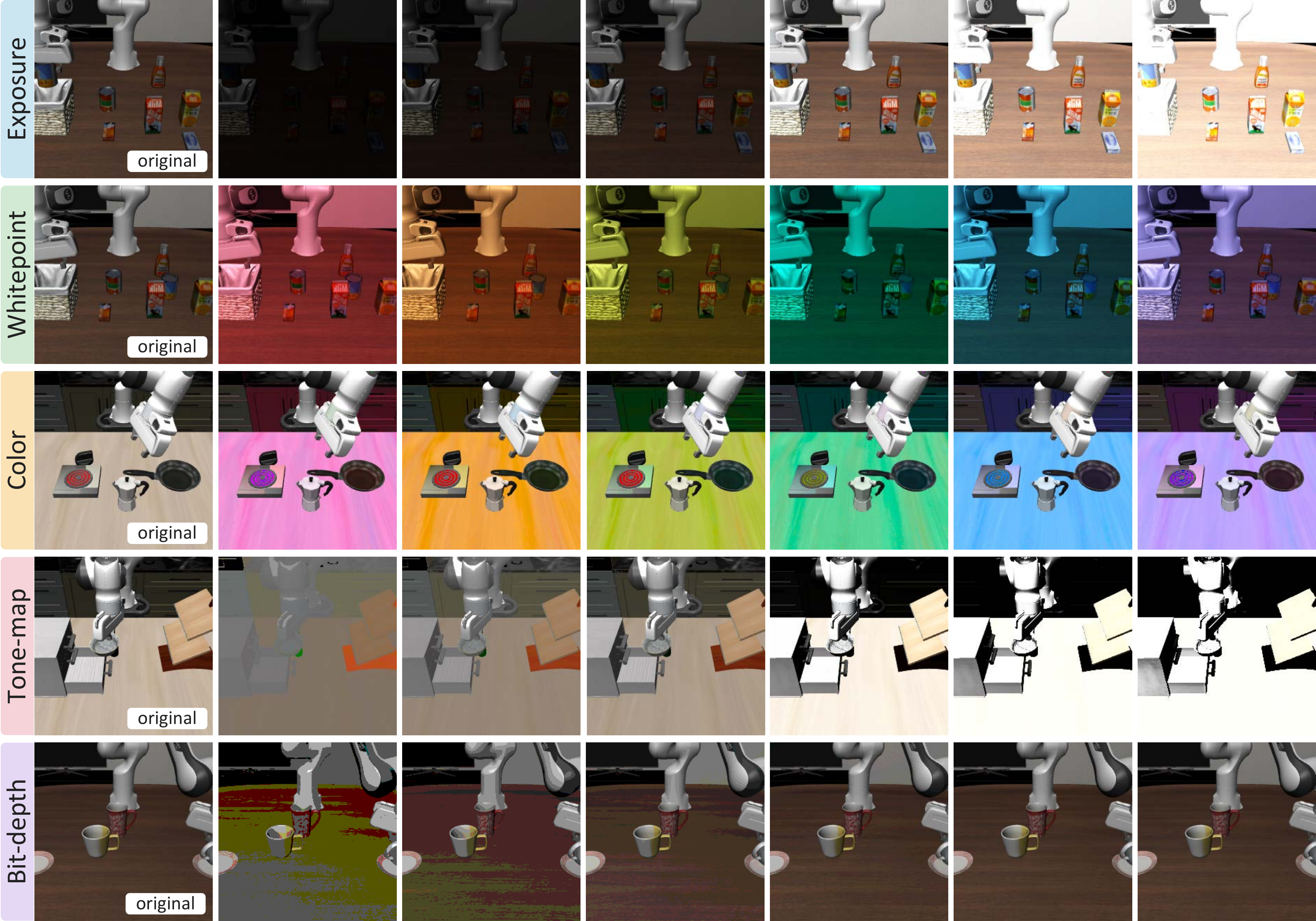}
\caption{Representative ISP perturbations on LIBERO~\citep{liu2023libero}. In each row, the first
column shows the default RGB rendering and the remaining columns show six
perturbed samples. Exposure uses EV offsets
$\{-3,-2,-1,+1,+2,+3\}$. White-point and relative-color response each use six
sampled chromatic transformations. Tone mapping uses six contrast settings,
and bit depth decreases from 7 to 2 bits.}
\label{fig:libero_isp_perturbations}
\end{figure}

For the \emph{tonal} axis, the BT.709 luminance weights are
$(0.2126,0.7152,0.0722)$, the logit pivot is $\tau=0.18$, and
\begin{equation}
    c_{\mathrm{tone}}\in\{0.05,0.3351,0.6458,2.5,5.0,12.0\}.
\end{equation}
The corresponding responses range from very flat to very high contrast.
$c_{\mathrm{tone}}=1$ is the identity and is not rerun as a perturbation. For the
\emph{bit-depth} axis, normalized RAW is evaluated with
$n\in\{2,3,4,5,6\}$, while default-ISP RGB is evaluated with
$n\in\{2,3,4,5,6,7\}$. Quantization is applied before the RGB ISP for RAW and
after rendering for RGB, with exposure, chromatic response, and tone fixed to
their defaults.
Figure~\ref{fig:libero_isp_perturbations} visualizes representative samples
from the four deterministic ISP perturbation families on
LIBERO~\citep{liu2023libero}.

\subsection{RoboTwin~2.0 Perturbation Settings}
\label{app:robotwin_perturbation_settings}

The selected perturbation and parameters are shared by the head, left-wrist,
and right-wrist cameras. For \emph{exposure},
RoboTwin~2.0~\citep{chen2025robotwin} uses the same EV values and four
representations as LIBERO~\citep{liu2023libero}.
Qwen3-OFT~\citep{starvla2026} omits only the $e=-8$ RGB-direct condition.
FastWAM~\citep{starvla2026} evaluates the two RGB representations,
omits both RAW representations, and also omits $e=-8$ RGB-direct. These are
sweep-coverage choices rather than limitations of the transform.

For \emph{noise}, RoboTwin~2.0~\citep{chen2025robotwin} uses the same six capture EVs, recovery gains,
$\sigma_{\mathrm{shot}}^2$, $\sigma_{\mathrm{read}}^2$, and base seed as
LIBERO~\citep{liu2023libero}. A draw uses a stable hash of the
base seed, frame index, view index, and capture EV. Unless explicitly set, the
frame index is a process-local monotonically increasing camera-call index and
the view index defaults to zero, yielding deterministic but distinct draws.
The capture EV, rather than the textual condition label, determines the actual
noise level. These noise coefficients are specific to the ISP perturbation
study and differ from the separate lighting/HDR benchmark coefficients.

\begin{figure}[t]
\centering
\includegraphics[width=\textwidth]{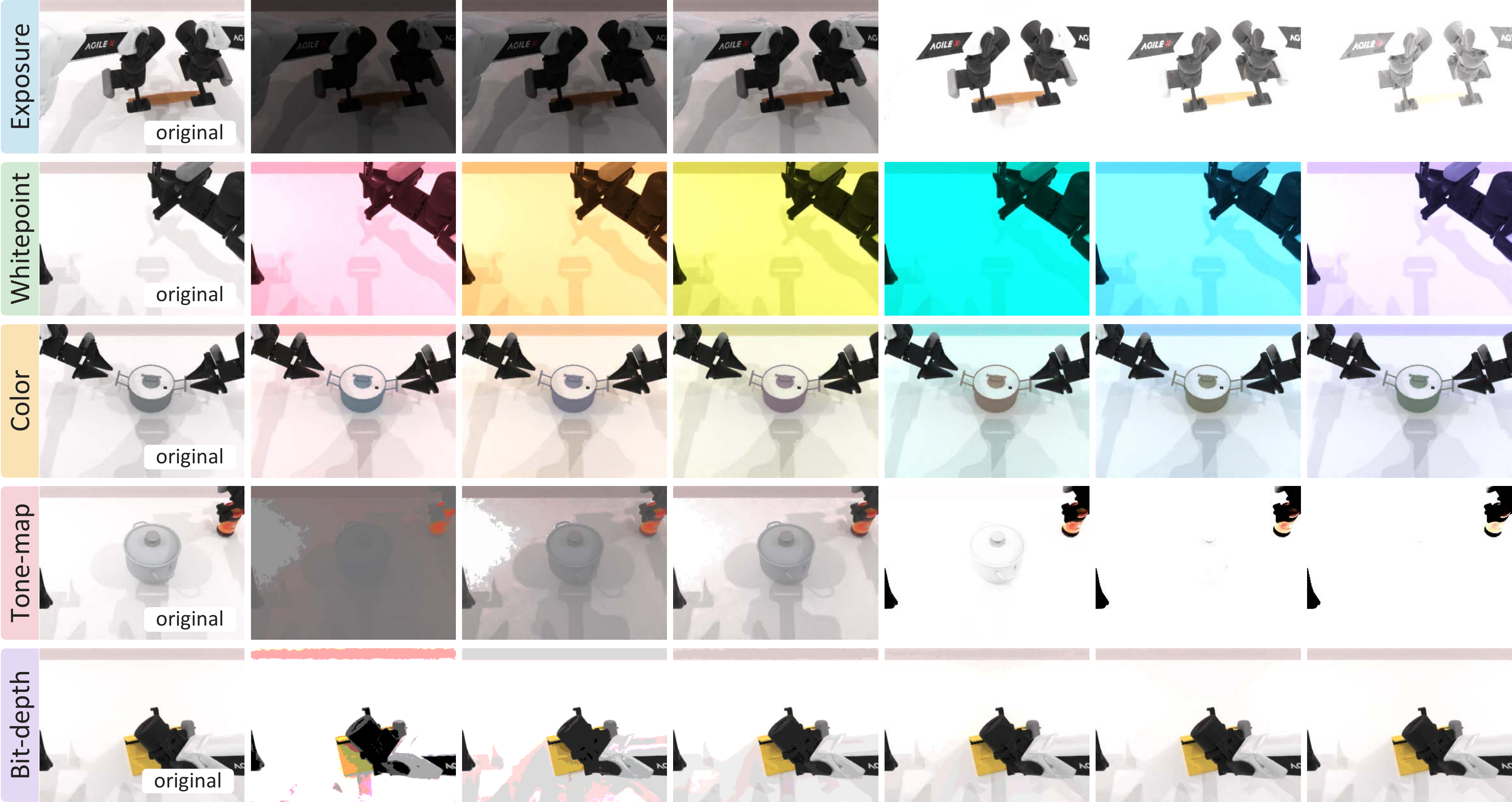}
\caption{Representative ISP perturbations on RoboTwin~2.0~\citep{chen2025robotwin}. In each row, the
first column shows the default RGB rendering and the remaining columns show
six perturbed samples. Exposure uses EV offsets
$\{-3,-2,-1,+1,+2,+3\}$. White-point and relative-color response each use six
sampled chromatic transformations. Tone mapping uses six contrast settings,
and bit depth decreases from 7 to 2 bits.}
\label{fig:robotwin_isp_perturbations}
\end{figure}

For \emph{chromatic response}, RoboTwin~2.0~\citep{chen2025robotwin} uses only
$\delta_{\mathrm{wp}}=0.075$ for white-point shifts and $s=5.0$ for relative OKLab color
shifts, with the same D65 anchor and eight angular directions as
LIBERO~\citep{liu2023libero}. For
\emph{tonal response}, it uses the same luminance definition and pivot but
\begin{equation}
    c_{\mathrm{tone}}\in\{0.05,0.3351,0.6458,1.5,2.5,3.5\},
\end{equation}
with a denominator floor of $10^{-6}$ in the luminance-preserving RGB rescale.
For \emph{bit depth}, $n\in\{2,3,4,5,6\}$ and quantization is applied only to
default-ISP RGB, rather than directly to pseudo-RAW10.
Figure~\ref{fig:robotwin_isp_perturbations} shows the corresponding ISP
perturbation visualizations on RoboTwin~2.0~\citep{chen2025robotwin}.

\section{Full ISP Sensitivity Results}

We report the four deterministic perturbation families and the completed
physical RAW sensor-noise sweep separately. The sensor-noise axis follows the
setup in Appendix~\ref{app:perturbation_settings}.
The columns correspond to
LIBERO-Spatial, LIBERO-Object, LIBERO-Goal, LIBERO-10, and their micro-average
from LIBERO~\citep{liu2023libero}.
All tables use the six VLA checkpoints shared by the perturbation sweeps. The
Original row denotes the default ISP without perturbation. Each non-original
row reports one concrete perturbation setting for the corresponding VLA
baseline. Each baseline block begins with its Original result and is separated
from the next baseline by a midrule.

For LIBERO, Tables~\ref{tab:appendix_exposure},
\ref{tab:appendix_noise}, \ref{tab:appendix_whitepoint},
\ref{tab:appendix_colorrelation}, \ref{tab:appendix_tonal},
and~\ref{tab:appendix_bitdepth} report exposure, sensor noise, white-point,
relative-color, tonal-response, and bit-depth sensitivity, respectively.
For RoboTwin~2.0, the corresponding results are reported in
Tables~\ref{tab:appendix_robotwin_bitdepth},
\ref{tab:appendix_robotwin_exposure}, \ref{tab:appendix_robotwin_noise},
\ref{tab:appendix_robotwin_whitepoint},
\ref{tab:appendix_robotwin_colorrelation},
and~\ref{tab:appendix_robotwin_tonal}. Finally,
Table~\ref{tab:appendix_ev_normalized} reports the effect of explicit RGB- and
RAW-domain exposure normalization on LIBERO.

\newcommand{\chromaswatch}[1]{{\setlength{\fboxsep}{0pt}\colorbox[rgb]{#1}{\rule{5.00ex}{0pt}\rule{0pt}{1.25ex}}}}
\definecolor{exposuredark}{RGB}{119,150,191}
\definecolor{exposurebright}{RGB}{255,239,211}
\newcommand{\exposureswatch}[1]{%
\begingroup
\pgfmathtruncatemacro{\exposuremix}{100*(#1)}%
\colorlet{exposureswatchcolor}{exposurebright!\exposuremix!exposuredark}%
\setlength{\fboxsep}{0pt}%
\colorbox{exposureswatchcolor}{\rule{5.00ex}{0pt}\rule{0pt}{1.25ex}}%
\endgroup}
\newcommand{\originalexposureswatch}{\exposureswatch{0.50}}
\newcommand{\tonegradient}{{\setlength{\fboxsep}{0pt}\colorbox[gray]{1.00}{\rule{0.35ex}{0pt}\rule{0pt}{1.25ex}}\colorbox[gray]{0.96}{\rule{0.35ex}{0pt}\rule{0pt}{1.25ex}}\colorbox[gray]{0.92}{\rule{0.35ex}{0pt}\rule{0pt}{1.25ex}}\colorbox[gray]{0.88}{\rule{0.35ex}{0pt}\rule{0pt}{1.25ex}}\colorbox[gray]{0.83}{\rule{0.35ex}{0pt}\rule{0pt}{1.25ex}}\colorbox[gray]{0.79}{\rule{0.35ex}{0pt}\rule{0pt}{1.25ex}}\colorbox[gray]{0.75}{\rule{0.35ex}{0pt}\rule{0pt}{1.25ex}}\colorbox[gray]{0.71}{\rule{0.35ex}{0pt}\rule{0pt}{1.25ex}}\colorbox[gray]{0.67}{\rule{0.35ex}{0pt}\rule{0pt}{1.25ex}}\colorbox[gray]{0.63}{\rule{0.35ex}{0pt}\rule{0pt}{1.25ex}}\colorbox[gray]{0.58}{\rule{0.35ex}{0pt}\rule{0pt}{1.25ex}}\colorbox[gray]{0.54}{\rule{0.35ex}{0pt}\rule{0pt}{1.25ex}}\colorbox[gray]{0.50}{\rule{0.35ex}{0pt}\rule{0pt}{1.25ex}}\colorbox[gray]{0.46}{\rule{0.35ex}{0pt}\rule{0pt}{1.25ex}}\colorbox[gray]{0.42}{\rule{0.35ex}{0pt}\rule{0pt}{1.25ex}}\colorbox[gray]{0.38}{\rule{0.35ex}{0pt}\rule{0pt}{1.25ex}}}}
\newcommand{\tonebar}{{\setlength{\fboxsep}{0pt}\setlength{\fboxrule}{0.35pt}\fcolorbox{white}{black}{\rule{1.00pt}{1.50ex}}}}
\newcommand{\toneswatch}[1]{{\makebox[5.60ex][l]{\makebox[0pt][l]{\tonegradient}\hspace{#1}\raisebox{-0.12ex}[0pt][0pt]{\tonebar}}}}
\newcommand{\originalswatch}{{\setlength{\fboxsep}{0pt}\fcolorbox{black}{white}{\rule{1.05ex}{0pt}\rule{0pt}{1.05ex}}}}
\newcommand{\originalchromaswatch}{{\setlength{\fboxsep}{0pt}\setlength{\fboxrule}{0.40pt}\fcolorbox[gray]{0.35}[gray]{1.00}{\rule{4.75ex}{0pt}\rule{0pt}{1.05ex}}}}
\newcommand{\bitunit}{{\setlength{\fboxsep}{0pt}\colorbox[gray]{0.35}{\rule{0.65ex}{0pt}\rule{0pt}{0.65ex}}}\hspace{0.25ex}}
\newcommand{\bitswatch}[1]{\raisebox{0.10ex}[0pt][0pt]{\ifcase#1\relax\or\bitunit\or\bitunit\bitunit\or\bitunit\bitunit\bitunit\or\bitunit\bitunit\bitunit\bitunit\or\bitunit\bitunit\bitunit\bitunit\bitunit\or\bitunit\bitunit\bitunit\bitunit\bitunit\bitunit\or\bitunit\bitunit\bitunit\bitunit\bitunit\bitunit\bitunit\or\bitunit\bitunit\bitunit\bitunit\bitunit\bitunit\bitunit\bitunit\fi}}
\definecolor{noisebg}{RGB}{232,232,232}
\definecolor{noiseink}{RGB}{82,82,82}
\definecolor{noiser}{RGB}{214,48,63}
\definecolor{noiseg}{RGB}{24,154,104}
\definecolor{noiseb}{RGB}{42,92,205}
\definecolor{noisec}{RGB}{20,166,188}
\definecolor{noisem}{RGB}{190,45,174}
\definecolor{noisey}{RGB}{220,151,25}
\newcommand{\noiseswatch}[1]{%
\tikz[baseline=-0.55ex,x=0.48ex,y=0.48ex]{%
  \fill[noisebg] (0,0) rectangle (10,2.4);%
  \foreach \x/\y/\c in {0.7/0.4/noiser,2.3/1.7/noiseg,4.1/0.8/noiseb,6.0/1.9/noisec,7.8/0.5/noisem,9.2/1.4/noisey}{\fill[\c] (\x,\y) rectangle ++(0.22,0.22);}%
  \ifnum#1>1 \foreach \x/\y/\c in {1.2/1.2/noiseb,2.9/0.3/noiser,3.5/2.0/noisey,4.9/1.4/noiseg,6.7/0.8/noisem,7.3/1.8/noisec,8.6/0.2/noiseb,9.6/2.0/noiser}{\fill[\c] (\x,\y) rectangle ++(0.23,0.23);}\fi%
  \ifnum#1>2 \foreach \x/\y/\c in {0.3/1.8/noiseg,1.8/0.2/noisem,2.0/2.0/noisec,3.2/1.1/noiseb,4.5/0.2/noisey,5.3/1.0/noiser,5.8/0.4/noiseb,6.5/1.5/noiseg,8.1/1.2/noiser,9.0/0.7/noisec}{\fill[\c] (\x,\y) rectangle ++(0.24,0.24);}\fi%
  \ifnum#1>3 \foreach \x/\y/\c in {0.2/0.7/noiseb,0.9/2.0/noisem,1.5/0.5/noisey,2.6/1.2/noisec,3.8/0.3/noiseg,4.0/1.7/noiser,4.8/2.1/noiseb,5.5/1.6/noisem,6.3/0.2/noisey,7.0/1.1/noiser,7.7/2.1/noiseg,8.4/0.8/noisec,9.4/0.2/noisem,9.7/1.1/noiseb}{\fill[\c] (\x,\y) rectangle ++(0.25,0.25);}\fi%
  \ifnum#1>4 \foreach \x/\y/\c in {0.1/0.1/noiser,0.4/1.2/noisec,0.6/2.1/noisey,1.0/0.8/noiseg,1.4/1.8/noiseb,1.9/0.9/noiser,2.2/0.1/noisec,2.5/2.0/noisem,3.0/0.5/noiseg,3.4/1.6/noisey,3.7/0.8/noiseb,4.3/1.2/noisem,4.6/0.6/noiser,5.0/2.0/noisec,5.6/0.7/noisey,6.1/1.2/noiseb,6.8/2.0/noiser,7.2/0.4/noisem,7.6/1.3/noisey,8.0/0.2/noiseg,8.5/1.8/noiseb,8.9/1.1/noiser,9.3/2.1/noisec,9.6/0.5/noisey}{\fill[\c] (\x,\y) rectangle ++(0.28,0.28);}\fi%
  \draw[noiseink!70,line width=0.25pt] (0,0) rectangle (10,2.4);%
}}
\newcommand{\originalnoiseswatch}{%
\tikz[baseline=-0.55ex,x=0.48ex,y=0.48ex]{%
  \fill[noisebg] (0,0) rectangle (10,2.4);%
  \draw[noiseink!70,line width=0.25pt] (0,0) rectangle (10,2.4);%
}}
\setlength{\LTcapwidth}{\textwidth}
{\small
\setlength{\tabcolsep}{3pt}
\begin{xltabular}{\textwidth}{@{}>{\raggedright\arraybackslash\hsize=1.00\hsize}X >{\raggedright\arraybackslash\hsize=0.95\hsize}X >{\centering\arraybackslash\hsize=0.65\hsize}X *{5}{>{\centering\arraybackslash\hsize=1.08\hsize}X}@{}}
\caption{LIBERO success rates (\%) for direct RGB inputs under exposure perturbations. EV denotes the exposure offset in stops, and the blue-to-cream bar indicates relative illumination.}\label{tab:appendix_exposure}\\
\toprule
Baseline & Exposure setting & Illum. & Spatial & Object & Goal & Long & Overall \\
\midrule
\endfirsthead
\endhead
\endfoot
\bottomrule
\endlastfoot
\multirow{12}{=}{Qwen3-OFT} & Original & \originalexposureswatch & 98.60 & 100.00 & 98.40 & 94.00 & 97.75 \\*
\cmidrule(lr){2-8}\noalign{\nobreak}
 & EV $-7$ & \exposureswatch{0.08} & 0.00 & 0.00 & 0.00 & 0.00 & 0.00 \\*
 & EV $-6$ & \exposureswatch{0.13} & 0.00 & 0.00 & 0.00 & 0.00 & 0.00 \\*
 & EV $-5$ & \exposureswatch{0.20} & 0.00 & 0.00 & 0.40 & 0.00 & 0.10 \\*
 & EV $-4$ & \exposureswatch{0.30} & 10.80 & 3.00 & 9.20 & 0.40 & 5.85 \\*
 & EV $-3$ & \exposureswatch{0.36} & 40.00 & 1.80 & 24.00 & 2.40 & 17.05 \\*
 & EV $-2$ & \exposureswatch{0.42} & 39.60 & 1.40 & 21.60 & 0.80 & 15.85 \\*
 & EV $-1$ & \exposureswatch{0.46} & 38.00 & 1.20 & 26.40 & 1.60 & 16.80 \\*
 & EV $+1$ & \exposureswatch{0.67} & 37.00 & 0.60 & 25.60 & 1.40 & 16.15 \\*
 & EV $+2$ & \exposureswatch{0.78} & 41.60 & 0.80 & 27.00 & 0.40 & 17.45 \\*
 & EV $+3$ & \exposureswatch{0.89} & 2.60 & 5.00 & 40.80 & 45.80 & 23.55 \\*
 & EV $+4$ & \exposureswatch{0.98} & 0.00 & 0.00 & 7.40 & 0.00 & 1.85 \\
\midrule
\multirow{12}{=}{Qwen3-PI} & Original & \originalexposureswatch & 98.40 & 99.00 & 98.00 & 95.20 & 97.65 \\*
\cmidrule(lr){2-8}\noalign{\nobreak}
 & EV $-7$ & \exposureswatch{0.08} & 0.00 & 0.00 & 4.80 & 0.00 & 1.20 \\*
 & EV $-6$ & \exposureswatch{0.13} & 0.00 & 0.00 & 4.20 & 0.00 & 1.05 \\*
 & EV $-5$ & \exposureswatch{0.20} & 2.60 & 3.00 & 0.60 & 0.00 & 1.55 \\*
 & EV $-4$ & \exposureswatch{0.30} & 57.40 & 59.40 & 38.80 & 16.20 & 42.95 \\*
 & EV $-3$ & \exposureswatch{0.36} & 96.60 & 98.40 & 90.00 & 52.20 & 84.30 \\*
 & EV $-2$ & \exposureswatch{0.42} & 98.00 & 98.40 & 97.20 & 86.00 & 94.90 \\*
 & EV $-1$ & \exposureswatch{0.46} & 98.60 & 98.80 & 96.60 & 93.80 & 96.95 \\*
 & EV $+1$ & \exposureswatch{0.67} & 96.80 & 97.80 & 97.40 & 95.80 & 96.95 \\*
 & EV $+2$ & \exposureswatch{0.78} & 96.60 & 91.40 & 95.80 & 65.60 & 87.35 \\*
 & EV $+3$ & \exposureswatch{0.89} & 0.80 & 4.00 & 27.60 & 25.00 & 14.35 \\*
 & EV $+4$ & \exposureswatch{0.98} & 0.00 & 0.20 & 3.80 & 2.60 & 1.65 \\
\midrule
\multirow{12}{=}{WM4A-Cosmo} & Original & \originalexposureswatch & 96.00 & 98.60 & 93.20 & 76.00 & 90.95 \\*
\cmidrule(lr){2-8}\noalign{\nobreak}
 & EV $-7$ & \exposureswatch{0.08} & 0.00 & 0.00 & 0.00 & 0.00 & 0.00 \\*
 & EV $-6$ & \exposureswatch{0.13} & 0.00 & 0.00 & 0.20 & 0.00 & 0.05 \\*
 & EV $-5$ & \exposureswatch{0.20} & 63.00 & 73.00 & 35.80 & 0.20 & 43.00 \\*
 & EV $-4$ & \exposureswatch{0.30} & 84.80 & 97.00 & 67.60 & 19.40 & 67.20 \\*
 & EV $-3$ & \exposureswatch{0.36} & 88.80 & 97.80 & 84.40 & 52.00 & 80.75 \\*
 & EV $-2$ & \exposureswatch{0.42} & 92.20 & 98.20 & 88.20 & 63.80 & 85.60 \\*
 & EV $-1$ & \exposureswatch{0.46} & 94.00 & 98.40 & 88.00 & 69.00 & 87.35 \\*
 & EV $+1$ & \exposureswatch{0.67} & 93.60 & 95.00 & 90.40 & 55.20 & 83.55 \\*
 & EV $+2$ & \exposureswatch{0.78} & 71.00 & 75.80 & 68.00 & 33.80 & 62.15 \\*
 & EV $+3$ & \exposureswatch{0.89} & 0.00 & 0.00 & 1.80 & 2.80 & 1.15 \\*
 & EV $+4$ & \exposureswatch{0.98} & 0.00 & 0.00 & 0.00 & 0.00 & 0.00 \\
\midrule
\multirow{12}{=}{WM4A-Wan} & Original & \originalexposureswatch & 96.40 & 97.80 & 93.20 & 87.60 & 93.75 \\*
\cmidrule(lr){2-8}\noalign{\nobreak}
 & EV $-7$ & \exposureswatch{0.08} & 0.00 & 0.00 & 0.00 & 0.00 & 0.00 \\*
 & EV $-6$ & \exposureswatch{0.13} & 0.00 & 0.00 & 1.20 & 0.00 & 0.30 \\*
 & EV $-5$ & \exposureswatch{0.20} & 2.20 & 0.00 & 9.40 & 0.00 & 2.90 \\*
 & EV $-4$ & \exposureswatch{0.30} & 21.40 & 35.00 & 49.40 & 0.00 & 26.45 \\*
 & EV $-3$ & \exposureswatch{0.36} & 34.40 & 51.60 & 58.00 & 1.80 & 36.45 \\*
 & EV $-2$ & \exposureswatch{0.42} & 30.40 & 41.80 & 45.60 & 8.60 & 31.60 \\*
 & EV $-1$ & \exposureswatch{0.46} & 28.40 & 45.00 & 42.00 & 8.80 & 31.05 \\*
 & EV $+1$ & \exposureswatch{0.67} & 27.80 & 65.40 & 40.60 & 11.40 & 36.30 \\*
 & EV $+2$ & \exposureswatch{0.78} & 0.00 & 34.60 & 13.40 & 0.60 & 12.15 \\*
 & EV $+3$ & \exposureswatch{0.89} & 0.00 & 0.00 & 3.20 & 0.00 & 0.80 \\*
 & EV $+4$ & \exposureswatch{0.98} & 0.00 & 0.00 & 0.00 & 0.00 & 0.00 \\
\midrule\pagebreak[4]
\multirow{12}{=}{PI0} & Original & \originalexposureswatch & 96.60 & 97.80 & 93.00 & 82.00 & 92.35 \\*
\cmidrule(lr){2-8}\noalign{\nobreak}
 & EV $-7$ & \exposureswatch{0.08} & 0.60 & 0.80 & 8.40 & 0.00 & 2.45 \\*
 & EV $-6$ & \exposureswatch{0.13} & 62.80 & 63.00 & 49.20 & 14.60 & 47.40 \\*
 & EV $-5$ & \exposureswatch{0.20} & 94.40 & 94.20 & 82.00 & 41.20 & 77.95 \\*
 & EV $-4$ & \exposureswatch{0.30} & 96.40 & 96.40 & 92.40 & 76.60 & 90.45 \\*
 & EV $-3$ & \exposureswatch{0.36} & 95.40 & 97.40 & 92.40 & 77.80 & 90.75 \\*
 & EV $-2$ & \exposureswatch{0.42} & 96.80 & 97.60 & 95.80 & 81.00 & 92.80 \\*
 & EV $-1$ & \exposureswatch{0.46} & 96.80 & 96.00 & 92.80 & 78.00 & 90.90 \\*
 & EV $+1$ & \exposureswatch{0.67} & 96.80 & 98.20 & 92.40 & 80.20 & 91.90 \\*
 & EV $+2$ & \exposureswatch{0.78} & 95.00 & 90.60 & 92.40 & 64.60 & 85.65 \\*
 & EV $+3$ & \exposureswatch{0.89} & 45.00 & 62.00 & 59.60 & 41.80 & 52.10 \\*
 & EV $+4$ & \exposureswatch{0.98} & 1.40 & 20.20 & 10.40 & 7.60 & 9.90 \\
\midrule
\multirow{12}{=}{PI05} & Original & \originalexposureswatch & 98.00 & 99.40 & 97.60 & 93.20 & 97.05 \\*
\cmidrule(lr){2-8}\noalign{\nobreak}
 & EV $-7$ & \exposureswatch{0.08} & 0.00 & 0.00 & 0.00 & 0.00 & 0.00 \\*
 & EV $-6$ & \exposureswatch{0.13} & 33.80 & 46.00 & 39.00 & 10.40 & 32.30 \\*
 & EV $-5$ & \exposureswatch{0.20} & 95.80 & 97.80 & 86.20 & 64.40 & 86.05 \\*
 & EV $-4$ & \exposureswatch{0.30} & 99.00 & 97.60 & 97.00 & 89.60 & 95.80 \\*
 & EV $-3$ & \exposureswatch{0.36} & 97.40 & 99.00 & 96.20 & 92.80 & 96.35 \\*
 & EV $-2$ & \exposureswatch{0.42} & 97.20 & 98.00 & 97.40 & 91.40 & 96.00 \\*
 & EV $-1$ & \exposureswatch{0.46} & 98.60 & 98.00 & 96.80 & 91.40 & 96.20 \\*
 & EV $+1$ & \exposureswatch{0.67} & 98.00 & 98.20 & 97.20 & 92.00 & 96.35 \\*
 & EV $+2$ & \exposureswatch{0.78} & 98.60 & 99.20 & 97.20 & 80.20 & 93.80 \\*
 & EV $+3$ & \exposureswatch{0.89} & 2.40 & 10.80 & 62.00 & 62.60 & 34.45 \\*
 & EV $+4$ & \exposureswatch{0.98} & 0.00 & 0.20 & 9.20 & 9.20 & 4.65 \\
\end{xltabular}
}

{\small
\setlength{\tabcolsep}{3pt}
\renewcommand{\arraystretch}{0.98}
\begin{xltabular}{\textwidth}{@{}>{\raggedright\arraybackslash\hsize=1.00\hsize}X >{\raggedright\arraybackslash\hsize=0.95\hsize}X >{\centering\arraybackslash\hsize=0.65\hsize}X *{5}{>{\centering\arraybackslash\hsize=1.08\hsize}X}@{}}
\caption{LIBERO success rates (\%) under physical RAW sensor-noise perturbations for each VLA baseline. Capture exposure decreases from EV $-1$ to EV $-6$. Increasing speckle density indicates increasing noise strength.}\label{tab:appendix_noise}\\
\toprule
Baseline & Noise level & Noise & Spatial & Object & Goal & Long & Overall \\
\midrule
\endfirsthead
\endhead
\endfoot
\bottomrule
\endlastfoot
\multirow[c]{7}{=}{Qwen3-OFT} & Original & \originalnoiseswatch & 98.60 & 100.00 & 98.40 & 94.00 & 97.75 \\*
\cmidrule(lr){2-8}
 & EV $-1$ & \noiseswatch{0} & 38.60 & 1.00 & 29.00 & 0.60 & 17.30 \\*
 & EV $-2$ & \noiseswatch{1} & 31.80 & 0.60 & 27.80 & 3.00 & 15.80 \\*
 & EV $-3$ & \noiseswatch{2} & 35.40 & 1.00 & 28.20 & 1.60 & 16.55 \\*
 & EV $-4$ & \noiseswatch{3} & 31.60 & 0.20 & 17.40 & 0.60 & 12.45 \\*
 & EV $-5$ & \noiseswatch{4} & 8.20 & 0.00 & 1.60 & 0.00 & 2.45 \\*
 & EV $-6$ & \noiseswatch{5} & 0.00 & 0.00 & 0.00 & 0.00 & 0.00 \\
\midrule
\multirow[c]{7}{=}{Qwen3-PI} & Original & \originalnoiseswatch & 98.40 & 99.00 & 98.00 & 95.20 & 97.65 \\*
\cmidrule(lr){2-8}
 & EV $-1$ & \noiseswatch{0} & 98.80 & 98.20 & 98.20 & 95.00 & 97.55 \\*
 & EV $-2$ & \noiseswatch{1} & 98.00 & 99.00 & 98.00 & 95.00 & 97.50 \\*
 & EV $-3$ & \noiseswatch{2} & 98.00 & 98.60 & 96.60 & 91.20 & 96.10 \\*
 & EV $-4$ & \noiseswatch{3} & 91.40 & 83.60 & 74.00 & 48.20 & 74.30 \\*
 & EV $-5$ & \noiseswatch{4} & 43.80 & 35.40 & 18.20 & 15.00 & 28.10 \\*
 & EV $-6$ & \noiseswatch{5} & 2.40 & 1.40 & 4.80 & 0.00 & 2.15 \\
\midrule
\multirow[c]{7}{=}{WM4A-Cosmo} & Original & \originalnoiseswatch & 96.00 & 98.60 & 93.20 & 76.00 & 90.95 \\*
\cmidrule(lr){2-8}
 & EV $-1$ & \noiseswatch{0} & 80.20 & 95.00 & 79.80 & 49.00 & 76.00 \\*
 & EV $-2$ & \noiseswatch{1} & 53.80 & 81.00 & 66.80 & 23.80 & 56.35 \\*
 & EV $-3$ & \noiseswatch{2} & 10.00 & 6.00 & 24.00 & 7.60 & 11.90 \\*
 & EV $-4$ & \noiseswatch{3} & 0.00 & 0.00 & 0.00 & 0.80 & 0.20 \\*
 & EV $-5$ & \noiseswatch{4} & 0.00 & 0.00 & 0.00 & 0.00 & 0.00 \\*
 & EV $-6$ & \noiseswatch{5} & 0.00 & 0.00 & 0.00 & 0.00 & 0.00 \\
\midrule
\multirow[c]{7}{=}{WM4A-Wan} & Original & \originalnoiseswatch & 96.40 & 97.80 & 93.20 & 87.60 & 93.75 \\*
\cmidrule(lr){2-8}
 & EV $-1$ & \noiseswatch{0} & 0.00 & 0.00 & 3.00 & 0.00 & 0.75 \\*
 & EV $-2$ & \noiseswatch{1} & 0.00 & 0.00 & 2.00 & 0.00 & 0.50 \\*
 & EV $-3$ & \noiseswatch{2} & 0.00 & 0.00 & 0.00 & 0.00 & 0.00 \\*
 & EV $-4$ & \noiseswatch{3} & 0.00 & 0.00 & 0.00 & 0.00 & 0.00 \\*
 & EV $-5$ & \noiseswatch{4} & 0.00 & 0.00 & 0.00 & 0.00 & 0.00 \\*
 & EV $-6$ & \noiseswatch{5} & 0.00 & 0.00 & 0.00 & 0.00 & 0.00 \\
\midrule
\multirow[c]{7}{=}{PI0} & Original & \originalnoiseswatch & 96.60 & 97.80 & 93.00 & 82.00 & 92.35 \\*
\cmidrule(lr){2-8}
 & EV $-1$ & \noiseswatch{0} & 97.40 & 96.80 & 93.20 & 81.20 & 92.15 \\*
 & EV $-2$ & \noiseswatch{1} & 95.80 & 97.80 & 91.60 & 78.20 & 90.85 \\*
 & EV $-3$ & \noiseswatch{2} & 96.00 & 97.60 & 89.60 & 71.60 & 88.70 \\*
 & EV $-4$ & \noiseswatch{3} & 94.80 & 94.20 & 80.80 & 56.20 & 81.50 \\*
 & EV $-5$ & \noiseswatch{4} & 43.60 & 70.20 & 35.60 & 16.00 & 41.35 \\*
 & EV $-6$ & \noiseswatch{5} & 0.20 & 0.40 & 9.60 & 0.00 & 2.55 \\
\midrule
\multirow[c]{7}{=}{PI05} & Original & \originalnoiseswatch & 98.00 & 99.40 & 97.60 & 93.20 & 97.05 \\*
\cmidrule(lr){2-8}
 & EV $-1$ & \noiseswatch{0} & 97.60 & 98.80 & 97.40 & 93.80 & 96.90 \\*
 & EV $-2$ & \noiseswatch{1} & 98.20 & 98.80 & 97.00 & 93.80 & 96.95 \\*
 & EV $-3$ & \noiseswatch{2} & 99.00 & 98.80 & 96.40 & 92.80 & 96.75 \\*
 & EV $-4$ & \noiseswatch{3} & 94.60 & 100.00 & 76.80 & 61.20 & 83.15 \\*
 & EV $-5$ & \noiseswatch{4} & 4.20 & 22.20 & 10.20 & 2.20 & 9.70 \\*
 & EV $-6$ & \noiseswatch{5} & 0.00 & 0.00 & 0.40 & 0.00 & 0.10 \\
\end{xltabular}
}

{\small
\setlength{\tabcolsep}{2.5pt}
\renewcommand{\arraystretch}{0.95}
\begin{xltabular}{\textwidth}{@{}>{\raggedright\arraybackslash\hsize=0.90\hsize}X >{\raggedright\arraybackslash\hsize=1.25\hsize}X >{\centering\arraybackslash\hsize=0.55\hsize}X *{5}{>{\centering\arraybackslash\hsize=1.06\hsize}X}@{}}
\caption{LIBERO success rates (\%) under white-point perturbations for each VLA baseline. The radius $\rho$ denotes displacement from D65 in the $u'v'$ chromaticity plane, and each colored bar represents one shift direction. Avg. averages the eight evaluated directions.}\label{tab:appendix_whitepoint}\\
\toprule
Baseline & White-point setting & Color & Spatial & Object & Goal & Long & Overall \\
\midrule
\endfirsthead
\endhead
\endfoot
\bottomrule
\endlastfoot
\multirow{20}{=}{Qwen3-OFT} & Original & \originalchromaswatch & 98.60 & 100.00 & 98.40 & 94.00 & 97.75 \\*
\cmidrule(lr){2-8}
 & \multirow{8}{=}{\shortstack[l]{$\rho=0.06$\\(moderate)}} & \chromaswatch{1.000,0.635,0.710} & 39.20 & 1.00 & 29.00 & 1.80 & 17.75 \\*
 &  & \chromaswatch{1.000,0.757,0.537} & 36.20 & 2.00 & 26.40 & 1.20 & 16.45 \\*
 &  & \chromaswatch{0.965,1.000,0.533} & 36.40 & 1.60 & 26.00 & 1.80 & 16.45 \\*
 &  & \chromaswatch{0.576,1.000,0.678} & 37.00 & 1.60 & 27.00 & 1.80 & 16.85 \\*
 &  & \chromaswatch{0.122,1.000,0.929} & 31.00 & 0.80 & 26.60 & 0.80 & 14.80 \\*
 &  & \chromaswatch{0.412,0.839,1.000} & 37.80 & 1.80 & 25.80 & 1.20 & 16.65 \\*
 &  & \chromaswatch{0.745,0.710,1.000} & 33.60 & 0.80 & 28.40 & 1.20 & 16.00 \\*
 &  & \chromaswatch{1.000,0.667,0.953} & 33.00 & 0.40 & 25.00 & 1.60 & 15.00 \\*
 &  & \textit{Avg.} & 35.52 & 1.25 & 26.77 & 1.43 & 16.24 \\*
\cmidrule(lr){2-8}
 & \multirow{8}{=}{\shortstack[l]{$\rho=0.08$\\(strong)}} & \chromaswatch{1.000,0.556,0.653} & 38.00 & 1.60 & 28.40 & 2.40 & 17.60 \\*
 &  & \chromaswatch{1.000,0.705,0.390} & 35.00 & 2.20 & 24.20 & 2.60 & 16.00 \\*
 &  & \chromaswatch{0.960,1.000,0.277} & 36.20 & 1.80 & 26.40 & 1.80 & 16.55 \\*
 &  & \chromaswatch{0.399,1.000,0.569} & 36.00 & 1.20 & 29.40 & 1.40 & 17.00 \\*
 &  & \chromaswatch{0.000,1.000,0.909} & 29.80 & 1.40 & 26.00 & 0.80 & 14.50 \\*
 &  & \chromaswatch{0.025,0.803,1.000} & 33.20 & 1.40 & 25.60 & 1.20 & 15.35 \\*
 &  & \chromaswatch{0.688,0.642,1.000} & 35.00 & 1.60 & 27.20 & 1.00 & 16.20 \\*
 &  & \chromaswatch{1.000,0.578,0.942} & 36.80 & 1.80 & 26.80 & 1.40 & 16.70 \\*
 &  & \textit{Avg.} & 35.00 & 1.62 & 26.75 & 1.57 & 16.24 \\*
\cmidrule(lr){2-8}
 & \textbf{Overall avg.} &  & 35.26 & 1.44 & 26.76 & 1.50 & 16.24 \\
\midrule
\multirow{20}{=}{Qwen3-PI} & Original & \originalchromaswatch & 98.40 & 99.00 & 98.00 & 95.20 & 97.65 \\*
\cmidrule(lr){2-8}
 & \multirow{8}{=}{\shortstack[l]{$\rho=0.06$\\(moderate)}} & \chromaswatch{1.000,0.635,0.710} & 97.40 & 98.00 & 98.20 & 95.40 & 97.25 \\*
 &  & \chromaswatch{1.000,0.757,0.537} & 96.40 & 99.60 & 98.40 & 95.00 & 97.35 \\*
 &  & \chromaswatch{0.965,1.000,0.533} & 98.60 & 99.40 & 98.00 & 95.00 & 97.75 \\*
 &  & \chromaswatch{0.576,1.000,0.678} & 98.80 & 97.00 & 98.40 & 95.20 & 97.35 \\*
 &  & \chromaswatch{0.122,1.000,0.929} & 97.80 & 98.80 & 97.60 & 92.20 & 96.60 \\*
 &  & \chromaswatch{0.412,0.839,1.000} & 97.60 & 98.60 & 99.00 & 94.80 & 97.50 \\*
 &  & \chromaswatch{0.745,0.710,1.000} & 98.80 & 98.20 & 98.60 & 94.80 & 97.60 \\*
 &  & \chromaswatch{1.000,0.667,0.953} & 98.20 & 97.60 & 98.60 & 94.80 & 97.30 \\*
 &  & \textit{Avg.} & 97.95 & 98.40 & 98.35 & 94.65 & 97.34 \\*
\cmidrule(lr){2-8}
 & \multirow{8}{=}{\shortstack[l]{$\rho=0.08$\\(strong)}} & \chromaswatch{1.000,0.556,0.653} & 98.60 & 98.20 & 98.40 & 94.80 & 97.50 \\*
 &  & \chromaswatch{1.000,0.705,0.390} & 97.80 & 99.20 & 97.60 & 94.20 & 97.20 \\*
 &  & \chromaswatch{0.960,1.000,0.277} & 97.80 & 98.20 & 96.00 & 94.80 & 96.70 \\*
 &  & \chromaswatch{0.399,1.000,0.569} & 98.40 & 98.20 & 98.60 & 90.80 & 96.50 \\*
 &  & \chromaswatch{0.000,1.000,0.909} & 97.40 & 98.40 & 97.40 & 93.00 & 96.55 \\*
 &  & \chromaswatch{0.025,0.803,1.000} & 98.80 & 97.20 & 97.80 & 92.40 & 96.55 \\*
 &  & \chromaswatch{0.688,0.642,1.000} & 98.40 & 98.00 & 98.40 & 95.40 & 97.55 \\*
 &  & \chromaswatch{1.000,0.578,0.942} & 98.40 & 98.20 & 98.00 & 94.60 & 97.30 \\*
 &  & \textit{Avg.} & 98.20 & 98.20 & 97.78 & 93.75 & 96.98 \\*
\cmidrule(lr){2-8}
 & \textbf{Overall avg.} &  & 98.08 & 98.30 & 98.06 & 94.20 & 97.16 \\
\midrule
\multirow{20}{=}{WM4A-Cosmo} & Original & \originalchromaswatch & 96.00 & 98.60 & 93.20 & 76.00 & 90.95 \\*
\cmidrule(lr){2-8}
 & \multirow{8}{=}{\shortstack[l]{$\rho=0.06$\\(moderate)}} & \chromaswatch{1.000,0.635,0.710} & 97.00 & 98.60 & 94.40 & 69.40 & 89.85 \\*
 &  & \chromaswatch{1.000,0.757,0.537} & 94.00 & 95.00 & 89.60 & 54.00 & 83.15 \\*
 &  & \chromaswatch{0.965,1.000,0.533} & 37.60 & 75.60 & 57.20 & 19.60 & 47.50 \\*
 &  & \chromaswatch{0.576,1.000,0.678} & 0.40 & 26.20 & 15.00 & 3.20 & 11.20 \\*
 &  & \chromaswatch{0.122,1.000,0.929} & 0.00 & 0.20 & 0.00 & 0.00 & 0.05 \\*
 &  & \chromaswatch{0.412,0.839,1.000} & 29.00 & 84.40 & 51.00 & 11.20 & 43.90 \\*
 &  & \chromaswatch{0.745,0.710,1.000} & 95.80 & 98.60 & 92.60 & 65.20 & 88.05 \\*
 &  & \chromaswatch{1.000,0.667,0.953} & 94.20 & 98.80 & 95.60 & 72.40 & 90.25 \\*
 &  & \textit{Avg.} & 56.00 & 72.17 & 61.92 & 36.88 & 56.74 \\*
\cmidrule(lr){2-8}
 & \multirow{8}{=}{\shortstack[l]{$\rho=0.08$\\(strong)}} & \chromaswatch{1.000,0.556,0.653} & 95.80 & 99.00 & 94.40 & 67.00 & 89.05 \\*
 &  & \chromaswatch{1.000,0.705,0.390} & 92.40 & 94.60 & 85.40 & 40.40 & 78.20 \\*
 &  & \chromaswatch{0.960,1.000,0.277} & 25.20 & 62.00 & 43.60 & 4.80 & 33.90 \\*
 &  & \chromaswatch{0.399,1.000,0.569} & 0.00 & 0.00 & 1.00 & 0.00 & 0.25 \\*
 &  & \chromaswatch{0.000,1.000,0.909} & 0.00 & 0.00 & 0.00 & 0.00 & 0.00 \\*
 &  & \chromaswatch{0.025,0.803,1.000} & 0.20 & 43.60 & 16.80 & 2.20 & 15.70 \\*
 &  & \chromaswatch{0.688,0.642,1.000} & 96.40 & 98.80 & 91.00 & 62.40 & 87.15 \\*
 &  & \chromaswatch{1.000,0.578,0.942} & 96.00 & 99.40 & 93.40 & 68.60 & 89.35 \\*
 &  & \textit{Avg.} & 50.75 & 62.17 & 53.20 & 30.68 & 49.20 \\*
\cmidrule(lr){2-8}
 & \textbf{Overall avg.} &  & 53.38 & 67.17 & 57.56 & 33.77 & 52.97 \\
\midrule
\multirow{20}{=}{WM4A-Wan} & Original & \originalchromaswatch & 96.40 & 97.80 & 93.20 & 87.60 & 93.75 \\*
\cmidrule(lr){2-8}
 & \multirow{8}{=}{\shortstack[l]{$\rho=0.06$\\(moderate)}} & \chromaswatch{1.000,0.635,0.710} & 7.80 & 46.20 & 33.80 & 0.80 & 22.15 \\*
 &  & \chromaswatch{1.000,0.757,0.537} & 1.40 & 24.60 & 15.00 & 0.00 & 10.25 \\*
 &  & \chromaswatch{0.965,1.000,0.533} & 0.00 & 1.20 & 0.80 & 0.00 & 0.50 \\*
 &  & \chromaswatch{0.576,1.000,0.678} & 0.20 & 0.00 & 1.80 & 0.00 & 0.50 \\*
 &  & \chromaswatch{0.122,1.000,0.929} & 0.00 & 0.00 & 4.20 & 0.00 & 1.05 \\*
 &  & \chromaswatch{0.412,0.839,1.000} & 0.00 & 0.00 & 7.40 & 0.00 & 1.85 \\*
 &  & \chromaswatch{0.745,0.710,1.000} & 1.60 & 31.60 & 24.20 & 4.20 & 15.40 \\*
 &  & \chromaswatch{1.000,0.667,0.953} & 3.00 & 49.40 & 29.40 & 2.80 & 21.15 \\*
 &  & \textit{Avg.} & 1.75 & 19.12 & 14.57 & 0.97 & 9.11 \\*
\cmidrule(lr){2-8}
 & \multirow{8}{=}{\shortstack[l]{$\rho=0.08$\\(strong)}} & \chromaswatch{1.000,0.556,0.653} & 3.00 & 40.60 & 29.80 & 0.20 & 18.40 \\*
 &  & \chromaswatch{1.000,0.705,0.390} & 0.80 & 21.80 & 9.40 & 0.00 & 8.00 \\*
 &  & \chromaswatch{0.960,1.000,0.277} & 0.20 & 0.80 & 1.80 & 0.00 & 0.70 \\*
 &  & \chromaswatch{0.399,1.000,0.569} & 0.00 & 0.00 & 1.40 & 0.00 & 0.35 \\*
 &  & \chromaswatch{0.000,1.000,0.909} & 0.00 & 0.00 & 4.00 & 0.00 & 1.00 \\*
 &  & \chromaswatch{0.025,0.803,1.000} & 0.00 & 0.00 & 5.40 & 0.00 & 1.35 \\*
 &  & \chromaswatch{0.688,0.642,1.000} & 0.60 & 17.80 & 19.40 & 2.40 & 10.05 \\*
 &  & \chromaswatch{1.000,0.578,0.942} & 1.20 & 44.20 & 23.00 & 1.80 & 17.55 \\*
 &  & \textit{Avg.} & 0.72 & 15.65 & 11.78 & 0.55 & 7.17 \\*
\cmidrule(lr){2-8}
 & \textbf{Overall avg.} &  & 1.24 & 17.39 & 13.18 & 0.76 & 8.14 \\
\midrule
\multirow{20}{=}{PI0} & Original & \originalchromaswatch & 96.60 & 97.80 & 93.00 & 82.00 & 92.35 \\*
\cmidrule(lr){2-8}
 & \multirow{8}{=}{\shortstack[l]{$\rho=0.06$\\(moderate)}} & \chromaswatch{1.000,0.635,0.710} & 96.60 & 96.40 & 93.60 & 80.40 & 91.75 \\*
 &  & \chromaswatch{1.000,0.757,0.537} & 96.40 & 98.00 & 92.00 & 79.00 & 91.35 \\*
 &  & \chromaswatch{0.965,1.000,0.533} & 97.20 & 95.60 & 94.80 & 77.40 & 91.25 \\*
 &  & \chromaswatch{0.576,1.000,0.678} & 97.40 & 96.80 & 93.00 & 81.00 & 92.05 \\*
 &  & \chromaswatch{0.122,1.000,0.929} & 96.20 & 97.00 & 93.20 & 81.20 & 91.90 \\*
 &  & \chromaswatch{0.412,0.839,1.000} & 96.20 & 97.20 & 94.60 & 78.20 & 91.55 \\*
 &  & \chromaswatch{0.745,0.710,1.000} & 97.40 & 99.00 & 92.20 & 80.60 & 92.30 \\*
 &  & \chromaswatch{1.000,0.667,0.953} & 96.60 & 98.20 & 93.80 & 80.00 & 92.15 \\*
 &  & \textit{Avg.} & 96.75 & 97.28 & 93.40 & 79.72 & 91.79 \\*
\cmidrule(lr){2-8}
 & \multirow{8}{=}{\shortstack[l]{$\rho=0.08$\\(strong)}} & \chromaswatch{1.000,0.556,0.653} & 96.80 & 97.40 & 93.40 & 81.00 & 92.15 \\*
 &  & \chromaswatch{1.000,0.705,0.390} & 96.60 & 97.00 & 91.20 & 75.40 & 90.05 \\*
 &  & \chromaswatch{0.960,1.000,0.277} & 97.60 & 96.20 & 92.00 & 79.00 & 91.20 \\*
 &  & \chromaswatch{0.399,1.000,0.569} & 96.80 & 96.80 & 93.20 & 78.80 & 91.40 \\*
 &  & \chromaswatch{0.000,1.000,0.909} & 97.60 & 98.20 & 92.60 & 82.20 & 92.65 \\*
 &  & \chromaswatch{0.025,0.803,1.000} & 96.40 & 98.40 & 91.80 & 76.80 & 90.85 \\*
 &  & \chromaswatch{0.688,0.642,1.000} & 97.20 & 97.00 & 93.80 & 78.80 & 91.70 \\*
 &  & \chromaswatch{1.000,0.578,0.942} & 97.60 & 97.00 & 94.80 & 79.20 & 92.15 \\*
 &  & \textit{Avg.} & 97.08 & 97.25 & 92.85 & 78.90 & 91.52 \\*
\cmidrule(lr){2-8}
 & \textbf{Overall avg.} &  & 96.91 & 97.26 & 93.12 & 79.31 & 91.65 \\
\midrule
\multirow{20}{=}{PI05} & Original & \originalchromaswatch & 98.00 & 99.40 & 97.60 & 93.20 & 97.05 \\*
\cmidrule(lr){2-8}
 & \multirow{8}{=}{\shortstack[l]{$\rho=0.06$\\(moderate)}} & \chromaswatch{1.000,0.635,0.710} & 98.20 & 99.20 & 96.80 & 93.40 & 96.90 \\*
 &  & \chromaswatch{1.000,0.757,0.537} & 98.60 & 98.00 & 96.80 & 92.40 & 96.45 \\*
 &  & \chromaswatch{0.965,1.000,0.533} & 99.20 & 98.00 & 97.40 & 90.80 & 96.35 \\*
 &  & \chromaswatch{0.576,1.000,0.678} & 99.20 & 98.40 & 97.00 & 92.00 & 96.65 \\*
 &  & \chromaswatch{0.122,1.000,0.929} & 98.40 & 98.80 & 96.40 & 91.00 & 96.15 \\*
 &  & \chromaswatch{0.412,0.839,1.000} & 98.20 & 99.20 & 98.20 & 92.60 & 97.05 \\*
 &  & \chromaswatch{0.745,0.710,1.000} & 98.80 & 97.40 & 97.80 & 92.60 & 96.65 \\*
 &  & \chromaswatch{1.000,0.667,0.953} & 98.00 & 98.40 & 97.80 & 91.00 & 96.30 \\*
 &  & \textit{Avg.} & 98.58 & 98.42 & 97.28 & 91.97 & 96.56 \\*
\cmidrule(lr){2-8}
 & \multirow{8}{=}{\shortstack[l]{$\rho=0.08$\\(strong)}} & \chromaswatch{1.000,0.556,0.653} & 98.80 & 98.20 & 98.40 & 93.60 & 97.25 \\*
 &  & \chromaswatch{1.000,0.705,0.390} & 99.00 & 97.60 & 97.60 & 91.60 & 96.45 \\*
 &  & \chromaswatch{0.960,1.000,0.277} & 98.00 & 98.20 & 97.40 & 91.80 & 96.35 \\*
 &  & \chromaswatch{0.399,1.000,0.569} & 98.80 & 99.00 & 97.20 & 92.00 & 96.75 \\*
 &  & \chromaswatch{0.000,1.000,0.909} & 98.60 & 98.80 & 97.80 & 92.00 & 96.80 \\*
 &  & \chromaswatch{0.025,0.803,1.000} & 98.20 & 98.40 & 97.20 & 90.00 & 95.95 \\*
 &  & \chromaswatch{0.688,0.642,1.000} & 98.20 & 98.40 & 96.80 & 91.40 & 96.20 \\*
 &  & \chromaswatch{1.000,0.578,0.942} & 98.00 & 98.20 & 97.60 & 91.40 & 96.30 \\*
 &  & \textit{Avg.} & 98.45 & 98.35 & 97.50 & 91.72 & 96.51 \\*
\cmidrule(lr){2-8}
 & \textbf{Overall avg.} &  & 98.51 & 98.39 & 97.39 & 91.85 & 96.53 \\
\midrule
\multirow{4}{=}{All baselines} & Original & \originalchromaswatch & 97.33 & 98.77 & 95.57 & 88.00 & 94.92 \\*
\cmidrule(lr){2-8}
 & \shortstack[l]{$\rho=0.06$\\(moderate)} & \textit{Avg.} & 64.42 & 64.44 & 65.38 & 50.94 & 61.30 \\
\cmidrule(lr){2-8}
 & \shortstack[l]{$\rho=0.08$\\(strong)} & \textit{Avg.} & 63.37 & 62.21 & 63.31 & 49.53 & 59.60 \\
 & \textbf{Overall avg.} &  & 63.90 & 63.33 & 64.35 & 50.23 & 60.45 \\
\end{xltabular}
}

{\small
\setlength{\tabcolsep}{2.5pt}
\renewcommand{\arraystretch}{0.95}
\begin{xltabular}{\textwidth}{@{}>{\raggedright\arraybackslash\hsize=0.90\hsize}X >{\raggedright\arraybackslash\hsize=1.25\hsize}X >{\centering\arraybackslash\hsize=0.55\hsize}X *{5}{>{\centering\arraybackslash\hsize=1.06\hsize}X}@{}}
\caption{LIBERO success rates (\%) under relative color transformations for each VLA baseline. The dimensionless factor $s$ scales OKLab chroma, and each colored bar represents one hue-rotation direction. Avg. averages the eight evaluated directions.}\label{tab:appendix_colorrelation}\\
\toprule
Baseline & Color scale & Color & Spatial & Object & Goal & Long & Overall \\
\midrule
\endfirsthead
\endhead
\endfoot
\bottomrule
\endlastfoot
\multirow{20}{=}{Qwen3-OFT} & Original & \originalchromaswatch & 98.60 & 100.00 & 98.40 & 94.00 & 97.75 \\*
\cmidrule(lr){2-8}
 & \multirow{8}{=}{\shortstack[l]{$s=4.5$\\(moderate)}} & \chromaswatch{0.949,0.655,0.722} & 29.00 & 1.40 & 25.00 & 0.80 & 14.05 \\*
 &  & \chromaswatch{0.941,0.761,0.541} & 36.20 & 1.20 & 25.40 & 1.20 & 16.00 \\*
 &  & \chromaswatch{0.906,0.910,0.518} & 29.80 & 1.20 & 26.20 & 0.60 & 14.45 \\*
 &  & \chromaswatch{0.569,0.847,0.647} & 27.00 & 1.60 & 25.00 & 2.40 & 14.00 \\*
 &  & \chromaswatch{0.514,0.851,0.831} & 28.80 & 1.20 & 28.20 & 2.20 & 15.10 \\*
 &  & \chromaswatch{0.541,0.737,0.910} & 28.20 & 1.00 & 26.00 & 0.80 & 14.00 \\*
 &  & \chromaswatch{0.702,0.655,0.910} & 30.40 & 0.80 & 24.60 & 1.80 & 14.40 \\*
 &  & \chromaswatch{0.898,0.651,0.855} & 30.00 & 1.00 & 26.00 & 0.60 & 14.40 \\*
 &  & \textit{Avg.} & 29.93 & 1.18 & 25.80 & 1.30 & 14.55 \\*
\cmidrule(lr){2-8}
 & \multirow{8}{=}{\shortstack[l]{$s=5.5$\\(strong)}} & \chromaswatch{0.941,0.357,0.471} & 28.60 & 0.80 & 26.20 & 1.00 & 14.15 \\*
 &  & \chromaswatch{0.929,0.604,0.196} & 32.20 & 1.00 & 24.40 & 1.40 & 14.75 \\*
 &  & \chromaswatch{0.831,0.847,0.188} & 29.00 & 1.80 & 25.00 & 1.60 & 14.35 \\*
 &  & \chromaswatch{0.220,0.737,0.424} & 28.00 & 1.20 & 28.00 & 2.40 & 14.90 \\*
 &  & \chromaswatch{0.153,0.741,0.706} & 29.20 & 0.40 & 28.60 & 1.80 & 15.00 \\*
 &  & \chromaswatch{0.208,0.553,0.824} & 29.00 & 1.00 & 25.60 & 0.40 & 14.00 \\*
 &  & \chromaswatch{0.455,0.357,0.816} & 30.20 & 1.00 & 27.60 & 1.80 & 15.15 \\*
 &  & \chromaswatch{0.824,0.357,0.722} & 29.60 & 0.80 & 26.80 & 1.20 & 14.60 \\*
 &  & \textit{Avg.} & 29.48 & 1.00 & 26.52 & 1.45 & 14.61 \\*
\cmidrule(lr){2-8}
 & \textbf{Overall avg.} &  & 29.70 & 1.09 & 26.16 & 1.38 & 14.58 \\
\midrule
\multirow{20}{=}{Qwen3-PI} & Original & \originalchromaswatch & 98.40 & 99.00 & 98.00 & 95.20 & 97.65 \\*
\cmidrule(lr){2-8}
 & \multirow{8}{=}{\shortstack[l]{$s=4.5$\\(moderate)}} & \chromaswatch{0.949,0.655,0.722} & 98.40 & 98.60 & 97.40 & 94.20 & 97.15 \\*
 &  & \chromaswatch{0.941,0.761,0.541} & 98.00 & 96.80 & 97.40 & 94.80 & 96.75 \\*
 &  & \chromaswatch{0.906,0.910,0.518} & 98.80 & 97.60 & 97.40 & 91.20 & 96.25 \\*
 &  & \chromaswatch{0.569,0.847,0.647} & 98.40 & 98.40 & 96.60 & 94.00 & 96.85 \\*
 &  & \chromaswatch{0.514,0.851,0.831} & 98.20 & 98.40 & 97.60 & 94.20 & 97.10 \\*
 &  & \chromaswatch{0.541,0.737,0.910} & 98.80 & 99.00 & 96.20 & 93.20 & 96.80 \\*
 &  & \chromaswatch{0.702,0.655,0.910} & 98.40 & 99.20 & 96.20 & 93.00 & 96.70 \\*
 &  & \chromaswatch{0.898,0.651,0.855} & 98.20 & 98.80 & 97.20 & 95.00 & 97.30 \\*
 &  & \textit{Avg.} & 98.40 & 98.35 & 97.00 & 93.70 & 96.86 \\*
\cmidrule(lr){2-8}
 & \multirow{8}{=}{\shortstack[l]{$s=5.5$\\(strong)}} & \chromaswatch{0.941,0.357,0.471} & 98.40 & 98.60 & 96.80 & 93.40 & 96.80 \\*
 &  & \chromaswatch{0.929,0.604,0.196} & 98.00 & 97.80 & 97.60 & 94.40 & 96.95 \\*
 &  & \chromaswatch{0.831,0.847,0.188} & 96.00 & 97.80 & 97.00 & 91.20 & 95.50 \\*
 &  & \chromaswatch{0.220,0.737,0.424} & 98.40 & 98.60 & 96.20 & 92.00 & 96.30 \\*
 &  & \chromaswatch{0.153,0.741,0.706} & 97.20 & 98.20 & 97.80 & 94.00 & 96.80 \\*
 &  & \chromaswatch{0.208,0.553,0.824} & 98.00 & 98.00 & 95.80 & 93.80 & 96.40 \\*
 &  & \chromaswatch{0.455,0.357,0.816} & 99.00 & 99.20 & 97.40 & 94.80 & 97.60 \\*
 &  & \chromaswatch{0.824,0.357,0.722} & 98.60 & 98.00 & 97.80 & 94.00 & 97.10 \\*
 &  & \textit{Avg.} & 97.95 & 98.28 & 97.05 & 93.45 & 96.68 \\*
\cmidrule(lr){2-8}
 & \textbf{Overall avg.} &  & 98.17 & 98.31 & 97.03 & 93.58 & 96.77 \\
\midrule
\multirow{20}{=}{WM4A-Cosmo} & Original & \originalchromaswatch & 96.00 & 98.60 & 93.20 & 76.00 & 90.95 \\*
\cmidrule(lr){2-8}
 & \multirow{8}{=}{\shortstack[l]{$s=4.5$\\(moderate)}} & \chromaswatch{0.949,0.655,0.722} & 82.80 & 95.80 & 85.00 & 40.00 & 75.90 \\*
 &  & \chromaswatch{0.941,0.761,0.541} & 58.40 & 91.40 & 67.00 & 36.20 & 63.25 \\*
 &  & \chromaswatch{0.906,0.910,0.518} & 0.00 & 12.80 & 7.00 & 0.00 & 4.95 \\*
 &  & \chromaswatch{0.569,0.847,0.647} & 0.00 & 14.40 & 5.40 & 0.00 & 4.95 \\*
 &  & \chromaswatch{0.514,0.851,0.831} & 2.60 & 58.40 & 8.00 & 0.20 & 17.30 \\*
 &  & \chromaswatch{0.541,0.737,0.910} & 38.80 & 82.20 & 24.20 & 3.60 & 37.20 \\*
 &  & \chromaswatch{0.702,0.655,0.910} & 84.40 & 95.00 & 70.60 & 16.80 & 66.70 \\*
 &  & \chromaswatch{0.898,0.651,0.855} & 91.00 & 96.00 & 84.60 & 44.20 & 78.95 \\*
 &  & \textit{Avg.} & 44.75 & 68.25 & 43.98 & 17.62 & 43.65 \\*
\cmidrule(lr){2-8}
 & \multirow{8}{=}{\shortstack[l]{$s=5.5$\\(strong)}} & \chromaswatch{0.941,0.357,0.471} & 74.80 & 96.00 & 79.40 & 38.80 & 72.25 \\*
 &  & \chromaswatch{0.929,0.604,0.196} & 41.20 & 89.00 & 53.80 & 31.20 & 53.80 \\*
 &  & \chromaswatch{0.831,0.847,0.188} & 0.00 & 6.60 & 6.60 & 0.00 & 3.30 \\*
 &  & \chromaswatch{0.220,0.737,0.424} & 0.00 & 5.40 & 2.60 & 0.00 & 2.00 \\*
 &  & \chromaswatch{0.153,0.741,0.706} & 0.40 & 44.80 & 4.60 & 0.00 & 12.45 \\*
 &  & \chromaswatch{0.208,0.553,0.824} & 22.00 & 76.80 & 14.20 & 3.00 & 29.00 \\*
 &  & \chromaswatch{0.455,0.357,0.816} & 76.40 & 92.20 & 60.40 & 12.80 & 60.45 \\*
 &  & \chromaswatch{0.824,0.357,0.722} & 85.40 & 95.00 & 83.60 & 40.60 & 76.15 \\*
 &  & \textit{Avg.} & 37.52 & 63.23 & 38.15 & 15.80 & 38.67 \\*
\cmidrule(lr){2-8}
 & \textbf{Overall avg.} &  & 41.14 & 65.74 & 41.06 & 16.71 & 41.16 \\
\midrule
\multirow{20}{=}{WM4A-Wan} & Original & \originalchromaswatch & 96.40 & 97.80 & 93.20 & 87.60 & 93.75 \\*
\cmidrule(lr){2-8}
 & \multirow{8}{=}{\shortstack[l]{$s=4.5$\\(moderate)}} & \chromaswatch{0.949,0.655,0.722} & 32.00 & 42.00 & 30.40 & 10.20 & 28.65 \\*
 &  & \chromaswatch{0.941,0.761,0.541} & 33.20 & 46.20 & 38.60 & 9.80 & 31.95 \\*
 &  & \chromaswatch{0.906,0.910,0.518} & 0.80 & 0.00 & 17.60 & 0.00 & 4.60 \\*
 &  & \chromaswatch{0.569,0.847,0.647} & 0.00 & 0.20 & 3.00 & 0.00 & 0.80 \\*
 &  & \chromaswatch{0.514,0.851,0.831} & 0.00 & 0.00 & 3.40 & 0.00 & 0.85 \\*
 &  & \chromaswatch{0.541,0.737,0.910} & 0.00 & 0.00 & 1.80 & 0.00 & 0.45 \\*
 &  & \chromaswatch{0.702,0.655,0.910} & 0.00 & 0.40 & 3.80 & 0.00 & 1.05 \\*
 &  & \chromaswatch{0.898,0.651,0.855} & 13.80 & 20.00 & 16.20 & 5.20 & 13.80 \\*
 &  & \textit{Avg.} & 9.97 & 13.60 & 14.35 & 3.15 & 10.27 \\*
\cmidrule(lr){2-8}
 & \multirow{8}{=}{\shortstack[l]{$s=5.5$\\(strong)}} & \chromaswatch{0.941,0.357,0.471} & 18.80 & 30.60 & 23.00 & 12.60 & 21.25 \\*
 &  & \chromaswatch{0.929,0.604,0.196} & 20.60 & 37.80 & 27.60 & 11.00 & 24.25 \\*
 &  & \chromaswatch{0.831,0.847,0.188} & 0.20 & 0.00 & 9.00 & 0.00 & 2.30 \\*
 &  & \chromaswatch{0.220,0.737,0.424} & 0.00 & 0.00 & 2.80 & 0.00 & 0.70 \\*
 &  & \chromaswatch{0.153,0.741,0.706} & 0.00 & 0.00 & 2.20 & 0.00 & 0.55 \\*
 &  & \chromaswatch{0.208,0.553,0.824} & 0.00 & 0.00 & 0.40 & 0.00 & 0.10 \\*
 &  & \chromaswatch{0.455,0.357,0.816} & 0.00 & 0.00 & 2.20 & 0.00 & 0.55 \\*
 &  & \chromaswatch{0.824,0.357,0.722} & 8.80 & 11.00 & 8.60 & 5.80 & 8.55 \\*
 &  & \textit{Avg.} & 6.05 & 9.93 & 9.47 & 3.67 & 7.28 \\*
\cmidrule(lr){2-8}
 & \textbf{Overall avg.} &  & 8.01 & 11.76 & 11.91 & 3.41 & 8.78 \\
\midrule
\multirow{20}{=}{PI0} & Original & \originalchromaswatch & 96.60 & 97.80 & 93.00 & 82.00 & 92.35 \\*
\cmidrule(lr){2-8}
 & \multirow{8}{=}{\shortstack[l]{$s=4.5$\\(moderate)}} & \chromaswatch{0.949,0.655,0.722} & 95.60 & 96.60 & 91.00 & 78.60 & 90.45 \\*
 &  & \chromaswatch{0.941,0.761,0.541} & 96.00 & 97.40 & 93.20 & 80.80 & 91.85 \\*
 &  & \chromaswatch{0.906,0.910,0.518} & 97.80 & 93.40 & 91.40 & 76.80 & 89.85 \\*
 &  & \chromaswatch{0.569,0.847,0.647} & 97.20 & 94.20 & 92.40 & 78.00 & 90.45 \\*
 &  & \chromaswatch{0.514,0.851,0.831} & 96.80 & 96.40 & 92.40 & 78.60 & 91.05 \\*
 &  & \chromaswatch{0.541,0.737,0.910} & 97.00 & 96.00 & 94.60 & 77.20 & 91.20 \\*
 &  & \chromaswatch{0.702,0.655,0.910} & 96.60 & 95.20 & 94.00 & 77.40 & 90.80 \\*
 &  & \chromaswatch{0.898,0.651,0.855} & 96.60 & 95.80 & 94.40 & 79.60 & 91.60 \\*
 &  & \textit{Avg.} & 96.70 & 95.62 & 92.92 & 78.38 & 90.91 \\*
\cmidrule(lr){2-8}
 & \multirow{8}{=}{\shortstack[l]{$s=5.5$\\(strong)}} & \chromaswatch{0.941,0.357,0.471} & 97.20 & 96.00 & 90.20 & 79.00 & 90.60 \\*
 &  & \chromaswatch{0.929,0.604,0.196} & 97.20 & 96.20 & 91.80 & 78.60 & 90.95 \\*
 &  & \chromaswatch{0.831,0.847,0.188} & 96.00 & 96.60 & 89.40 & 77.00 & 89.75 \\*
 &  & \chromaswatch{0.220,0.737,0.424} & 97.40 & 95.80 & 91.80 & 75.60 & 90.15 \\*
 &  & \chromaswatch{0.153,0.741,0.706} & 97.40 & 95.40 & 90.40 & 79.20 & 90.60 \\*
 &  & \chromaswatch{0.208,0.553,0.824} & 95.80 & 94.80 & 91.60 & 79.40 & 90.40 \\*
 &  & \chromaswatch{0.455,0.357,0.816} & 96.00 & 96.60 & 94.40 & 80.40 & 91.85 \\*
 &  & \chromaswatch{0.824,0.357,0.722} & 98.20 & 95.60 & 93.00 & 79.60 & 91.60 \\*
 &  & \textit{Avg.} & 96.90 & 95.88 & 91.58 & 78.60 & 90.74 \\*
\cmidrule(lr){2-8}
 & \textbf{Overall avg.} &  & 96.80 & 95.75 & 92.25 & 78.49 & 90.82 \\
\midrule
\multirow{20}{=}{PI05} & Original & \originalchromaswatch & 98.00 & 99.40 & 97.60 & 93.20 & 97.05 \\*
\cmidrule(lr){2-8}
 & \multirow{8}{=}{\shortstack[l]{$s=4.5$\\(moderate)}} & \chromaswatch{0.949,0.655,0.722} & 98.20 & 98.40 & 96.80 & 91.60 & 96.25 \\*
 &  & \chromaswatch{0.941,0.761,0.541} & 98.60 & 99.20 & 97.40 & 93.20 & 97.10 \\*
 &  & \chromaswatch{0.906,0.910,0.518} & 97.20 & 96.80 & 97.60 & 92.20 & 95.95 \\*
 &  & \chromaswatch{0.569,0.847,0.647} & 98.20 & 96.80 & 97.40 & 94.40 & 96.70 \\*
 &  & \chromaswatch{0.514,0.851,0.831} & 98.40 & 96.60 & 96.60 & 94.20 & 96.45 \\*
 &  & \chromaswatch{0.541,0.737,0.910} & 98.20 & 97.20 & 98.60 & 92.60 & 96.65 \\*
 &  & \chromaswatch{0.702,0.655,0.910} & 98.40 & 99.20 & 96.20 & 91.60 & 96.35 \\*
 &  & \chromaswatch{0.898,0.651,0.855} & 98.60 & 97.60 & 97.80 & 90.80 & 96.20 \\*
 &  & \textit{Avg.} & 98.22 & 97.72 & 97.30 & 92.58 & 96.46 \\*
\cmidrule(lr){2-8}
 & \multirow{8}{=}{\shortstack[l]{$s=5.5$\\(strong)}} & \chromaswatch{0.941,0.357,0.471} & 97.40 & 98.60 & 97.80 & 91.60 & 96.35 \\*
 &  & \chromaswatch{0.929,0.604,0.196} & 98.40 & 99.00 & 97.60 & 91.00 & 96.50 \\*
 &  & \chromaswatch{0.831,0.847,0.188} & 98.40 & 96.80 & 98.00 & 93.20 & 96.60 \\*
 &  & \chromaswatch{0.220,0.737,0.424} & 97.60 & 96.00 & 96.40 & 90.80 & 95.20 \\*
 &  & \chromaswatch{0.153,0.741,0.706} & 98.20 & 97.80 & 97.40 & 91.20 & 96.15 \\*
 &  & \chromaswatch{0.208,0.553,0.824} & 98.80 & 97.40 & 98.00 & 92.20 & 96.60 \\*
 &  & \chromaswatch{0.455,0.357,0.816} & 98.20 & 98.00 & 97.20 & 92.40 & 96.45 \\*
 &  & \chromaswatch{0.824,0.357,0.722} & 98.40 & 98.00 & 97.20 & 91.60 & 96.30 \\*
 &  & \textit{Avg.} & 98.17 & 97.70 & 97.45 & 91.75 & 96.27 \\*
\cmidrule(lr){2-8}
 & \textbf{Overall avg.} &  & 98.20 & 97.71 & 97.38 & 92.16 & 96.36 \\
\midrule
\multirow{4}{=}{All baselines} & Original & \originalchromaswatch & 97.33 & 98.77 & 95.57 & 88.00 & 94.92 \\*
\cmidrule(lr){2-8}
 & \shortstack[l]{$s=4.5$\\(moderate)} & \textit{Avg.} & 63.00 & 62.45 & 61.89 & 47.79 & 58.78 \\
\cmidrule(lr){2-8}
 & \shortstack[l]{$s=5.5$\\(strong)} & \textit{Avg.} & 61.01 & 61.00 & 60.04 & 47.45 & 57.38 \\
 & \textbf{Overall avg.} &  & 62.00 & 61.73 & 60.96 & 47.62 & 58.08 \\
\end{xltabular}
}
{\small
\setlength{\tabcolsep}{3pt}
\begin{xltabular}{\textwidth}{@{}>{\raggedright\arraybackslash\hsize=1.00\hsize}X >{\raggedright\arraybackslash\hsize=0.95\hsize}X >{\centering\arraybackslash\hsize=0.65\hsize}X *{5}{>{\centering\arraybackslash\hsize=1.08\hsize}X}@{}}
\caption{LIBERO success rates (\%) under tonal-response perturbations for each VLA baseline. The dimensionless parameter $c$ controls tonal contrast, with $c=1$ denoting the default response. The vertical marker indicates the setting on the grayscale scale.}\label{tab:appendix_tonal}\\
\toprule
Baseline & Setting & Tonal & Spatial & Object & Goal & Long & Overall \\
\midrule
\endfirsthead
\endhead
\endfoot
\bottomrule
\endlastfoot
\multirow{7}{=}{Qwen3-OFT} & Original & \toneswatch{2.80ex} & 98.60 & 100.00 & 98.40 & 94.00 & 97.75 \\*
\cmidrule(lr){2-8}
 & $c=0.05$ & \toneswatch{0.00ex} & 8.40 & 0.00 & 5.60 & 0.20 & 3.55 \\*
 & $c=0.34$ & \toneswatch{1.79ex} & 35.60 & 1.20 & 24.40 & 0.80 & 15.50 \\*
 & $c=0.65$ & \toneswatch{2.40ex} & 34.00 & 0.40 & 27.00 & 1.20 & 15.65 \\*
 & $c=2.50$ & \toneswatch{3.83ex} & 37.60 & 2.80 & 23.20 & 1.60 & 16.30 \\*
 & $c=5.00$ & \toneswatch{4.61ex} & 32.20 & 3.40 & 23.00 & 0.00 & 14.65 \\*
 & $c=12.00$ & \toneswatch{5.35ex} & 33.00 & 1.00 & 26.20 & 0.00 & 15.05 \\
\midrule
\multirow{7}{=}{Qwen3-PI} & Original & \toneswatch{2.80ex} & 98.40 & 99.00 & 98.00 & 95.20 & 97.65 \\*
\cmidrule(lr){2-8}
 & $c=0.05$ & \toneswatch{0.00ex} & 7.60 & 47.20 & 13.80 & 19.00 & 21.90 \\*
 & $c=0.34$ & \toneswatch{1.79ex} & 98.40 & 98.40 & 96.60 & 92.60 & 96.50 \\*
 & $c=0.65$ & \toneswatch{2.40ex} & 97.40 & 98.20 & 98.40 & 94.40 & 97.10 \\*
 & $c=2.50$ & \toneswatch{3.83ex} & 97.80 & 98.00 & 96.00 & 89.80 & 95.40 \\*
 & $c=5.00$ & \toneswatch{4.61ex} & 91.40 & 83.20 & 80.60 & 18.20 & 68.35 \\*
 & $c=12.00$ & \toneswatch{5.35ex} & 77.60 & 64.40 & 59.80 & 4.40 & 51.55 \\
\midrule
\multirow{7}{=}{WM4A-Cosmo} & Original & \toneswatch{2.80ex} & 96.00 & 98.60 & 93.20 & 76.00 & 90.95 \\*
\cmidrule(lr){2-8}
 & $c=0.05$ & \toneswatch{0.00ex} & 5.40 & 0.00 & 0.00 & 0.00 & 1.35 \\*
 & $c=0.34$ & \toneswatch{1.79ex} & 91.00 & 91.00 & 78.20 & 53.00 & 78.30 \\*
 & $c=0.65$ & \toneswatch{2.40ex} & 95.20 & 97.00 & 89.80 & 63.20 & 86.30 \\*
 & $c=2.50$ & \toneswatch{3.83ex} & 92.00 & 96.40 & 80.20 & 30.60 & 74.80 \\*
 & $c=5.00$ & \toneswatch{4.61ex} & 84.00 & 85.20 & 72.00 & 5.20 & 61.60 \\*
 & $c=12.00$ & \toneswatch{5.35ex} & 41.20 & 41.80 & 49.00 & 0.60 & 33.15 \\
\midrule
\multirow{7}{=}{WM4A-Wan} & Original & \toneswatch{2.80ex} & 96.40 & 97.80 & 93.20 & 87.60 & 93.75 \\*
\cmidrule(lr){2-8}
 & $c=0.05$ & \toneswatch{0.00ex} & 0.20 & 0.00 & 0.00 & 0.00 & 0.05 \\*
 & $c=0.34$ & \toneswatch{1.79ex} & 0.60 & 0.00 & 1.40 & 0.20 & 0.55 \\*
 & $c=0.65$ & \toneswatch{2.40ex} & 8.00 & 17.40 & 22.40 & 4.80 & 13.15 \\*
 & $c=2.50$ & \toneswatch{3.83ex} & 63.40 & 81.40 & 74.60 & 8.20 & 56.90 \\*
 & $c=5.00$ & \toneswatch{4.61ex} & 51.00 & 58.20 & 49.40 & 0.00 & 39.65 \\*
 & $c=12.00$ & \toneswatch{5.35ex} & 14.80 & 11.60 & 16.40 & 0.00 & 10.70 \\
\midrule
\multirow{7}{=}{PI0} & Original & \toneswatch{2.80ex} & 96.60 & 97.80 & 93.00 & 82.00 & 92.35 \\*
\cmidrule(lr){2-8}
 & $c=0.05$ & \toneswatch{0.00ex} & 89.80 & 96.00 & 88.40 & 66.60 & 85.20 \\*
 & $c=0.34$ & \toneswatch{1.79ex} & 97.20 & 97.80 & 93.20 & 80.60 & 92.20 \\*
 & $c=0.65$ & \toneswatch{2.40ex} & 97.20 & 97.40 & 93.80 & 83.40 & 92.95 \\*
 & $c=2.50$ & \toneswatch{3.83ex} & 95.80 & 97.60 & 92.60 & 81.40 & 91.85 \\*
 & $c=5.00$ & \toneswatch{4.61ex} & 95.20 & 86.40 & 90.80 & 59.20 & 82.90 \\*
 & $c=12.00$ & \toneswatch{5.35ex} & 91.80 & 79.00 & 76.80 & 21.80 & 67.35 \\
\midrule
\multirow{7}{=}{PI05} & Original & \toneswatch{2.80ex} & 98.00 & 99.40 & 97.60 & 93.20 & 97.05 \\*
\cmidrule(lr){2-8}
 & $c=0.05$ & \toneswatch{0.00ex} & 98.20 & 96.20 & 95.80 & 77.20 & 91.85 \\*
 & $c=0.34$ & \toneswatch{1.79ex} & 97.60 & 98.60 & 97.60 & 93.00 & 96.70 \\*
 & $c=0.65$ & \toneswatch{2.40ex} & 99.00 & 98.60 & 98.20 & 92.40 & 97.05 \\*
 & $c=2.50$ & \toneswatch{3.83ex} & 98.40 & 98.40 & 97.60 & 91.60 & 96.50 \\*
 & $c=5.00$ & \toneswatch{4.61ex} & 98.80 & 98.80 & 97.40 & 78.80 & 93.45 \\*
 & $c=12.00$ & \toneswatch{5.35ex} & 97.00 & 96.40 & 88.60 & 38.40 & 80.10 \\
\end{xltabular}
}
\pagebreak
{\small
\setlength{\tabcolsep}{3pt}
\begin{xltabular}{\textwidth}{@{}>{\raggedright\arraybackslash\hsize=0.90\hsize}X >{\raggedright\arraybackslash\hsize=0.80\hsize}X >{\raggedright\arraybackslash\hsize=0.80\hsize}X *{5}{>{\centering\arraybackslash\hsize=1.10\hsize}X}@{}}
\caption{LIBERO success rates (\%) under uniform quantization for each VLA baseline. Bit denotes the precision $B$ in bits. RAW and RGB denote direct quantization of the RAW and RGB inputs, respectively.}\label{tab:appendix_bitdepth}\\
\toprule
Baseline & Input & Bit & Spatial & Object & Goal & Long & Overall \\
\midrule
\endfirsthead
\endhead
\endfoot
\bottomrule
\endlastfoot
\multirow{12}{=}{Qwen3-OFT} & Original & \bitswatch{8} & 98.60 & 100.00 & 98.40 & 94.00 & 97.75 \\*
\cmidrule(lr){2-8}
 & \multirow{5}{=}{RAW} & \bitswatch{6} & 99.00 & 99.60 & 98.80 & 92.80 & 97.55 \\*
 &  & \bitswatch{5} & 98.20 & 99.60 & 98.00 & 93.40 & 97.30 \\*
 &  & \bitswatch{4} & 95.80 & 75.20 & 96.00 & 91.20 & 89.55 \\*
 &  & \bitswatch{3} & 96.20 & 99.20 & 95.40 & 82.80 & 93.40 \\*
 &  & \bitswatch{2} & 60.40 & 38.00 & 37.20 & 13.40 & 37.25 \\*
\cmidrule(lr){2-8}
 & \multirow{6}{=}{RGB} & \bitswatch{7} & 99.20 & 99.80 & 98.80 & 94.80 & 98.15 \\*
 &  & \bitswatch{6} & 37.40 & 2.00 & 27.00 & 1.40 & 16.95 \\*
 &  & \bitswatch{5} & 34.20 & 0.80 & 25.60 & 2.40 & 15.75 \\*
 &  & \bitswatch{4} & 34.00 & 1.00 & 24.00 & 1.00 & 15.00 \\*
 &  & \bitswatch{3} & 39.80 & 0.20 & 25.20 & 0.40 & 16.40 \\*
 &  & \bitswatch{2} & 29.80 & 0.00 & 27.80 & 0.00 & 14.40 \\
\midrule
\multirow{12}{=}{Qwen3-PI} & Original & \bitswatch{8} & 98.40 & 99.00 & 98.00 & 95.20 & 97.65 \\*
\cmidrule(lr){2-8}
 & \multirow{5}{=}{RAW} & \bitswatch{6} & 98.80 & 98.40 & 97.20 & 84.00 & 94.60 \\*
 &  & \bitswatch{5} & 98.80 & 97.20 & 96.60 & 80.20 & 93.20 \\*
 &  & \bitswatch{4} & 97.20 & 95.00 & 97.20 & 80.40 & 92.45 \\*
 &  & \bitswatch{3} & 94.00 & 95.80 & 78.20 & 72.00 & 85.00 \\*
 &  & \bitswatch{2} & 54.00 & 73.40 & 47.40 & 13.00 & 46.95 \\*
\cmidrule(lr){2-8}
 & \multirow{6}{=}{RGB} & \bitswatch{7} & 98.40 & 99.00 & 97.40 & 95.40 & 97.55 \\*
 &  & \bitswatch{6} & 99.60 & 98.40 & 98.20 & 96.00 & 98.05 \\*
 &  & \bitswatch{5} & 98.60 & 99.00 & 97.80 & 95.00 & 97.60 \\*
 &  & \bitswatch{4} & 98.60 & 99.20 & 97.20 & 97.20 & 98.05 \\*
 &  & \bitswatch{3} & 96.80 & 98.80 & 97.60 & 89.00 & 95.55 \\*
 &  & \bitswatch{2} & 90.00 & 80.40 & 90.60 & 66.80 & 81.95 \\
\midrule
\multirow{12}{=}{WM4A-Cosmo} & Original & \bitswatch{8} & 96.00 & 98.60 & 93.20 & 76.00 & 90.95 \\*
\cmidrule(lr){2-8}
 & \multirow{5}{=}{RAW} & \bitswatch{6} & 56.20 & 80.60 & 64.00 & 29.40 & 57.55 \\*
 &  & \bitswatch{5} & 34.20 & 59.80 & 47.20 & 22.00 & 40.80 \\*
 &  & \bitswatch{4} & 4.80 & 18.40 & 10.00 & 15.20 & 12.10 \\*
 &  & \bitswatch{3} & 0.80 & 0.00 & 2.40 & 1.60 & 1.20 \\*
 &  & \bitswatch{2} & 0.00 & 0.00 & 0.00 & 0.00 & 0.00 \\*
\cmidrule(lr){2-8}
 & \multirow{6}{=}{RGB} & \bitswatch{7} & 95.40 & 98.80 & 91.60 & 68.80 & 88.65 \\*
 &  & \bitswatch{6} & 94.20 & 98.60 & 92.80 & 68.80 & 88.60 \\*
 &  & \bitswatch{5} & 95.60 & 99.20 & 89.80 & 69.60 & 88.55 \\*
 &  & \bitswatch{4} & 76.20 & 96.60 & 81.20 & 67.80 & 80.45 \\*
 &  & \bitswatch{3} & 55.80 & 91.20 & 80.80 & 43.00 & 67.70 \\*
 &  & \bitswatch{2} & 10.60 & 22.40 & 29.80 & 4.40 & 16.80 \\
\midrule
\multirow{12}{=}{WM4A-Wan} & Original & \bitswatch{8} & 96.40 & 97.80 & 93.20 & 87.60 & 93.75 \\*
\cmidrule(lr){2-8}
 & \multirow{5}{=}{RAW} & \bitswatch{6} & 0.40 & 4.60 & 10.40 & 0.00 & 3.85 \\*
 &  & \bitswatch{5} & 0.00 & 0.20 & 3.80 & 0.00 & 1.00 \\*
 &  & \bitswatch{4} & 0.20 & 0.00 & 5.60 & 0.00 & 1.45 \\*
 &  & \bitswatch{3} & 0.00 & 0.00 & 7.80 & 0.00 & 1.95 \\*
 &  & \bitswatch{2} & 1.40 & 0.00 & 1.20 & 0.00 & 0.65 \\*
\cmidrule(lr){2-8}
 & \multirow{6}{=}{RGB} & \bitswatch{7} & 26.40 & 50.20 & 43.20 & 13.00 & 33.20 \\*
 &  & \bitswatch{6} & 27.60 & 38.40 & 38.60 & 11.40 & 29.00 \\*
 &  & \bitswatch{5} & 14.20 & 13.60 & 30.60 & 3.20 & 15.40 \\*
 &  & \bitswatch{4} & 3.80 & 3.40 & 24.60 & 0.60 & 8.10 \\*
 &  & \bitswatch{3} & 1.00 & 1.40 & 32.00 & 0.40 & 8.70 \\*
 &  & \bitswatch{2} & 4.80 & 16.60 & 24.20 & 1.00 & 11.65 \\
\midrule\pagebreak[4]
\multirow{12}{=}{PI0} & Original & \bitswatch{8} & 96.60 & 97.80 & 93.00 & 82.00 & 92.35 \\*
\cmidrule(lr){2-8}
 & \multirow{5}{=}{RAW} & \bitswatch{6} & 97.40 & 95.80 & 91.40 & 80.00 & 91.15 \\*
 &  & \bitswatch{5} & 96.80 & 96.20 & 93.20 & 74.00 & 90.05 \\*
 &  & \bitswatch{4} & 96.60 & 95.40 & 89.80 & 78.40 & 90.05 \\*
 &  & \bitswatch{3} & 96.40 & 95.20 & 84.20 & 69.00 & 86.20 \\*
 &  & \bitswatch{2} & 83.20 & 91.40 & 74.40 & 22.40 & 67.85 \\*
\cmidrule(lr){2-8}
 & \multirow{6}{=}{RGB} & \bitswatch{7} & 97.00 & 97.00 & 93.20 & 79.40 & 91.65 \\*
 &  & \bitswatch{6} & 95.80 & 97.80 & 93.60 & 81.60 & 92.20 \\*
 &  & \bitswatch{5} & 97.40 & 97.20 & 92.00 & 79.00 & 91.40 \\*
 &  & \bitswatch{4} & 96.20 & 97.00 & 90.40 & 78.80 & 90.60 \\*
 &  & \bitswatch{3} & 95.40 & 95.60 & 89.20 & 76.80 & 89.25 \\*
 &  & \bitswatch{2} & 92.60 & 95.20 & 86.80 & 66.80 & 85.35 \\
\midrule
\multirow{12}{=}{PI05} & Original & \bitswatch{8} & 98.00 & 99.40 & 97.60 & 93.20 & 97.05 \\*
\cmidrule(lr){2-8}
 & \multirow{5}{=}{RAW} & \bitswatch{6} & 97.60 & 98.20 & 97.60 & 92.80 & 96.55 \\*
 &  & \bitswatch{5} & 97.00 & 98.80 & 96.60 & 93.80 & 96.55 \\*
 &  & \bitswatch{4} & 99.20 & 98.40 & 95.20 & 93.00 & 96.45 \\*
 &  & \bitswatch{3} & 98.80 & 98.40 & 96.00 & 86.60 & 94.95 \\*
 &  & \bitswatch{2} & 89.20 & 96.60 & 69.20 & 37.00 & 73.00 \\*
\cmidrule(lr){2-8}
 & \multirow{6}{=}{RGB} & \bitswatch{7} & 99.00 & 98.60 & 97.40 & 92.60 & 96.90 \\*
 &  & \bitswatch{6} & 98.40 & 98.20 & 97.00 & 93.60 & 96.80 \\*
 &  & \bitswatch{5} & 98.00 & 99.40 & 98.00 & 94.00 & 97.35 \\*
 &  & \bitswatch{4} & 99.40 & 98.20 & 98.60 & 92.40 & 97.15 \\*
 &  & \bitswatch{3} & 98.80 & 97.20 & 98.80 & 91.20 & 96.50 \\*
 &  & \bitswatch{2} & 98.00 & 96.40 & 94.20 & 80.40 & 92.25 \\
\end{xltabular}
}

\subsection{RoboTwin~2.0 ISP Perturbation Results}
\label{app:robotwin_isp_perturbations}

We additionally evaluate Qwen3-OFT and FastWAM across five matched ISP
perturbation axes on RoboTwin~2.0. The Easy and Hard results are shown side by
side, with each entry aggregating 50 rollouts on each of the same 13
representative tasks. For exposure, we report only the direct RGB input used
by both backbones. Original denotes the unperturbed RoboTwin~2.0 RGB
observation.

{\small
\setlength{\tabcolsep}{4pt}
\renewcommand{\arraystretch}{0.96}
\begin{xltabular}{\textwidth}{@{}>{\raggedright\arraybackslash}X >{\raggedright\arraybackslash}X *{4}{>{\centering\arraybackslash}X}@{}}
\caption{RoboTwin~2.0 success rates (\%) under uniform RGB quantization. Bit denotes the precision $B$ in bits. Results are aggregated over the shared 13-task subset.}
\label{tab:appendix_robotwin_bitdepth}\\
\toprule
Setting & Bit & \multicolumn{2}{c}{Qwen3-OFT} & \multicolumn{2}{c}{FastWAM} \\
\cmidrule(lr){3-4}\cmidrule(lr){5-6}
 & & Easy & Hard & Easy & Hard \\
\midrule
\endfirsthead
\endhead
\endfoot
\bottomrule
\endlastfoot
Original & \bitswatch{8} & 89.08 & 91.08 & 95.38 & 94.92 \\*
\cmidrule(lr){1-6}
$B=7$ & \bitswatch{7} & 90.46 & 91.67 & 95.85 & 94.31 \\*
$B=6$ & \bitswatch{6} & 90.31 & 92.36 & 94.62 & 95.54 \\*
$B=5$ & \bitswatch{5} & 90.31 & 91.54 & 93.38 & 95.08 \\*
$B=4$ & \bitswatch{4} & 89.54 & 89.69 & 91.85 & 92.77 \\*
$B=3$ & \bitswatch{3} & 88.77 & 87.23 & 90.15 & 83.08 \\*
$B=2$ & \bitswatch{2} & 70.92 & 70.31 & 69.69 & 48.46 \\
\end{xltabular}
}

{\small
\setlength{\tabcolsep}{4pt}
\renewcommand{\arraystretch}{0.96}
\begin{xltabular}{\textwidth}{@{}>{\raggedright\arraybackslash}X >{\centering\arraybackslash}X *{4}{>{\centering\arraybackslash}X}@{}}
\caption{RoboTwin~2.0 success rates (\%) for direct RGB inputs under exposure perturbations. EV denotes the exposure offset in stops, and the grayscale bar indicates relative brightness.}\label{tab:appendix_robotwin_exposure}\\
\toprule
Setting & Exposure & \multicolumn{2}{c}{Qwen3-OFT} & \multicolumn{2}{c}{FastWAM} \\
\cmidrule(lr){3-4}\cmidrule(lr){5-6}
 & & Easy & Hard & Easy & Hard \\
\midrule
\endfirsthead
\endhead
\endfoot
\bottomrule
\endlastfoot
Original & \originalexposureswatch & 89.08 & 91.08 & 95.38 & 94.92 \\*
\cmidrule(lr){1-6}
EV $-7$ & \exposureswatch{0.08} & 1.54 & 0.00 & 12.00 & 0.77 \\*
EV $-6$ & \exposureswatch{0.17} & 28.92 & 0.77 & 50.46 & 4.62 \\*
EV $-5$ & \exposureswatch{0.26} & 72.15 & 7.38 & 80.62 & 49.08 \\*
EV $-4$ & \exposureswatch{0.35} & 86.46 & 53.23 & 89.23 & 86.92 \\*
EV $-3$ & \exposureswatch{0.44} & 87.52 & 81.00 & 91.70 & 93.80 \\*
EV $-2$ & \exposureswatch{0.53} & 88.00 & 86.28 & 93.80 & 95.10 \\*
EV $-1$ & \exposureswatch{0.62} & 88.44 & 86.92 & 95.50 & 95.70 \\*
EV $+1$ & \exposureswatch{0.71} & 66.92 & 88.62 & 67.38 & 93.54 \\*
EV $+2$ & \exposureswatch{0.80} & 37.08 & 83.23 & 53.08 & 86.15 \\*
EV $+3$ & \exposureswatch{0.89} & 10.62 & 43.69 & 21.38 & 62.62 \\*
EV $+4$ & \exposureswatch{0.98} & 6.15 & 9.69 & 0.00 & 13.08 \\
\end{xltabular}
}

{\small
\setlength{\tabcolsep}{4pt}
\renewcommand{\arraystretch}{0.96}
\begin{xltabular}{\textwidth}{@{}>{\raggedright\arraybackslash}X >{\centering\arraybackslash}X *{4}{>{\centering\arraybackslash}X}@{}}
\caption{RoboTwin~2.0 success rates (\%) under physical RAW sensor-noise perturbations. Capture exposure decreases from EV $-1$ to EV $-6$, and increasing speckle density indicates increasing noise strength. Each reported result is aggregated over 650 episodes.}\label{tab:appendix_robotwin_noise}\\
\toprule
Noise level & Noise & \multicolumn{2}{c}{Qwen3-OFT} & \multicolumn{2}{c}{FastWAM} \\
\cmidrule(lr){3-4}\cmidrule(lr){5-6}
 & & Easy & Hard & Easy & Hard \\
\midrule
\endfirsthead
\endhead
\endfoot
\bottomrule
\endlastfoot
Original & \originalnoiseswatch & 89.08 & 91.08 & 95.38 & 94.92 \\*
\cmidrule(lr){1-6}
EV $-1$ & \noiseswatch{0} & 90.15 & 90.28 & 92.80 & 91.50 \\*
EV $-2$ & \noiseswatch{1} & 90.15 & 90.15 & 90.62 & 83.38 \\*
EV $-3$ & \noiseswatch{2} & 90.15 & 87.54 & 83.69 & 58.92 \\*
EV $-4$ & \noiseswatch{3} & 89.23 & 72.15 & 68.62 & 25.08 \\*
EV $-5$ & \noiseswatch{4} & 83.54 & 24.15 & 40.46 & 1.54 \\*
EV $-6$ & \noiseswatch{5} & 57.85 & 4.00 & 3.54 & 0.00 \\
\end{xltabular}
}

{\small
\setlength{\tabcolsep}{4pt}
\renewcommand{\arraystretch}{0.96}
\begin{xltabular}{\textwidth}{@{}>{\raggedright\arraybackslash}X >{\centering\arraybackslash}X *{4}{>{\centering\arraybackslash}X}@{}}
\caption{RoboTwin~2.0 success rates (\%) under white-point perturbations. The radius is fixed to $\rho=0.075$ around D65 in the $u'v'$ chromaticity plane, and each colored bar represents one shift direction.}\label{tab:appendix_robotwin_whitepoint}\\
\toprule
Direction & Color & \multicolumn{2}{c}{Qwen3-OFT} & \multicolumn{2}{c}{FastWAM} \\
\cmidrule(lr){3-4}\cmidrule(lr){5-6}
 & & Easy & Hard & Easy & Hard \\
\midrule
\endfirsthead
\endhead
\endfoot
\bottomrule
\endlastfoot
Original & \originalchromaswatch & 89.08 & 91.08 & 95.38 & 94.92 \\*
\cmidrule(lr){1-6}
$\theta=0^\circ$ & \chromaswatch{1.000,0.556,0.653} & 90.00 & 88.92 & 93.85 & 93.23 \\*
$\theta=45^\circ$ & \chromaswatch{1.000,0.705,0.390} & 89.23 & 89.54 & 94.15 & 93.08 \\*
$\theta=90^\circ$ & \chromaswatch{0.960,1.000,0.277} & 89.69 & 90.00 & 95.54 & 94.62 \\*
$\theta=135^\circ$ & \chromaswatch{0.399,1.000,0.569} & 89.38 & 89.54 & 94.31 & 95.38 \\*
$\theta=180^\circ$ & \chromaswatch{0.000,1.000,0.909} & 88.00 & 89.85 & 95.38 & 95.23 \\*
$\theta=225^\circ$ & \chromaswatch{0.025,0.803,1.000} & 86.77 & 90.15 & 93.85 & 92.77 \\*
$\theta=270^\circ$ & \chromaswatch{0.688,0.642,1.000} & 91.08 & 91.23 & 96.15 & 94.92 \\*
$\theta=315^\circ$ & \chromaswatch{1.000,0.578,0.942} & 89.54 & 92.15 & 96.00 & 95.23 \\
\end{xltabular}
}

{\small
\setlength{\tabcolsep}{4pt}
\renewcommand{\arraystretch}{0.96}
\begin{xltabular}{\textwidth}{@{}>{\raggedright\arraybackslash}X >{\centering\arraybackslash}X *{4}{>{\centering\arraybackslash}X}@{}}
\caption{RoboTwin~2.0 success rates (\%) under relative color transformations. The dimensionless factor is fixed to $s=5.0$ in OKLab space, and each colored bar represents one hue-rotation direction.}\label{tab:appendix_robotwin_colorrelation}\\
\toprule
Direction & Color & \multicolumn{2}{c}{Qwen3-OFT} & \multicolumn{2}{c}{FastWAM} \\
\cmidrule(lr){3-4}\cmidrule(lr){5-6}
 & & Easy & Hard & Easy & Hard \\
\midrule
\endfirsthead
\endhead
\endfoot
\bottomrule
\endlastfoot
Original & \originalchromaswatch & 89.08 & 91.08 & 95.38 & 94.92 \\*
\cmidrule(lr){1-6}
$\theta=0^\circ$ & \chromaswatch{0.949,0.655,0.722} & 88.15 & 86.92 & 92.31 & 94.15 \\*
$\theta=45^\circ$ & \chromaswatch{0.941,0.761,0.541} & 85.38 & 87.85 & 92.00 & 92.15 \\*
$\theta=90^\circ$ & \chromaswatch{0.906,0.910,0.518} & 87.23 & 90.31 & 88.15 & 87.54 \\*
$\theta=135^\circ$ & \chromaswatch{0.569,0.847,0.647} & 80.77 & 85.08 & 82.77 & 83.08 \\*
$\theta=180^\circ$ & \chromaswatch{0.514,0.851,0.831} & 79.08 & 82.00 & 80.00 & 76.31 \\*
$\theta=225^\circ$ & \chromaswatch{0.541,0.737,0.910} & 79.69 & 79.08 & 86.15 & 80.31 \\*
$\theta=270^\circ$ & \chromaswatch{0.702,0.655,0.910} & 77.08 & 78.62 & 85.69 & 83.08 \\*
$\theta=315^\circ$ & \chromaswatch{0.898,0.651,0.855} & 80.46 & 78.62 & 86.00 & 82.77 \\
\end{xltabular}
}

{\small
\setlength{\tabcolsep}{4pt}
\renewcommand{\arraystretch}{0.96}
\begin{xltabular}{\textwidth}{@{}>{\raggedright\arraybackslash}X >{\centering\arraybackslash}X *{4}{>{\centering\arraybackslash}X}@{}}
\caption{RoboTwin~2.0 success rates (\%) under tonal-response perturbations. The dimensionless parameter $c$ controls tonal contrast, with $c=1$ denoting the conceptual default response.}\label{tab:appendix_robotwin_tonal}\\
\toprule
Setting & Tonal & \multicolumn{2}{c}{Qwen3-OFT} & \multicolumn{2}{c}{FastWAM} \\
\cmidrule(lr){3-4}\cmidrule(lr){5-6}
 & & Easy & Hard & Easy & Hard \\
\midrule
\endfirsthead
\endhead
\endfoot
\bottomrule
\endlastfoot
Original & \toneswatch{2.80ex} & 89.08 & 91.08 & 95.38 & 94.92 \\*
\cmidrule(lr){1-6}
$c=0.05$ & \toneswatch{0.00ex} & 58.00 & 35.38 & 62.62 & 49.23 \\*
$c=0.34$ & \toneswatch{1.79ex} & 88.77 & 88.31 & 93.08 & 93.38 \\*
$c=0.65$ & \toneswatch{2.40ex} & 87.23 & 90.31 & 93.38 & 93.69 \\*
$c=1.50$ & \toneswatch{3.30ex} & 88.92 & 91.38 & 95.69 & 94.31 \\*
$c=2.50$ & \toneswatch{3.83ex} & 64.92 & 91.38 & 85.23 & 92.00 \\*
$c=3.50$ & \toneswatch{4.15ex} & 55.08 & 88.46 & 66.62 & 88.31 \\
\end{xltabular}
}

\section{Effect of RAW \& RGB Normalization}

Table~\ref{tab:appendix_ev_normalized} compares direct exposure normalization
in the RGB and RAW domains. Because linear RAW measurements retain a wider
range of scene-referred intensity information before nonlinear tone mapping,
display encoding, and associated clipping, they generally provide greater
exposure latitude than rendered RGB observations. Consequently, when the same
normalization principle is applied to both representations, RAW-domain
normalization preserves more recoverable shadow and highlight structure for
the subsequent ISP and policy. This leads to substantially higher average
success rates for most VLA baselines, particularly under severe exposure
shifts, although the magnitude of the benefit remains architecture dependent.
These results align with the motivation of \methodname{} described in
Section~\ref{sec:rawvla}, where retaining and adapting scene-referred
information before RGB rendering provides a more effective interface for
robust embodied control than correcting an already rendered observation.

{\small
\setlength{\tabcolsep}{2.2pt}
\renewcommand{\arraystretch}{0.95}
\begin{xltabular}{\textwidth}{@{}>{\raggedright\arraybackslash\hsize=0.85\hsize}X >{\raggedright\arraybackslash\hsize=1.05\hsize}X >{\raggedright\arraybackslash\hsize=0.45\hsize}X >{\centering\arraybackslash\hsize=0.75\hsize}X *{5}{>{\centering\arraybackslash\hsize=1.00\hsize}X}@{}}
\caption{LIBERO success rates (\%) after RGB- and RAW-domain exposure normalization for each VLA baseline. EV denotes the exposure offset in stops. Original is the default RGB input at EV $+0$.}\label{tab:appendix_ev_normalized}\\
\toprule
Baseline & Input & EV & Level & Spatial & Object & Goal & Long & Overall \\
\midrule
\endfirsthead
\endhead
\endfoot
\bottomrule
\endlastfoot
\multirow{21}{=}{Qwen3-OFT} & Original & $+0$ & \originalexposureswatch & 98.60 & 100.00 & 98.40 & 94.00 & 97.75 \\*
\cmidrule(lr){2-9}
 & \multirow{10}{=}{Normalized RGB} & $-8$ & \exposureswatch{0.04} & 0.00 & 0.00 & 0.00 & 0.00 & 0.00 \\*
 &  & $-7$ & \exposureswatch{0.08} & 0.20 & 0.00 & 0.00 & 0.00 & 0.05 \\*
 &  & $-6$ & \exposureswatch{0.13} & 96.00 & 90.40 & 72.80 & 34.80 & 73.50 \\*
 &  & $-5$ & \exposureswatch{0.20} & 34.20 & 1.60 & 22.60 & 0.20 & 14.65 \\*
 &  & $-4$ & \exposureswatch{0.30} & 38.20 & 1.60 & 20.80 & 0.80 & 15.35 \\*
 &  & $+1$ & \exposureswatch{0.67} & 37.60 & 1.60 & 27.60 & 2.00 & 17.20 \\*
 &  & $+2$ & \exposureswatch{0.78} & 31.60 & 2.40 & 25.00 & 0.60 & 14.90 \\*
 &  & $+3$ & \exposureswatch{0.89} & 13.00 & 47.60 & 68.80 & 56.00 & 46.35 \\*
 &  & $+4$ & \exposureswatch{0.98} & 0.00 & 0.00 & 9.80 & 0.00 & 2.45 \\*
 &  &  & \textit{Avg.} & 27.87 & 16.13 & 27.49 & 10.49 & 20.49 \\*
\cmidrule(lr){2-9}
 & \multirow{10}{=}{Normalized RAW} & $-8$ & \exposureswatch{0.04} & 90.40 & 88.80 & 82.00 & 20.20 & 70.35 \\*
 &  & $-7$ & \exposureswatch{0.08} & 98.40 & 99.00 & 95.00 & 93.20 & 96.40 \\*
 &  & $-6$ & \exposureswatch{0.13} & 98.60 & 99.00 & 99.20 & 89.80 & 96.65 \\*
 &  & $-5$ & \exposureswatch{0.20} & 36.00 & 2.00 & 26.40 & 0.20 & 16.15 \\*
 &  & $-4$ & \exposureswatch{0.30} & 36.40 & 2.00 & 25.80 & 1.20 & 16.35 \\*
 &  & $+1$ & \exposureswatch{0.67} & 36.20 & 2.20 & 26.40 & 1.80 & 16.65 \\*
 &  & $+2$ & \exposureswatch{0.78} & 35.00 & 1.60 & 22.00 & 1.60 & 15.05 \\*
 &  & $+3$ & \exposureswatch{0.89} & 37.80 & 84.40 & 83.00 & 50.80 & 64.00 \\*
 &  & $+4$ & \exposureswatch{0.98} & 0.00 & 0.00 & 8.40 & 0.00 & 2.10 \\*
 &  &  & \textit{Avg.} & 52.09\positivechange{24.22} & 42.11\positivechange{25.98} & 52.02\positivechange{24.53} & 28.76\positivechange{18.27} & 43.74\positivechange{23.25} \\
\midrule\pagebreak[4]
\multirow{21}{=}{Qwen3-PI} & Original & $+0$ & \originalexposureswatch & 98.40 & 99.00 & 98.00 & 95.20 & 97.65 \\*
\cmidrule(lr){2-9}
 & \multirow{10}{=}{Normalized RGB} & $-8$ & \exposureswatch{0.04} & 0.00 & 0.00 & 3.60 & 0.00 & 0.90 \\*
 &  & $-7$ & \exposureswatch{0.08} & 1.20 & 0.00 & 0.00 & 5.40 & 1.65 \\*
 &  & $-6$ & \exposureswatch{0.13} & 85.80 & 88.40 & 66.60 & 27.40 & 67.05 \\*
 &  & $-5$ & \exposureswatch{0.20} & 97.80 & 98.00 & 95.00 & 92.40 & 95.80 \\*
 &  & $-4$ & \exposureswatch{0.30} & 99.40 & 96.20 & 97.60 & 94.20 & 96.85 \\*
 &  & $+1$ & \exposureswatch{0.67} & 98.80 & 98.60 & 98.40 & 95.20 & 97.75 \\*
 &  & $+2$ & \exposureswatch{0.78} & 98.00 & 93.80 & 97.80 & 81.00 & 92.65 \\*
 &  & $+3$ & \exposureswatch{0.89} & 5.00 & 21.20 & 48.00 & 39.00 & 28.30 \\*
 &  & $+4$ & \exposureswatch{0.98} & 0.00 & 0.60 & 13.00 & 4.80 & 4.60 \\*
 &  &  & \textit{Avg.} & 54.00 & 55.20 & 57.78 & 48.82 & 53.95 \\*
\cmidrule(lr){2-9}
 & \multirow{10}{=}{Normalized RAW} & $-8$ & \exposureswatch{0.04} & 91.20 & 81.60 & 67.00 & 7.80 & 61.90 \\*
 &  & $-7$ & \exposureswatch{0.08} & 93.20 & 96.80 & 81.40 & 82.60 & 88.50 \\*
 &  & $-6$ & \exposureswatch{0.13} & 97.60 & 97.00 & 96.60 & 78.00 & 92.30 \\*
 &  & $-5$ & \exposureswatch{0.20} & 97.60 & 97.40 & 95.80 & 78.80 & 92.40 \\*
 &  & $-4$ & \exposureswatch{0.30} & 98.20 & 97.20 & 96.80 & 80.00 & 93.05 \\*
 &  & $+1$ & \exposureswatch{0.67} & 98.80 & 98.40 & 96.00 & 82.00 & 93.80 \\*
 &  & $+2$ & \exposureswatch{0.78} & 98.60 & 96.20 & 97.00 & 82.20 & 93.50 \\*
 &  & $+3$ & \exposureswatch{0.89} & 10.20 & 51.40 & 51.80 & 30.00 & 35.85 \\*
 &  & $+4$ & \exposureswatch{0.98} & 0.20 & 1.20 & 11.80 & 0.00 & 3.30 \\*
 &  &  & \textit{Avg.} & 76.18\positivechange{22.18} & 79.69\positivechange{24.49} & 77.13\positivechange{19.36} & 57.93\positivechange{9.11} & 72.73\positivechange{18.78} \\
\midrule
\multirow{21}{=}{WM4A-Cosmo} & Original & $+0$ & \originalexposureswatch & 96.00 & 98.60 & 93.20 & 76.00 & 90.95 \\*
\cmidrule(lr){2-9}
 & \multirow{10}{=}{Normalized RGB} & $-8$ & \exposureswatch{0.04} & 0.00 & 0.00 & 0.00 & 0.00 & 0.00 \\*
 &  & $-7$ & \exposureswatch{0.08} & 0.00 & 0.00 & 0.00 & 0.00 & 0.00 \\*
 &  & $-6$ & \exposureswatch{0.13} & 4.20 & 15.00 & 16.20 & 2.20 & 9.40 \\*
 &  & $-5$ & \exposureswatch{0.20} & 58.20 & 93.40 & 63.60 & 13.80 & 57.25 \\*
 &  & $-4$ & \exposureswatch{0.30} & 72.00 & 94.60 & 75.80 & 45.80 & 72.05 \\*
 &  & $+1$ & \exposureswatch{0.67} & 95.40 & 97.00 & 91.20 & 62.40 & 86.50 \\*
 &  & $+2$ & \exposureswatch{0.78} & 70.60 & 73.40 & 69.00 & 40.80 & 63.45 \\*
 &  & $+3$ & \exposureswatch{0.89} & 0.00 & 0.00 & 1.20 & 1.40 & 0.65 \\*
 &  & $+4$ & \exposureswatch{0.98} & 0.00 & 0.00 & 0.00 & 0.00 & 0.00 \\*
 &  &  & \textit{Avg.} & 33.38 & 41.49 & 35.22 & 18.49 & 32.14 \\*
\cmidrule(lr){2-9}
 & \multirow{10}{=}{Normalized RAW} & $-8$ & \exposureswatch{0.04} & 0.00 & 0.00 & 0.00 & 0.00 & 0.00 \\*
 &  & $-7$ & \exposureswatch{0.08} & 0.00 & 0.00 & 9.80 & 2.80 & 3.15 \\*
 &  & $-6$ & \exposureswatch{0.13} & 7.20 & 17.80 & 32.40 & 7.40 & 16.20 \\*
 &  & $-5$ & \exposureswatch{0.20} & 30.40 & 62.40 & 46.00 & 21.40 & 40.05 \\*
 &  & $-4$ & \exposureswatch{0.30} & 52.60 & 82.00 & 61.60 & 26.40 & 55.65 \\*
 &  & $+1$ & \exposureswatch{0.67} & 71.20 & 85.80 & 66.40 & 31.80 & 63.80 \\*
 &  & $+2$ & \exposureswatch{0.78} & 47.20 & 72.20 & 50.20 & 25.00 & 48.65 \\*
 &  & $+3$ & \exposureswatch{0.89} & 0.00 & 9.20 & 21.20 & 6.60 & 9.25 \\*
 &  & $+4$ & \exposureswatch{0.98} & 0.00 & 0.00 & 0.00 & 0.00 & 0.00 \\*
 &  &  & \textit{Avg.} & 23.18\negativechange{10.20} & 36.60\negativechange{4.89} & 31.96\negativechange{3.27} & 13.49\negativechange{5.00} & 26.31\negativechange{5.84} \\
\midrule
\multirow{21}{=}{WM4A-Wan} & Original & $+0$ & \originalexposureswatch & 96.40 & 97.80 & 93.20 & 87.60 & 93.75 \\*
\cmidrule(lr){2-9}
 & \multirow{10}{=}{Normalized RGB} & $-8$ & \exposureswatch{0.04} & 0.00 & 0.00 & 0.00 & 0.00 & 0.00 \\*
 &  & $-7$ & \exposureswatch{0.08} & 0.20 & 0.00 & 0.20 & 0.00 & 0.10 \\*
 &  & $-6$ & \exposureswatch{0.13} & 5.00 & 18.80 & 11.80 & 0.00 & 8.90 \\*
 &  & $-5$ & \exposureswatch{0.20} & 1.20 & 2.00 & 15.60 & 0.00 & 4.70 \\*
 &  & $-4$ & \exposureswatch{0.30} & 0.00 & 0.00 & 2.60 & 0.00 & 0.65 \\*
 &  & $+1$ & \exposureswatch{0.67} & 27.40 & 48.40 & 41.40 & 9.80 & 31.75 \\*
 &  & $+2$ & \exposureswatch{0.78} & 0.40 & 5.80 & 12.20 & 1.20 & 4.90 \\*
 &  & $+3$ & \exposureswatch{0.89} & 0.40 & 0.00 & 1.60 & 0.00 & 0.50 \\*
 &  & $+4$ & \exposureswatch{0.98} & 0.40 & 0.00 & 0.00 & 0.00 & 0.10 \\*
 &  &  & \textit{Avg.} & 3.89 & 8.33 & 9.49 & 1.22 & 5.73 \\*
\cmidrule(lr){2-9}
 & \multirow{10}{=}{Normalized RAW} & $-8$ & \exposureswatch{0.04} & 0.20 & 0.00 & 0.60 & 0.00 & 0.20 \\*
 &  & $-7$ & \exposureswatch{0.08} & 0.00 & 0.00 & 7.60 & 0.00 & 1.90 \\*
 &  & $-6$ & \exposureswatch{0.13} & 0.00 & 0.00 & 2.60 & 0.00 & 0.65 \\*
 &  & $-5$ & \exposureswatch{0.20} & 0.00 & 0.60 & 6.60 & 0.00 & 1.80 \\*
 &  & $-4$ & \exposureswatch{0.30} & 0.00 & 5.20 & 11.20 & 0.00 & 4.10 \\*
 &  & $+1$ & \exposureswatch{0.67} & 4.00 & 12.40 & 18.20 & 0.00 & 8.65 \\*
 &  & $+2$ & \exposureswatch{0.78} & 0.20 & 8.60 & 7.00 & 0.00 & 3.95 \\*
 &  & $+3$ & \exposureswatch{0.89} & 0.80 & 0.00 & 2.20 & 0.00 & 0.75 \\*
 &  & $+4$ & \exposureswatch{0.98} & 0.00 & 0.00 & 0.00 & 0.00 & 0.00 \\*
 &  &  & \textit{Avg.} & 0.58\negativechange{3.31} & 2.98\negativechange{5.36} & 6.22\negativechange{3.27} & 0.00\negativechange{1.22} & 2.44\negativechange{3.29} \\
\midrule
\multirow{21}{=}{PI0} & Original & $+0$ & \originalexposureswatch & 96.60 & 97.80 & 93.00 & 82.00 & 92.35 \\*
\cmidrule(lr){2-9}
 & \multirow{10}{=}{Normalized RGB} & $-8$ & \exposureswatch{0.04} & 0.00 & 0.00 & 7.80 & 0.00 & 1.95 \\*
 &  & $-7$ & \exposureswatch{0.08} & 46.00 & 16.40 & 28.20 & 12.00 & 25.65 \\*
 &  & $-6$ & \exposureswatch{0.13} & 93.40 & 93.60 & 78.40 & 36.00 & 75.35 \\*
 &  & $-5$ & \exposureswatch{0.20} & 96.60 & 96.60 & 91.80 & 75.00 & 90.00 \\*
 &  & $-4$ & \exposureswatch{0.30} & 97.80 & 97.80 & 92.60 & 80.20 & 92.10 \\*
 &  & $+1$ & \exposureswatch{0.67} & 95.20 & 97.80 & 93.80 & 81.00 & 91.95 \\*
 &  & $+2$ & \exposureswatch{0.78} & 95.20 & 90.80 & 92.80 & 65.20 & 86.00 \\*
 &  & $+3$ & \exposureswatch{0.89} & 59.60 & 71.00 & 68.60 & 44.40 & 60.90 \\*
 &  & $+4$ & \exposureswatch{0.98} & 1.40 & 20.20 & 20.20 & 15.20 & 14.25 \\*
 &  &  & \textit{Avg.} & 65.02 & 64.91 & 63.80 & 45.44 & 59.79 \\*
\cmidrule(lr){2-9}
 & \multirow{10}{=}{Normalized RAW} & $-8$ & \exposureswatch{0.04} & 93.40 & 91.40 & 78.60 & 25.40 & 72.20 \\*
 &  & $-7$ & \exposureswatch{0.08} & 96.00 & 95.60 & 90.60 & 66.60 & 87.20 \\*
 &  & $-6$ & \exposureswatch{0.13} & 96.00 & 95.60 & 91.60 & 79.20 & 90.60 \\*
 &  & $-5$ & \exposureswatch{0.20} & 97.40 & 97.00 & 94.40 & 76.80 & 91.40 \\*
 &  & $-4$ & \exposureswatch{0.30} & 96.20 & 97.40 & 91.80 & 81.60 & 91.75 \\*
 &  & $+1$ & \exposureswatch{0.67} & 98.00 & 97.60 & 93.00 & 81.80 & 92.60 \\*
 &  & $+2$ & \exposureswatch{0.78} & 97.60 & 97.00 & 92.00 & 80.60 & 91.80 \\*
 &  & $+3$ & \exposureswatch{0.89} & 90.00 & 85.00 & 91.00 & 58.60 & 81.15 \\*
 &  & $+4$ & \exposureswatch{0.98} & 21.40 & 42.40 & 42.20 & 32.40 & 34.60 \\*
 &  &  & \textit{Avg.} & 87.33\positivechange{22.31} & 88.78\positivechange{23.87} & 85.02\positivechange{21.22} & 64.78\positivechange{19.33} & 81.48\positivechange{21.68} \\
\midrule
\multirow{21}{=}{PI05} & Original & $+0$ & \originalexposureswatch & 98.00 & 99.40 & 97.60 & 93.20 & 97.05 \\*
\cmidrule(lr){2-9}
 & \multirow{10}{=}{Normalized RGB} & $-8$ & \exposureswatch{0.04} & 0.00 & 0.00 & 0.40 & 0.00 & 0.10 \\*
 &  & $-7$ & \exposureswatch{0.08} & 23.80 & 7.80 & 17.00 & 18.80 & 16.85 \\*
 &  & $-6$ & \exposureswatch{0.13} & 97.00 & 97.80 & 80.20 & 58.40 & 83.35 \\*
 &  & $-5$ & \exposureswatch{0.20} & 97.80 & 96.20 & 96.80 & 92.00 & 95.70 \\*
 &  & $-4$ & \exposureswatch{0.30} & 98.00 & 97.80 & 95.80 & 92.60 & 96.05 \\*
 &  & $+1$ & \exposureswatch{0.67} & 98.60 & 98.40 & 96.40 & 94.40 & 96.95 \\*
 &  & $+2$ & \exposureswatch{0.78} & 98.40 & 97.40 & 97.20 & 85.60 & 94.65 \\*
 &  & $+3$ & \exposureswatch{0.89} & 3.00 & 14.00 & 78.40 & 63.20 & 39.65 \\*
 &  & $+4$ & \exposureswatch{0.98} & 0.00 & 0.00 & 13.00 & 14.00 & 6.75 \\*
 &  &  & \textit{Avg.} & 57.40 & 56.60 & 63.91 & 57.67 & 58.89 \\*
\cmidrule(lr){2-9}
 & \multirow{10}{=}{Normalized RAW} & $-8$ & \exposureswatch{0.04} & 98.00 & 98.20 & 89.80 & 48.80 & 83.70 \\*
 &  & $-7$ & \exposureswatch{0.08} & 98.80 & 96.80 & 97.00 & 90.40 & 95.75 \\*
 &  & $-6$ & \exposureswatch{0.13} & 98.80 & 97.60 & 97.80 & 92.60 & 96.70 \\*
 &  & $-5$ & \exposureswatch{0.20} & 98.40 & 98.60 & 98.80 & 91.80 & 96.90 \\*
 &  & $-4$ & \exposureswatch{0.30} & 98.00 & 98.60 & 97.20 & 91.40 & 96.30 \\*
 &  & $+1$ & \exposureswatch{0.67} & 98.60 & 98.40 & 96.00 & 90.20 & 95.80 \\*
 &  & $+2$ & \exposureswatch{0.78} & 98.60 & 97.60 & 97.20 & 91.00 & 96.10 \\*
 &  & $+3$ & \exposureswatch{0.89} & 95.80 & 96.60 & 97.60 & 78.40 & 92.10 \\*
 &  & $+4$ & \exposureswatch{0.98} & 0.80 & 0.40 & 42.00 & 38.40 & 20.40 \\*
 &  &  & \textit{Avg.} & 87.31\positivechange{29.91} & 86.98\positivechange{30.38} & 90.38\positivechange{26.47} & 79.22\positivechange{21.56} & 85.97\positivechange{27.08} \\
\end{xltabular}
}

\bibliography{iclr2027_conference}
\bibliographystyle{iclr2027_conference}

\end{document}